\documentclass{article}

\usepackage{microtype}
\usepackage{graphicx}
\usepackage{subfigure}
\usepackage{booktabs}
\usepackage{hyperref}

\usepackage[accepted]{icml2024}
\usepackage{amsmath}
\usepackage{amssymb}
\usepackage{mathtools}
\usepackage{amsthm}
\usepackage[capitalize,noabbrev]{cleveref}
\usepackage[textsize=tiny]{todonotes}

\usepackage{listings}

\theoremstyle{plain}

\theoremstyle{definition}

\theoremstyle{remark}

\icmltitlerunning{Hidden Holes}

\begin{document}

\twocolumn[
\icmltitle{Hidden Holes \\ \large topological aspects of language models}



\icmlsetsymbol{equal}{*}

\begin{icmlauthorlist}
\icmlauthor{Stephen Fitz}{keio}
\icmlauthor{Peter Romero}{keio}
\icmlauthor{Jiyan Jonas Schneider}{keio}
\end{icmlauthorlist}

\icmlaffiliation{keio}{Keio University, Tokyo, Japan}

\icmlcorrespondingauthor{Stephen Fitz}{mail@stephenfitz.net}

\icmlkeywords{Language Models, Topological Data Analysis, Natural Language Processing, Machine Learning, Deep Learning, Neural Networks, Representation Learning}

\vskip 0.3in
]



\printAffiliationsAndNotice{}  

\begin{abstract}
We explore the topology of representation manifolds arising in autoregressive neural language models trained on raw text data.
In order to study their properties, we introduce tools from computational algebraic topology, which we use as a basis for a measure of topological complexity, that we call \emph{perforation}.

Using this measure, we study the evolution of topological structure in GPT based large language models across depth and time during training.
We then compare these to gated recurrent models, and show that the latter exhibit more topological complexity, with a distinct pattern of changes common to all natural languages but absent from synthetically generated data. The paper presents a detailed analysis of the representation manifolds derived by these models based on studying the shapes of vector clouds induced by them as they are conditioned on sentences from corpora of natural language text.

The methods developed in this paper are novel in the field and based on mathematical apparatus that might be unfamiliar to the target audience.
To help with that we introduce the minimum necessary theory, and provide additional visualizations in the appendices.

The main contribution of the paper is a striking observation about the topological structure of the transformer as compared to LSTM based neural architectures.
It suggests that further research into mathematical properties of these neural networks is necessary to understand the operation of large transformer language models. We hope this work inspires further explorations in this direction within the NLP community.
\end{abstract}

\section{Introduction}
\label{introduction}

\begin{figure}[!h]
  \centering
  \includegraphics[width=0.48\textwidth]{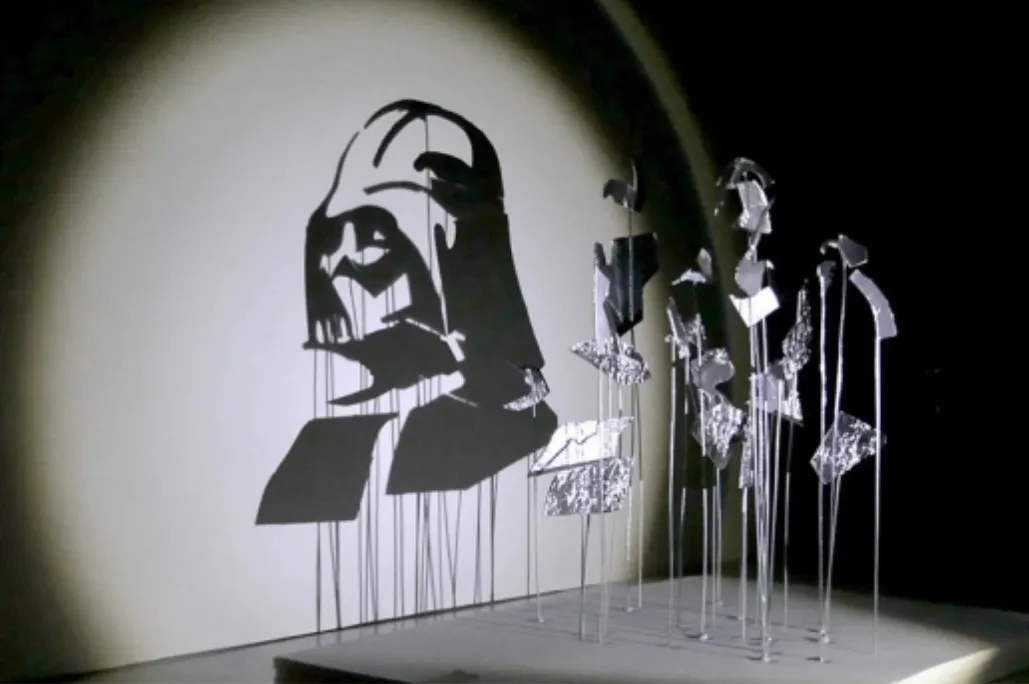}
  \caption{An art installation by "Red" Hong Yi \cite{franco2016amazing} showing the projection of a 3D sculpture onto a 2D plane. The projection can be misleading, while the original higher dimensional data has a more complex structure. In this analogy the text output from an LLM is the projected image. Our study develops tools to describe the shape of the internal representation manifolds (the sculpture) directly. We also track the evolution of those shapes throughout model training, and how they change with respect to choices of training data used.}
  \label{projection_sculpture}
\end{figure}

Large language models and NLP systems based on them are currently at the forefront of research and applications of artificial intelligence.
Most efforts in this area, however, focus on analyzing the model outputs.
Significantly less work has been done so far on studying the structure of their internal representations.
The power of neural network models stems from their ability to derive informative representations of inputs into sub-manifolds of high-dimensional real vector spaces.
Ultimately, this mapping of discrete text input into manifolds induced from large quantities of raw text is at the core of emergent abilities in large language models.
It is precisely the topology and geometry of those vector spaces that allow AI assistants such as Chat-GPT to perform their cognitive functions.
When we interact with language models on the level of their output, we are looking at the projections of these high dimensional representations back onto a discrete representation.
The natural topological structure of these language embeddings, which define the \emph{"thoughts"} of the system, is lost in this projection.
We are thus looking at shadows of complex high-dimensional objects, that could have nontrivial shapes (see figure \ref{projection_sculpture}).

This paper investigates topological aspects of representation spaces induced by neural network models trained on natural language data.
Instead of working directly with model outputs, we investigate how the internal structure of the hidden layers evolves during training as the models process inputs corresponding to natural language text and contrast it with synthetically generated inputs.
In analogy to studying a human subject, our approach would be akin to analyzing the brain activity of a person as they are reading a book, and comparing it to the brain activity of a person who is reading unnatural text generated by a random process, as opposed to relying on subject's introspection through dialogue.
Our method is therefore based on intrinsic evaluations, which is in contrast to most contemporary approaches, which are predominantly extrinsic and behavioral in nature.
In these mainstream methods, the abilities of the model are examined with techniques based on prompting and analyzing the outputs, or by proxy of performance on downstream tasks.
Instead, we focus on the analysis of activation patterns within the models' neural networks from a topological perspective.
The tools we use here, are novel in the field of natural language processing, and we hope this paper inspires the AI community to look into these and similar methods in future studies.

Although the largest NLP systems are currently based on transformer architectures such as GPT, gated recurrent models, such as LSTM have been used extensively, and are still deployed in many AI systems dealing with natural language inputs.
Furthermore, recurrent neural networks are increasingly combined with self-attention architectures such as GPT, leading to improved performance and new abilities of these augmented models \cite{bulatov2023scaling}.
Recurrent models based on state space architectures recently outperformed transformers on language modelling as well as downstream NLP tasks, while exhibiting superior scaling properties \cite{gu2023mamba}.
We applied our analysis to both transformer and recurrent language models in order to compare the topological structure of their representation manifolds.
Studies of algebraic and topological aspects of hidden state trajectories in gated recurrent networks can lead to re-parametrization and topological regularization techniques making these architectures more suitable to LLM setting, especially when combined with transformer architectures.

As of now, it is still a mystery why certain behaviors emerge in large language models, and little is understood about the structure of their representation manifolds.
The more we know about properties of embedding spaces emergent in the context of language models trained on raw text data, the closer we get to answering these questions.
Additionally, neural language models, as it currently  stands, tend to be extremely inefficient and over-parametrized.
Understanding the structure of their representation spaces can help re-parametrization efforts aiming at model compression and development of more sustainable NLP systems.
In this paper we explore topological aspects of neural language models in an effort to better understand the shapes of their activation manifolds.

\section{Relation to Other Work}
\label{relation_to_other_work}

Gunnar Carlson applied topological data analysis to patches of pixels from naturally occurring images \cite{carlsson2008local}. 
The analysis led to a conclusion that the shape of the image manifold under study could be approximated by a Klein bottle. 
This realization led to a novel compression algorithm for images taking advantage of a parametrization of the pixel space that mapped 3x3 patches of images onto points on a sub-manifold homeomorphic to the Klein bottle within the image manifold. 
To the best of our knowledge such approaches have not been applied in the field of natural language processing.
Recent developments in deep learning hint at the utility of understanding topological structure of data in improving representational power of neural information processing systems.
For instance, \cite{fuchs2020se} show that augmenting the self-attention mechanism of transformer architectures with an inductive prior encoding roto-translational structure leads to significant gains in robustness for point cloud and graph data modeling.
The idea of encoding manifold structure by algebraically tying synaptic connection weights is at the core of advancements in neural artificial intelligence \cite{hinton2011transforming} \cite{jumper2021highly}.
It is likely that augmentations to neural architectures informed by deep understanding of topological structure of data, will provide a source of future advancements in this area.
A better understanding of the structure of representation spaces induced by neural networks from training data can aid research efforts aimed at gaining deeper insights into the inner workings of black box neural systems such as large language models.
These insights will be useful to other researchers working on topics such as:
\begin{itemize}
  \item producing more informative or compressed representations
  \item improving the robustness of neural models
  \item interpretability of AI systems
  \item improving the parameter efficiency of large models, which might have a lower intrinsic dimensionality than the number of parameters used
  \item new methods of regularization with topological priors
  \item theoretical understanding of the representational power of neural networks
  \item new approaches to the design and training of neural network systems
  \item rethinking current approaches to language modeling
\end{itemize}

In order to make further progress in many of these areas, it will become increasingly important to develop new techniques for probing the internal activity of large neural systems.
The work presented here is a step in this direction.

\section{Methods}

\begin{figure}[!h]
	\centering
	\includegraphics[width=0.3\textwidth]{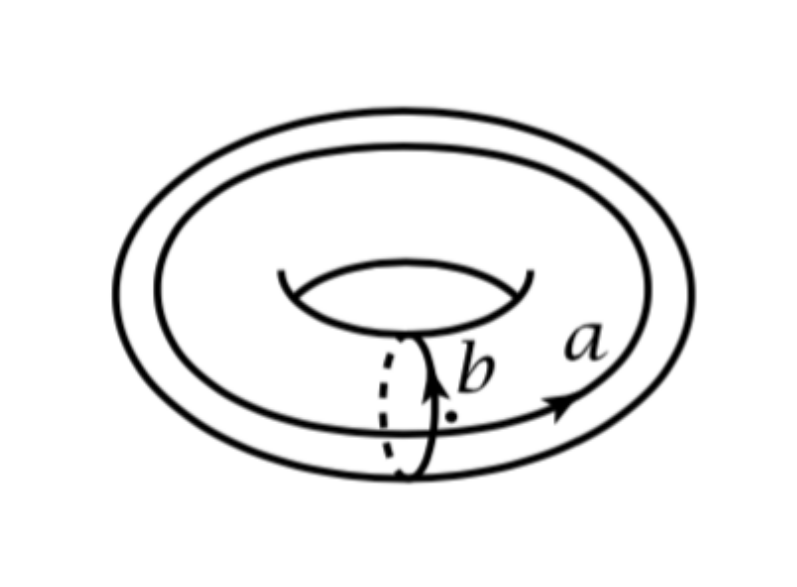}
	\includegraphics[width=0.2\textwidth]{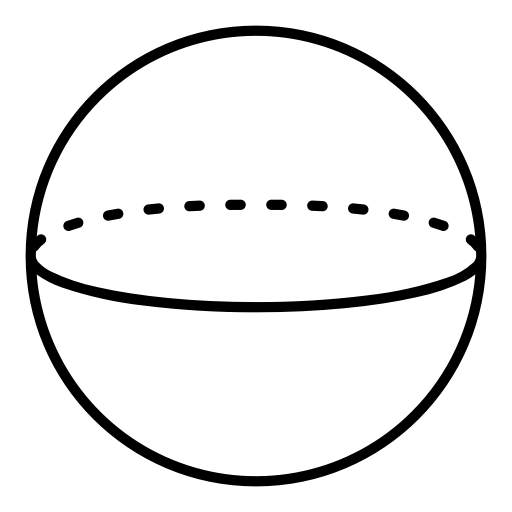}
  \caption{The torus (top) has a single connected component which corresponds to a single $0$-dimensional "hole". It has two $1$-dimensional holes - one measured by the class of loops wrapping around the main circle (represented by $a$) and another by the class of loops going through the middle hole (represented by $b$). These loops are independent because there is no way to continuously deform any of the loops in the first class into any of the loops in the second class. Finally, the torus has a single $2$-dimensional hole, which is generated by the tire shaped cavity inside its surface. These numbers of independent holes are the \emph{Betti numbers} of the torus. Thus, the Betti numbers for the torus are [1, 2, 1]. The sphere (bottom) by contrast has no non-trivial loops in the first dimension. This is because every loop on a sphere can be continuously deformed into any other loop on its surface, and they are all contractible to a point. The Betti numbers of the sphere are thus [1, 0, 1]. Both of these are 2-dimensional manifolds embedded in a 3-dimensional ambient space, so the sequence of Betti numbers has only 3 elements. In general, for higher dimensional manifolds such as those we study in this work, the sequence will continue and the n-th number can be interpreted as counting n-dimensional cells wrapped around (n+1)-dimensional cavities in the manifold's topological structure.}
	\label{betti_numbers}
\end{figure}

\begin{figure}[!h]
  \centering
  \includegraphics[width=0.48\textwidth]{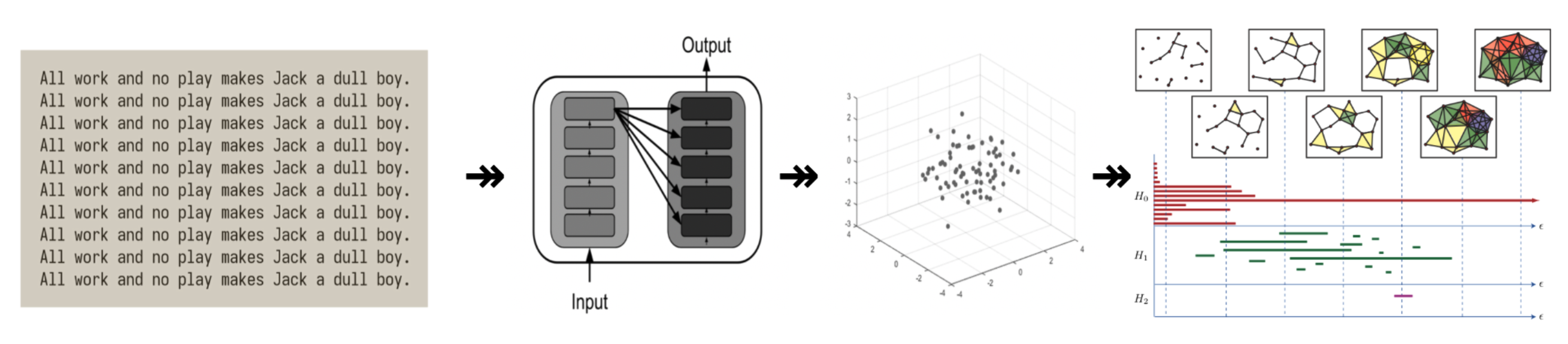}
  \caption{The main data analysis pipeline. Raw text is transformed into a sequence of high dimensional point clouds by the neural layers of a language model. The shapes of these point clouds are then analyzed from a topological perspective by computing persistent homology modules. The free ranks of these algebraic objects keep track of the number of holes emerging within the manifolds from which these point clouds are sampled. We also used additional topological analysis methods, which are explained in the appendices.}
  \label{analysis_pipeline}
\end{figure}

Our method can be decomposed into the following stages, which we repeat at each epoch of training:

\begin{enumerate}
  \item select a sample of sentences from the training corpus
  \item feed the sentences into the language model and record the hidden layer activations from the model (the hidden state)
  \item compute topological features of the hidden state
\end{enumerate}

The representations collected in step 2 were stored in separate tensors - one per each sentence processed by the models.
They form a unit of analysis for the topological computations performed in step 3.
Each sentence generates a 3D tensor of floating point values with the following axes: \lstinline{number_of_tokens} $\times$ \lstinline{state_dimension} $\times$ \lstinline{number_of_epochs}, where \lstinline{state_dimension} refers to the hidden state dimension, which can be one of the following: self-attention layer output, recurrent cell state, token embedding in case of the input layer.

We apply three different methods of topological analysis to the hidden state tensors: persistent homology, simplicial mapping approximation, sliding window embedding.
The main approach is based on persistence modules \cite{zomorodian2004computing}.
Detailed description is provided in the appendices.
We can interpreted the information these methods provide in the following way.
The tokens of a sentence are transformed into clouds of points (vectors associated with the tokens) by the neural network layers.
These clouds can be thought of as finite samples from neighborhoods of some underlying manifolds.
These methods produce data that is subsequently summarized into sequences of integers (traditionally called Betti numbers in classical homology theory), which count the number of independent holes (homology classes) in those hidden manifolds (see figure \ref{betti_numbers} for examples).
Finally, we use those integer sequences to compute a summary statistic, which we call \emph{perforation}.
It is a measure of topological complexity of the hidden manifolds, which we use to track the evolution of the representation spaces induced by the neural networks as they are trained on natural language data.
We designed perforation to be a simple and intuitive measure of topological complexity, which can be easily computed from the output of the topological analysis methods we use.
It has the following useful properties:
\begin{itemize}
  \item it is a positive scalar, which increases with the number of holes in the hidden manifolds
  \item higher dimensional holes are given more weight, as they require more complex topological structure to produce
  \item it bijectively encodes the Betti number sequences of the hidden manifolds, which is a homotopy invariant - meaning that topologically equivalent shapes will produce the same perforation
\end{itemize}

\begin{center}
  \setlength{\fboxsep}{1em}
  \setlength{\fboxrule}{0.1em}
  \noindent\fcolorbox{black}{lightgray}{%
    \minipage[t]{\dimexpr0.98\linewidth-2\fboxsep-2\fboxrule\relax}
    \textbf{\underline{Definition - perforation:}}
    \newline
    \newline
    Given a vector space representation of a sentence $s$ under its language model induced from a corpus of text, as described previously, let $n$ be the maximum dimension of its \emph{holes} (as measured by topological analysis methods - c.f. Appendix C). Then the \textbf{perforation} of the given sentence $s$ is defined to be $\phi(s) = H_1\log{2} + H_2\log{3} + \dots + H_n\log{p_n}$. Here $p_n$ denotes the n-th prime, and $H_n$ is the number of independent \emph{holes} (free rank of homology) in dimension $n$. Hence, perforation is the sum of Betti numbers, weighted by logarithms of consecutive primes.
    \endminipage
  }
\end{center}

\textbf{Proposition:} Perforation is a homotopy invariant.
\newline
\textbf{Proof:}
Let $\phi(s)$ be the perforation of some sentence $s$.
That is $\phi(s) = \sum_{i=1}^{n} H_i\log{p_i}$, where $H_i$ is the count of holes in dimension $i$ of the manifold approximated by the point cloud of representation vectors corresponding to the tokens of $s$ under some language model, and $p_i$ is the i-th prime number.
Note that the quantity $p = e^{\phi(s)}$ (where $e$ is the Euler constant) is an integer, and is of the form $2^{H_1}*3^{H_2}*\dots*p_n^{H_n}$.
Due to the Fundamental Theorem of Arithmetic, this exponentiated perforation uniquely encodes the sequence of persistent Betti numbers because there is only one way to factor it as a product of primes as above. Since Betti numbers are a homotopy invariant \cite{hatcher2001algebraic}, and perforation can always be decoded into the sequence of Betti numbers by performing exponentiation and prime factorization as above, it follows that perforation is a homotopy invariant (as it bijectively encodes the sequence of homology ranks). $\blacksquare$

\begin{figure}
  \centering
  \includegraphics[width=0.48\textwidth]{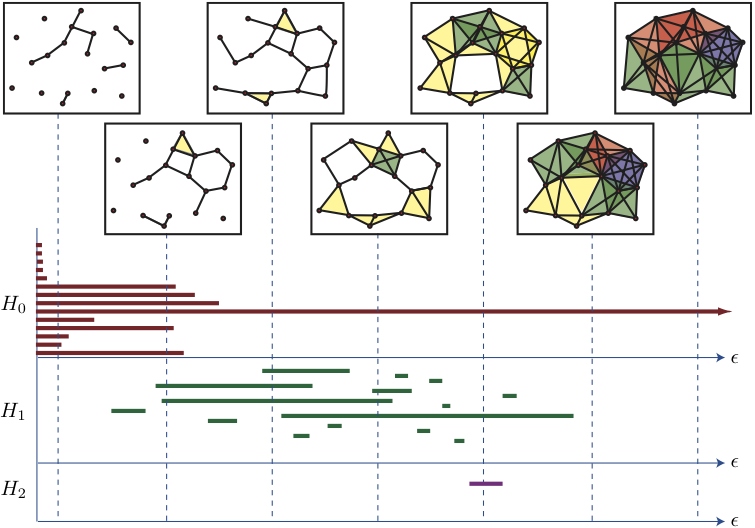}
  \caption{A filtration of Vietoris-Rips complexes with distance parameter $\epsilon$ on a set of points embedded in an ambient metric space, and the associated persistent homology barcodes. \cite{ghrist2008barcodes}}
  \label{persistence}
\end{figure}

\begin{figure}
	\centering
	\includegraphics[width=0.48\textwidth]{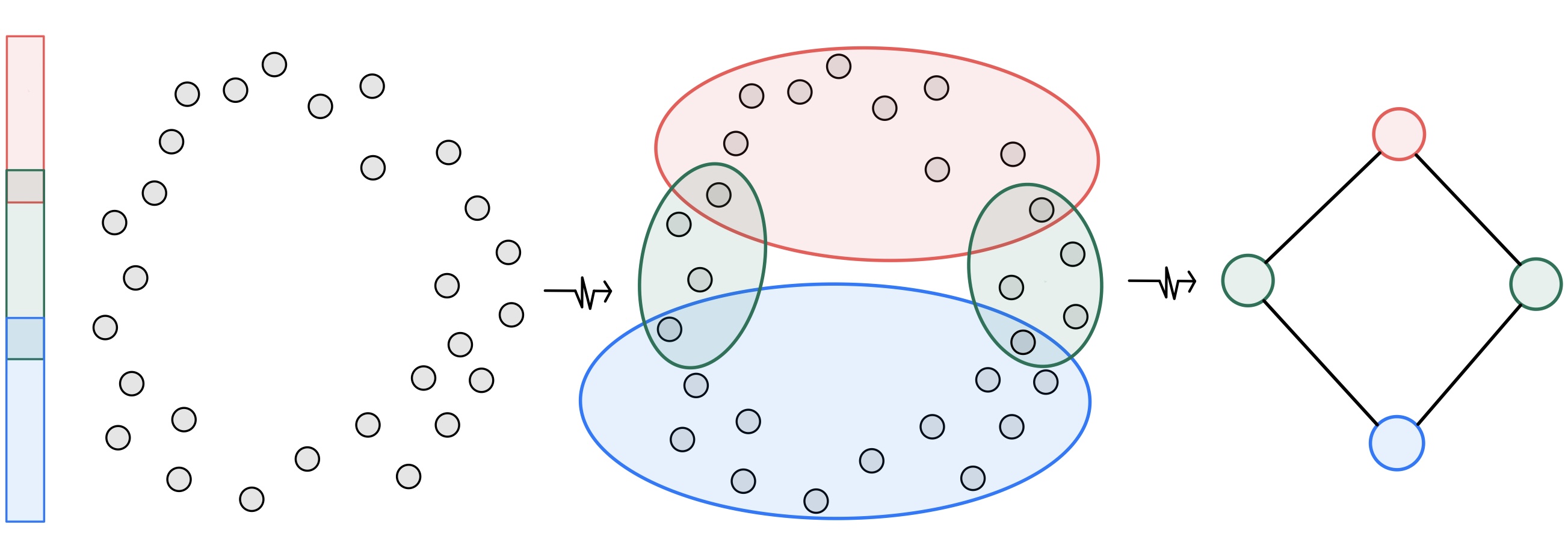}
  \caption{Inducing topological structure from a point cloud representing noisy samples from a neighborhood of a 1-dimensional submanifold ($\mathbb{S}^2$) of a 2-dimensional ambient embedding space ($\mathbb{R}^2$). This method can be adjusted to produce simplicial complex summaries in all dimensions \cite{singh2007topological}.}
	\label{mapper}
\end{figure}

\begin{figure}
	\centering
	\includegraphics[width=0.48\textwidth]{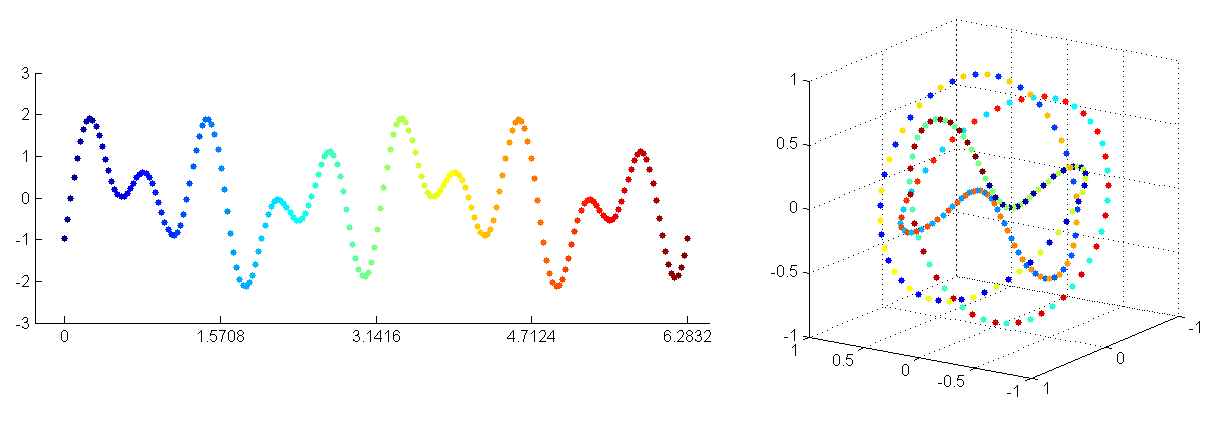}
        \caption{We can reinterpret a time series of values as a geometric object by performing a sliding window embedding \cite{perea2015sliding}. The resulting point cloud can then be interpreted as noisy samples from an underlying manifold. The topology of this manifold can be studied using tools from computational algebraic topology. It reveals intrinsic properties of the original time series that are not easily captured by standard methods.}
	\label{sliding_window_embedding}
\end{figure}

We count the number of persistent homology classes \cite{edelsbrunner2008persistent} as a proxy for estimating the number of holes in the manifold underlying our vector clouds.
In order to do that, we compute persistence barcode diagrams \cite{ghrist2008barcodes} from a Vietoris-Rips filtration of the point cloud of activations associated to each sentence.
We grow $\epsilon$ balls around each token vector in the ambient representation space, and record patterns of intersections between sets of neighborhoods as the value of $\epsilon$ radius increases.
The bars indicate when a hole (in various dimensions) is born and when it dies.
Holes are born when they are formed, and die when they are filled by a higher dimensional object.
For instance, a 2-dimensional hole is born when a pattern of surfaces (generated by triple intersections) of vector neighborhoods surrounds a 3-dimensional cavity in the ambient space.
It dies at a larger $\epsilon$ value, when the cavity is filled by a 3-dimensional volume.
These changes are recorded by bars that start at the birth and end at death values of $\epsilon$ associated to each hole (see figure \ref{persistence} for an illustration of the process, and refer to Appendix C for details).

We also apply topological methods based on approximating the shapes of our point clouds with a simplicial complex of a lower dimension.
This involves clustering the points based on a cover of a projection to a lower dimensional subspace (see figure \ref{mapper} for an example and Appendix C for details).
Finally we look at another method of topological data analysis suitable to time series data.
We view our sentences as time series of token embeddings and apply a sliding window to re-represent them as a collection of point clouds generated from considering consecutive tokens together (see figure \ref{sliding_window_embedding} for an illustration).

\section{Results}

We started by analyzing the data from the recurrent models in order to establish a baseline for the topological complexity of the representation spaces induced by neural language models.
We then compared the results to the transformer models.

\begin{figure}
	\centering
  \includegraphics[width=0.22\textwidth]{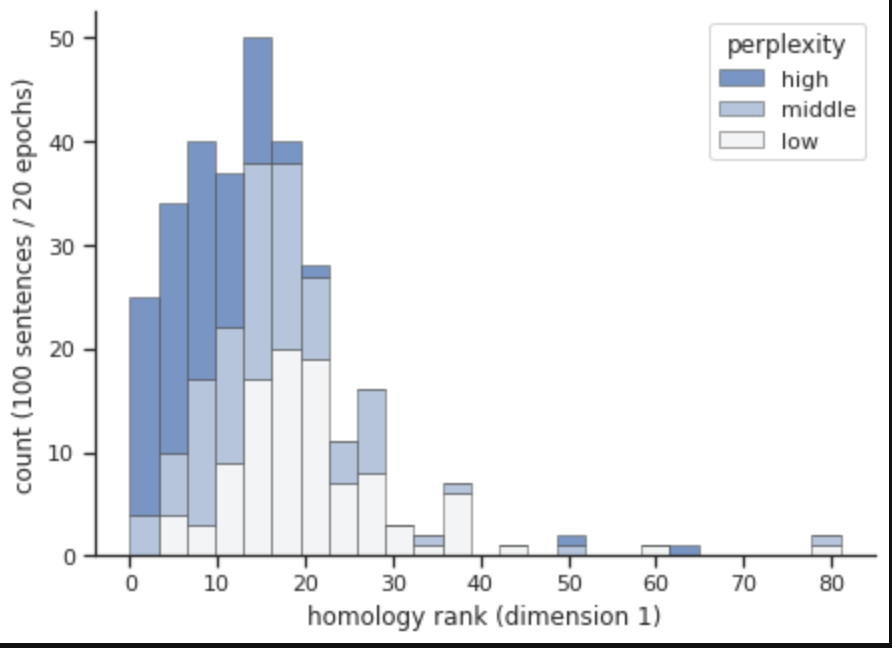}
	\includegraphics[width=0.22\textwidth]{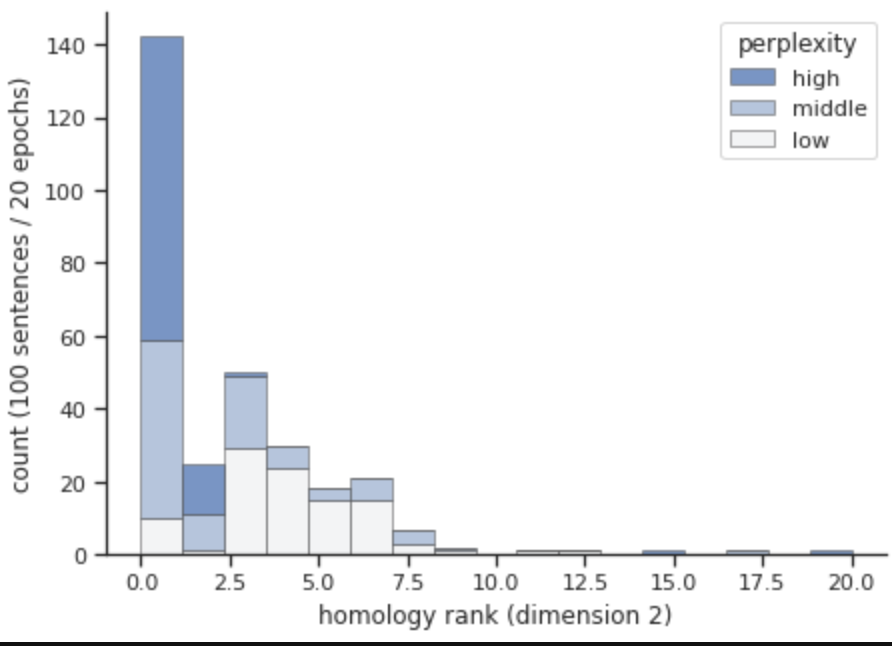}
	\includegraphics[width=0.22\textwidth]{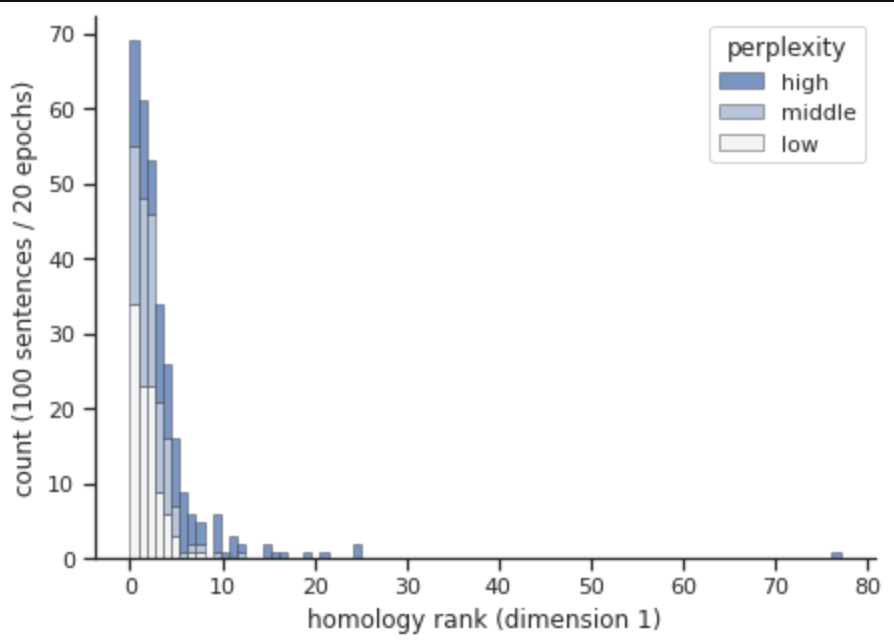}
	\includegraphics[width=0.22\textwidth]{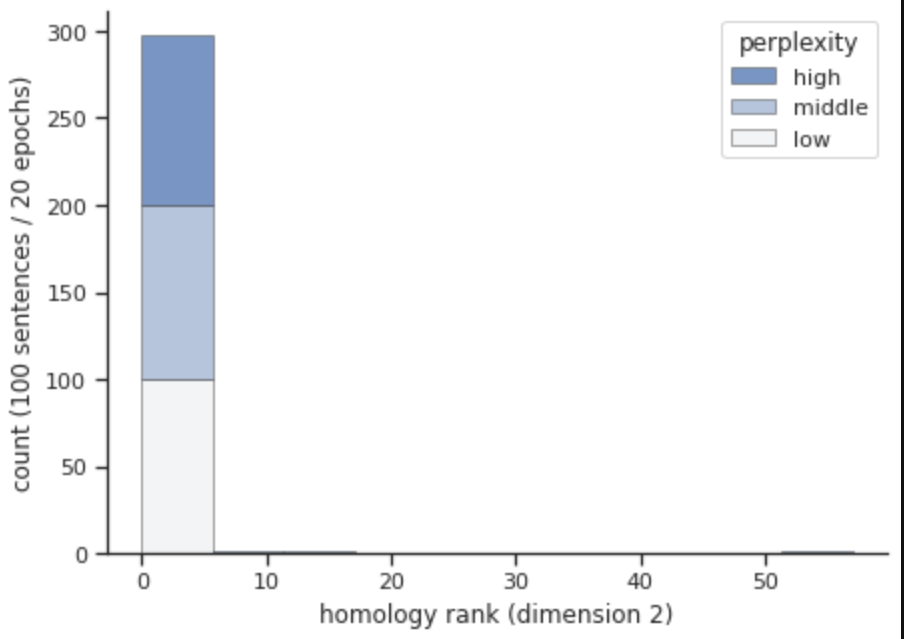}
  \caption{Visualization of the two most informative homology dimensions (loops and closed surfaces) in the representation manifold of an LSTM language model during training. As the model converges (low perplexity) we observe more topological complexity transferred to hidden states (top) and the corresponding reduction in topological complexity of input embeddings (bottom).}
	\label{betti_hist_epochs}
\end{figure}

Initially, we looked at the token embeddings (input level representations induced by the neural network).
These did not seem to have complex topological structure, and were rather ball like, with no cavities in higher dimensions.
We then inspected representations deeper into the neural network stack, comparing topology of input layer embeddings with the hidden states of the language model for the same sentences side by side.
After computing persistence diagrams of hidden layer representations, we observed a surprising pattern.
Although both the input (embedding) and hidden layer (cell state) representations of the sentences evolved during training - that is the vectors corresponding to tokens of each sentence moved around the ambient space - the former did not form cavities, while the latter did.
In other words, the topology of the embeddings remained nearly constant during that process, while the deeper representations (hidden states of the system) of the same sentences changed shape significantly, developing complex topological structures.
Figure \ref{betti_hist_epochs} shows the shift of homology rank distributions in two most informative dimensions (1 and 2) as corpus perplexity decreases during model convergence.

\begin{figure}
	\centering
  \includegraphics[width=0.22\textwidth]{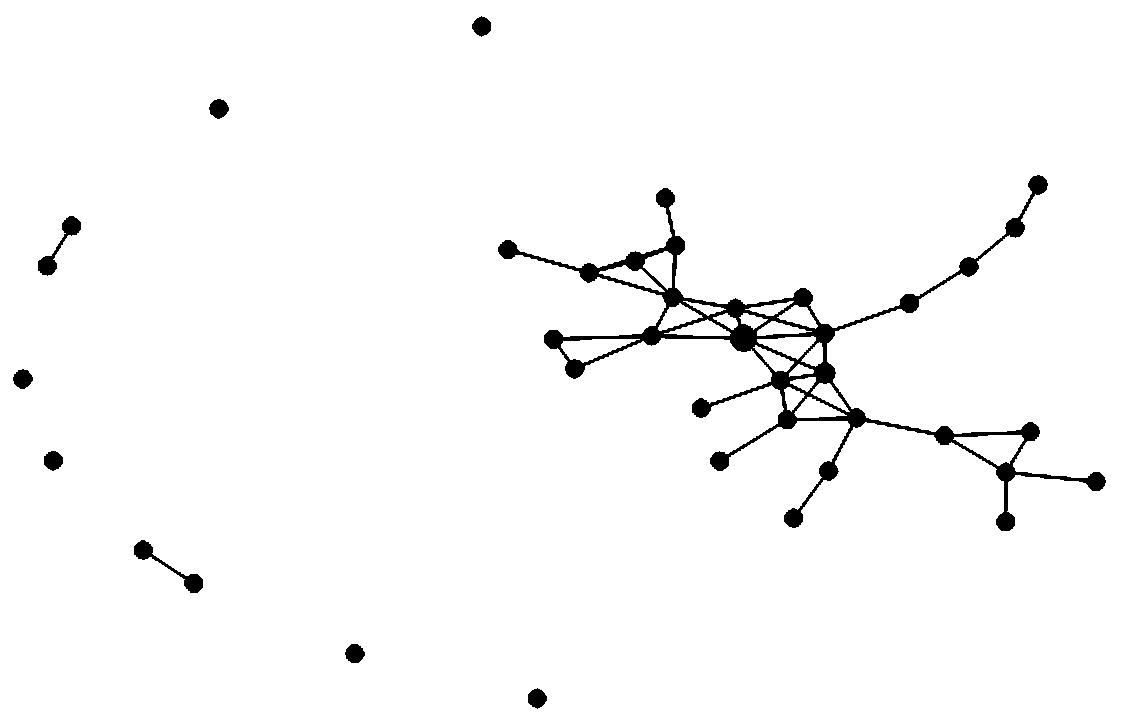}
	\includegraphics[width=0.22\textwidth]{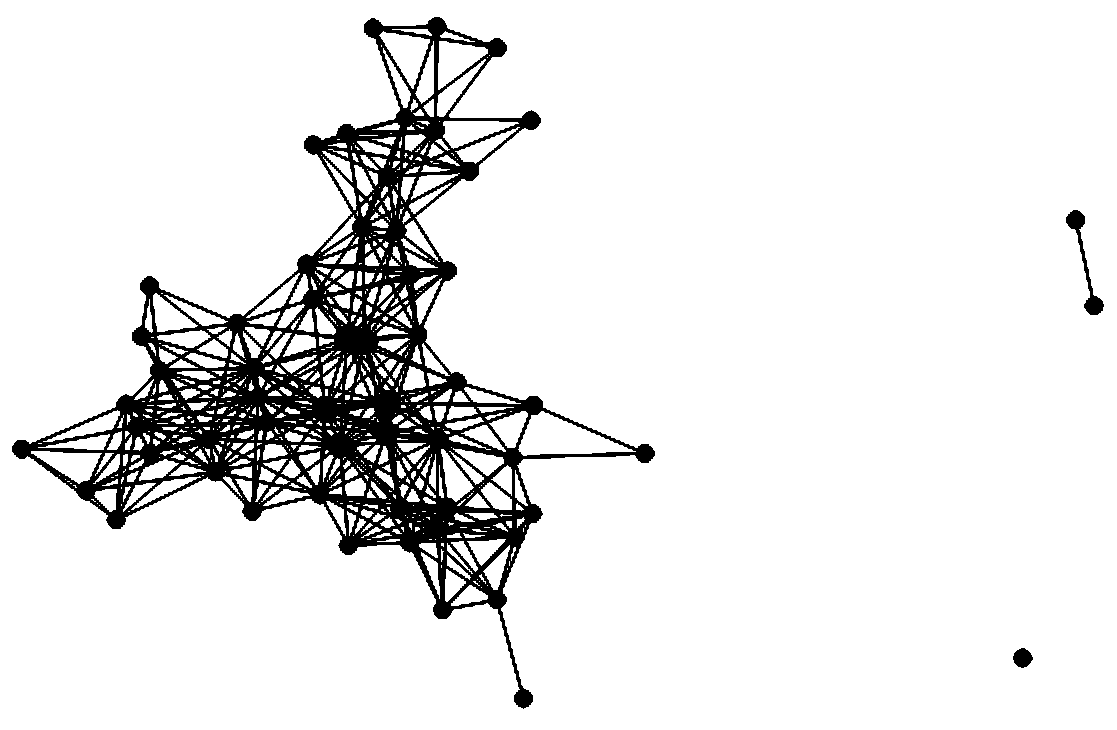}
  \includegraphics[width=0.22\textwidth]{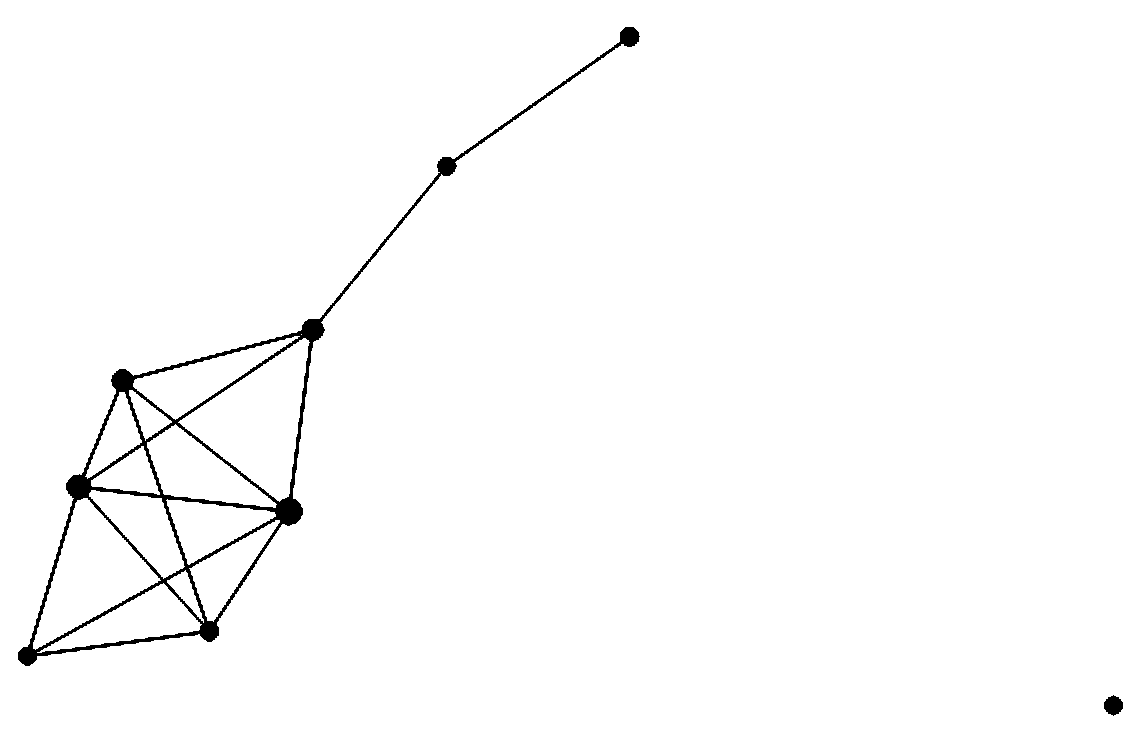}
	\includegraphics[width=0.22\textwidth]{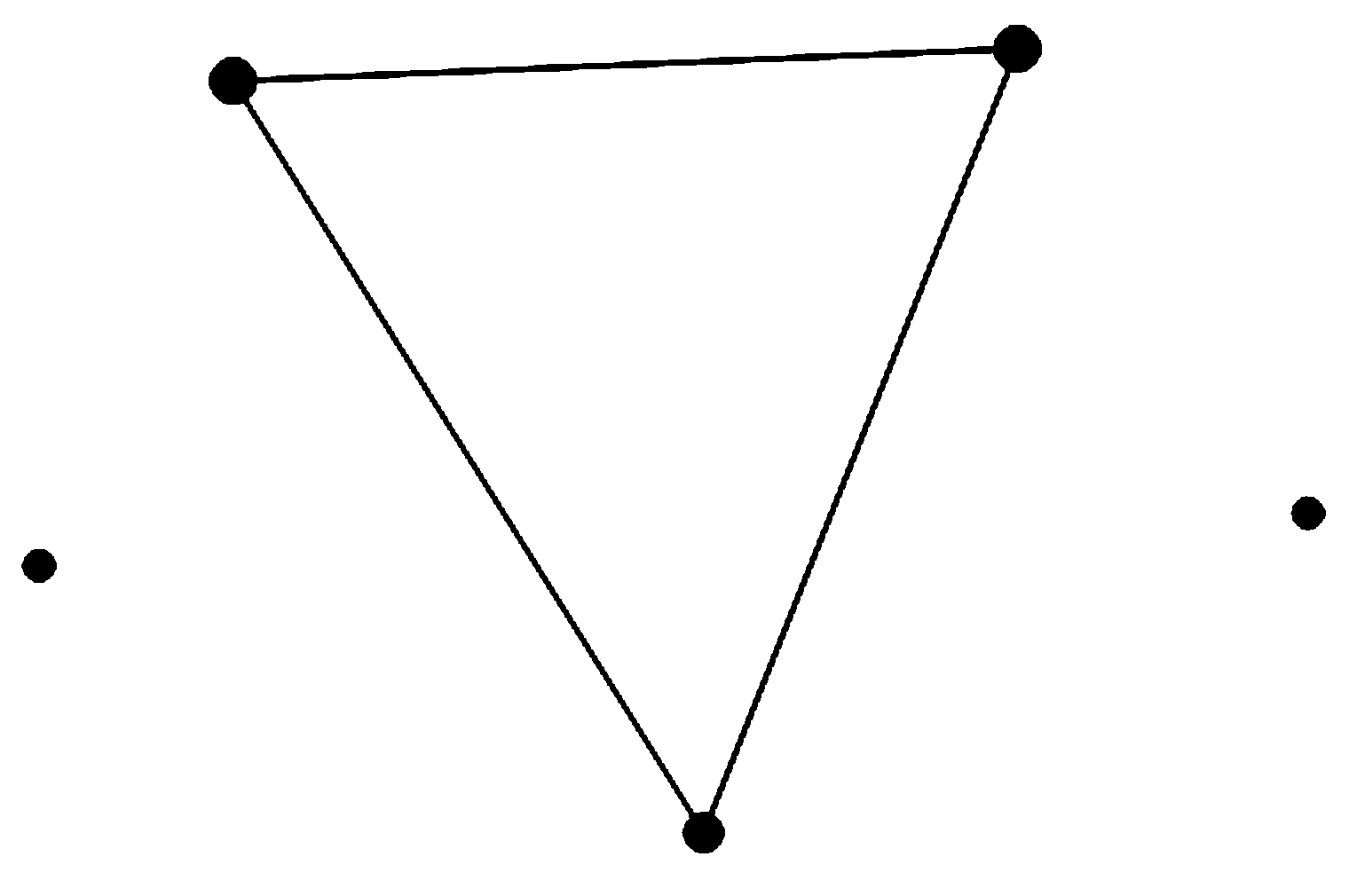}
  \caption{Graph approximations (simplicial mapper) to point clouds corresponding to tokens of a random text sample in English. Hidden states are on top and embeddings on the bottom.}
	\label{hid_emb_mapper}
\end{figure}

Encouraged by this discovery, we decided to survey the embedding manifold and the hidden representation manifold with the simplicial mapping technique (as in figure \ref{mapper}).
We produced graphs (i.e. simplicial mapping approximation of dimension 1) for individual sentences, and larger sections of the corpus in order to visualize the two spaces in a human readable format.
This resulted in visualizations exhibiting striking differences between the two manifolds, as viewed through the lens of simplicial mapping.
The top of figure \ref{hid_emb_mapper} shows a visualization resulting from simplicial mapping of hidden state vectors of our LSTM language model corresponding to a random sample from English.
We see that the graph is quite complex with multiple connected components, and intricate topological structure of edges.
The bottom of figure \ref{hid_emb_mapper} shows visualization of the same text in the embedding space.
We observe that the resulting graphs are significantly simpler.
Appendix C shows more visualizations of individual sentences, as well as approximations generated from entire documents.

\begin{figure}
	\centering
	\includegraphics[width=0.22\textwidth]{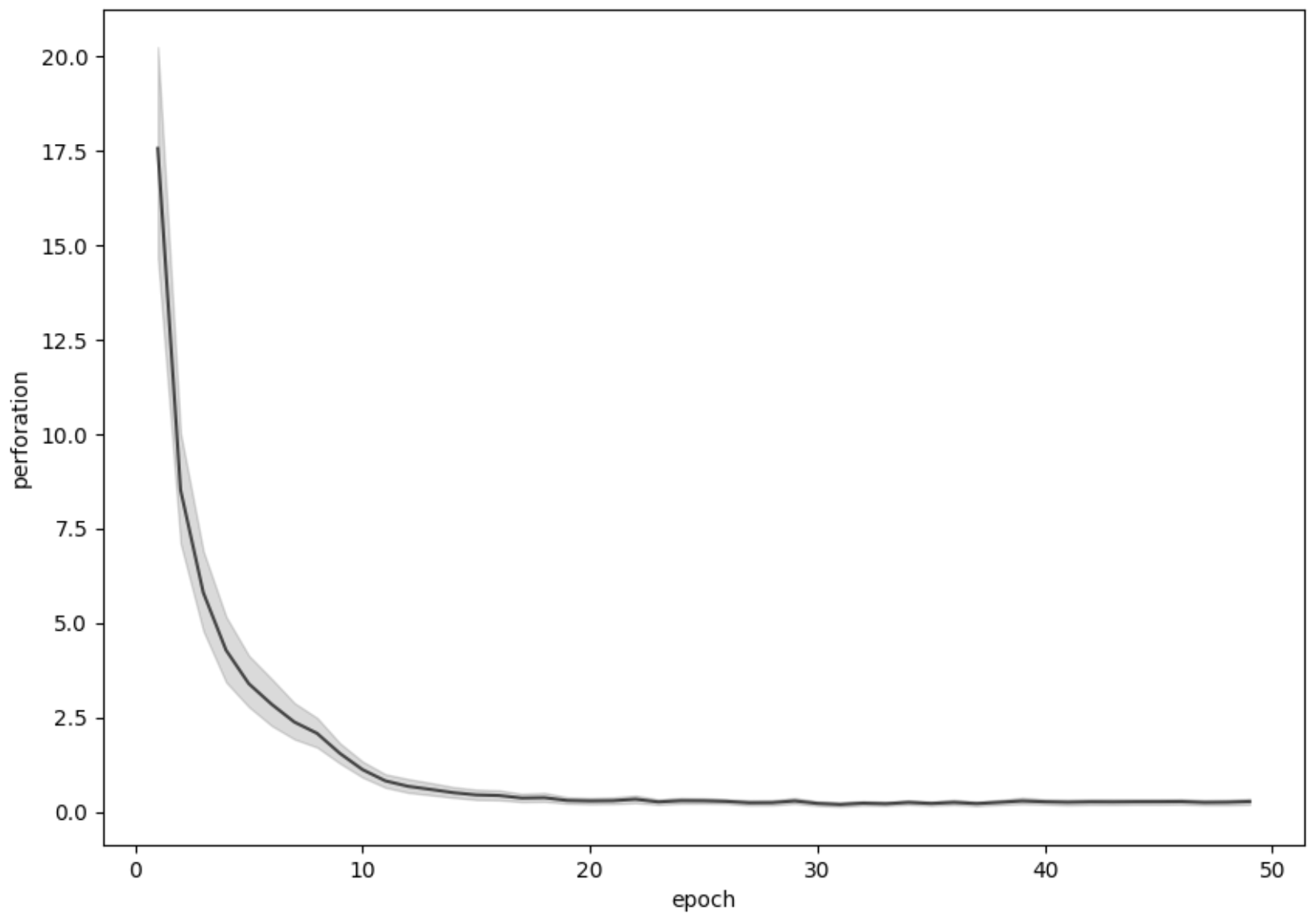}
	\includegraphics[width=0.22\textwidth]{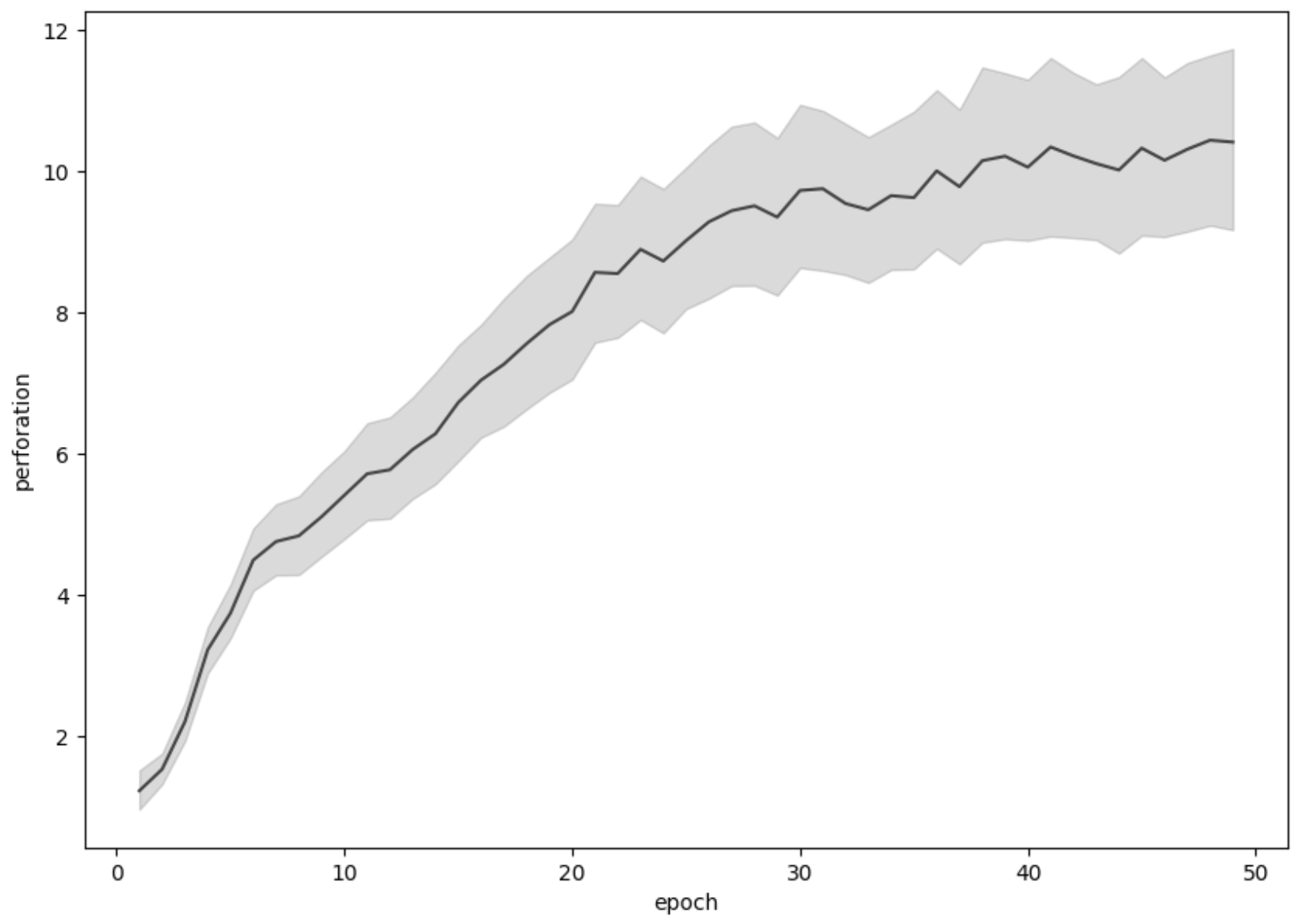}
	\includegraphics[width=0.22\textwidth]{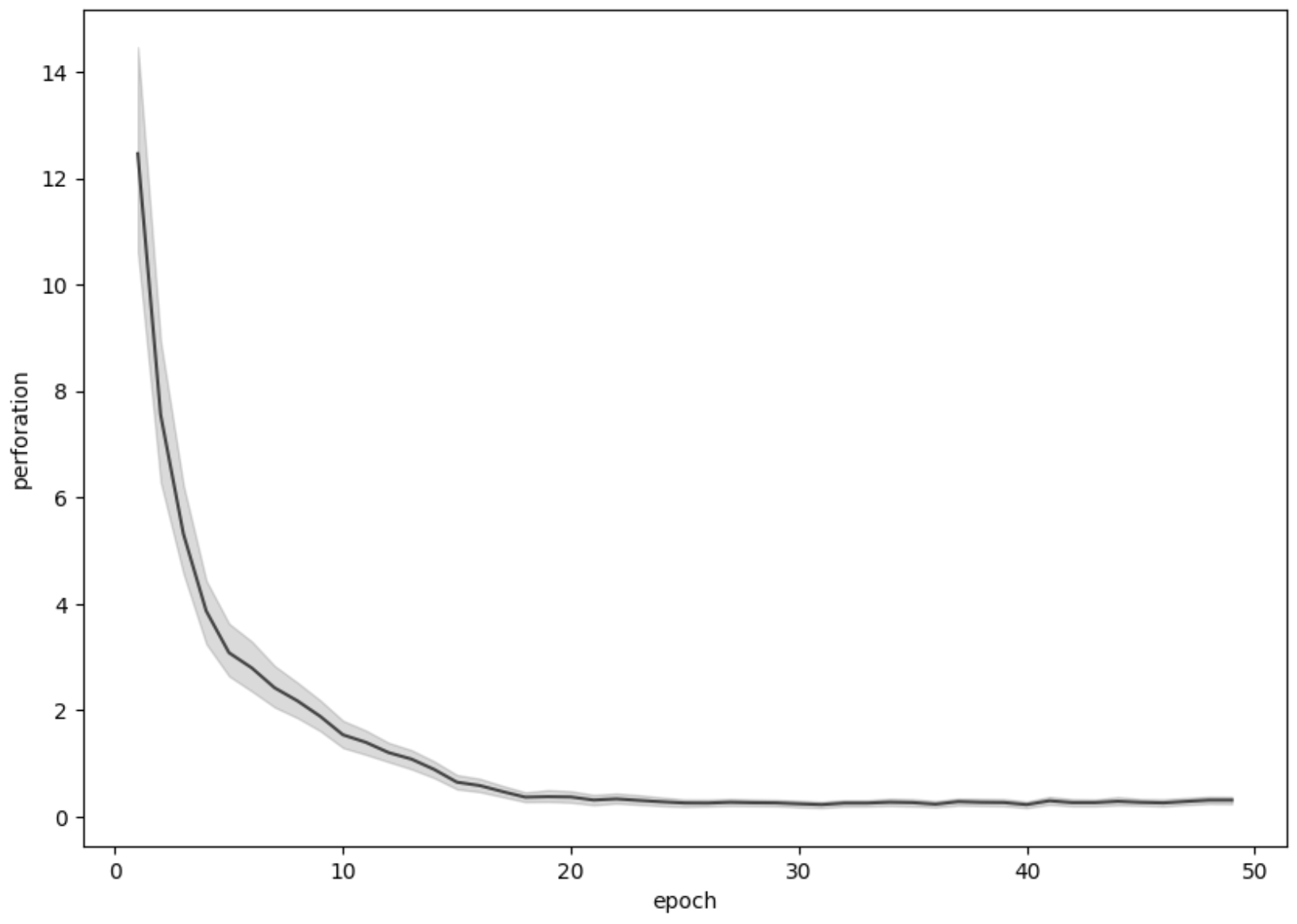}
	\includegraphics[width=0.22\textwidth]{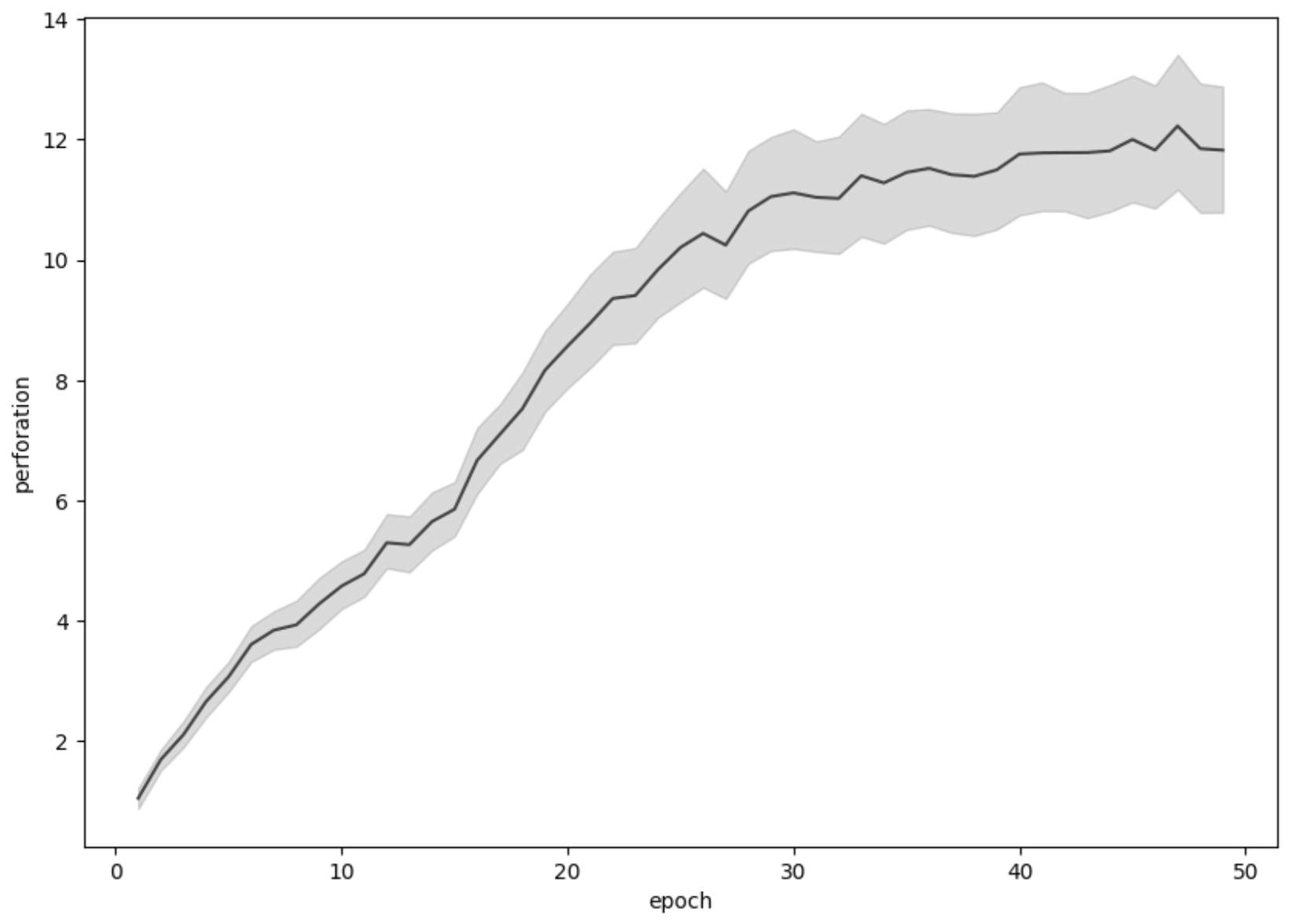}
	\includegraphics[width=0.22\textwidth]{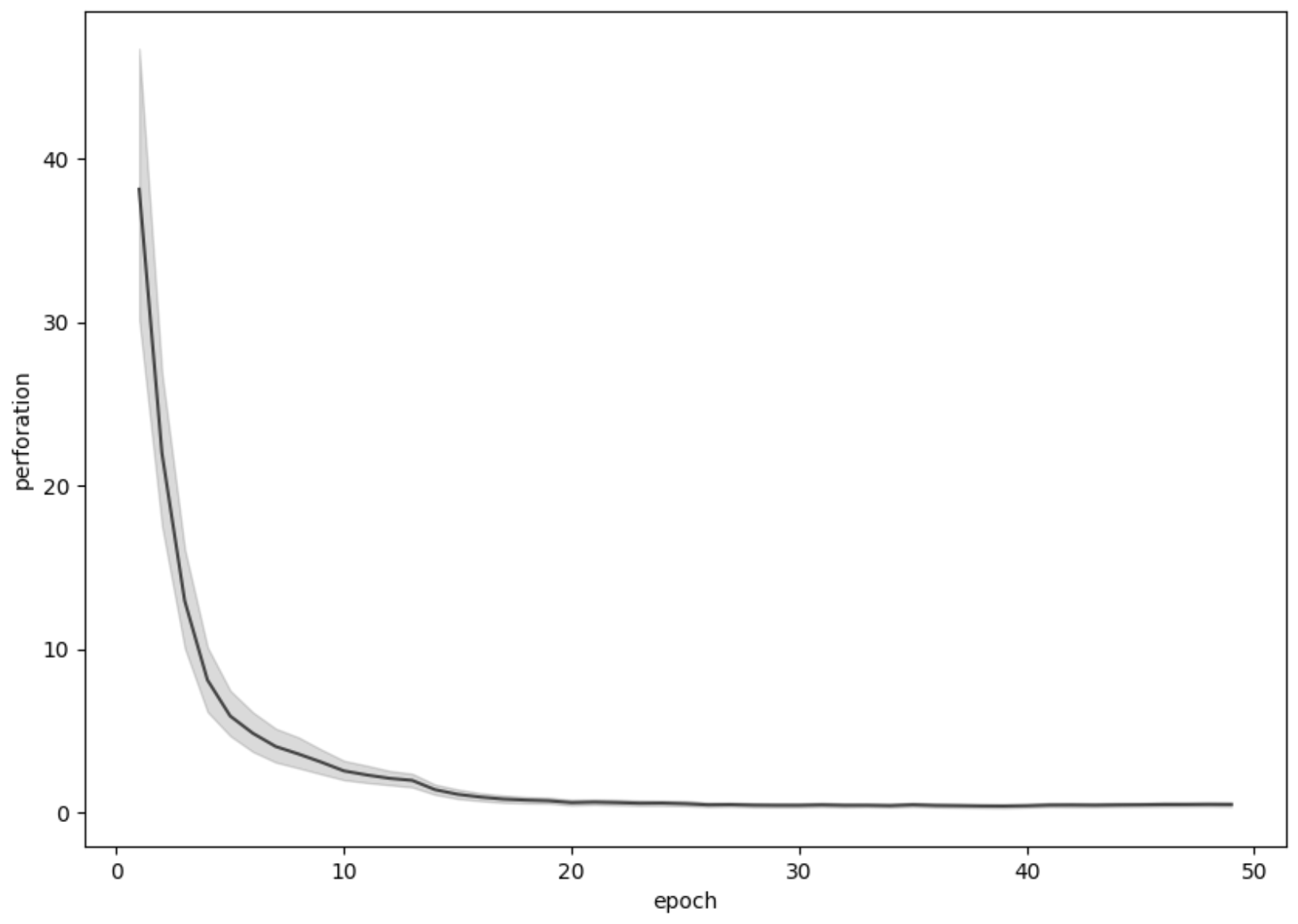}
	\includegraphics[width=0.22\textwidth]{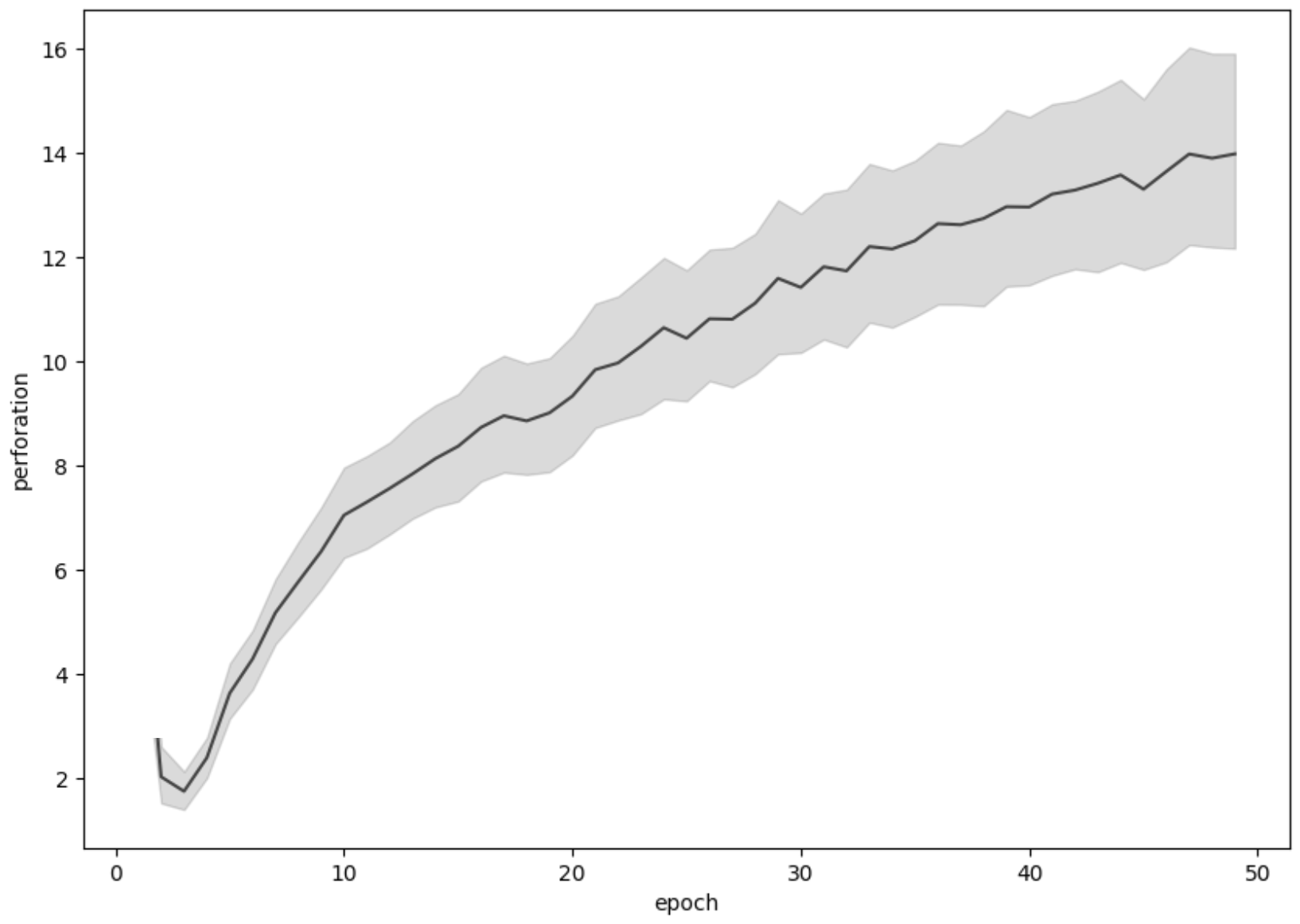}
  \caption{Perforation for embeddings (left) and hidden states (right) over epochs of training. Mean (center line) and 98\% interval shown (shaded area). The plots above correspond to the same language model trained from scratch on three equal size corpora of natural languages. From top to bottom: English, German, Japanese.}
	\label{perf50}
\end{figure}

This difference in topological complexity between embeddings and hidden states held across sentence sizes and across languages (we trained models on English, German, and Japanese).
In order to see the changes in complexity across depth and time during training, we sampled 2000 sentences at each epoch of training.
We then computed their persistent homology and used it to determine perforation (our one point summary of topological complexity) for each sentence.
We plotted perforation values for the input layer token embedding, and the hidden layer representations (mean over the sample and 98\% interval).
In the case of the LSTM model, we observed that perforation of deep contextualized representations increases with learning.
More precisely, during model training, there is an upwards trend in perforation over epochs in the hidden states of the language model as perplexity decreases.
This suggests that perforation can be used as a diagnostic tool for determining model convergence, and defining early stopping criteria for language models.
Input embeddings by comparison show no increase in perforation, and in fact decrease as the model converges showing a reversed relationship with perplexity.
These results suggests that the topological structure is transferred from the embedding space to the hidden states during training.
Furthermore, this topological relationship between global and contextualized word representations with respect to perplexity (higher epochs are lower perplexity) appears uniformly in all natural language corpora examined (figure \ref{perf50} shows results for input and hidden perforation for three languages).
All natural languages exhibited inverse relationship between input and hidden layer representations of sentences.
In particular, the input representations usually start with perforation in mid-teens and drop to near zero within the first 20 epochs, after which they remain nearly flat for the rest of the training process.
By contrast, the hidden state representations start near zero and increase to mid-teens during the same period, and stay nearly flat afterwards.
The drop in the perforation of input representations seems more sudden than the rise in hidden state representations, and some growth in topological complexity still happens in the hidden layers after a delay from when the input embeddings lose their initial perforation.

\begin{figure}
	\centering
	\includegraphics[width=0.45\textwidth]{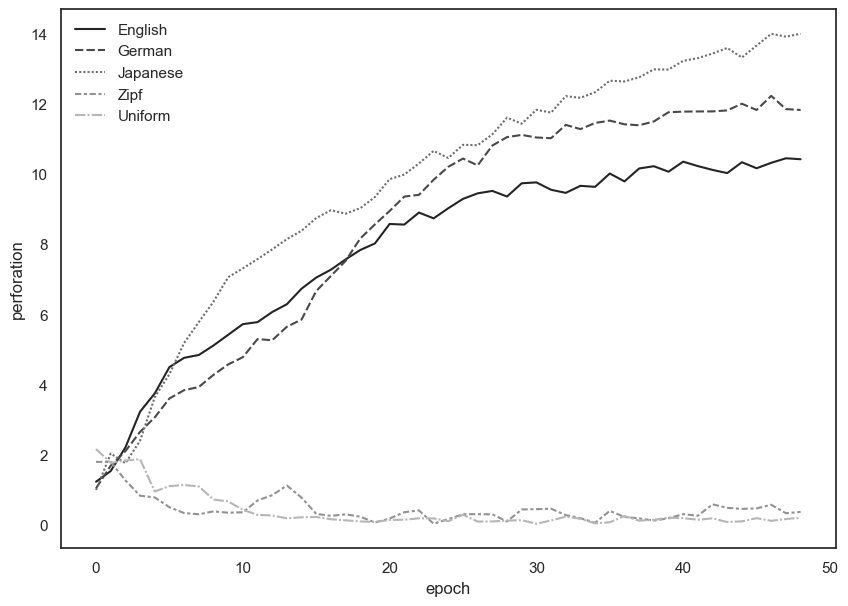}
	\caption{Combined perforation plot for natural and synthetic data. Hidden state evolution shown. Synthetic perforation oscillates near zero (two curves at the bottom) while natural perforation increases throughout language model training.}
	\label{combperf50}
\end{figure}

After discovering this relationship and establishing that it holds for all natural languages that we tested, we wanted to see if this effect is particular to natural language data, or is it a property of LSTM networks that would produce similar outcomes regardless of the corpus used.
If the phenomenon depended on the training data, we wanted to see if it can be reproduced with fake randomly generated data, which does not encode natural language structure.
For this purpose we generated two synthetic corpora: Zipf and Uniform.
Both of these contained the same number of sentences, with the same sentence length distributions, and same lexicon as the natural language corpora used before.
The difference was that they destroy natural language structure, generating meaningless sentences by sampling word tokens at random.
The Zipf corpus preserves the unigram frequency distribution found in natural corpora, while the Uniform corpus samples all tokens with equal probability, destroying all statistical properties of those languages.
Surprisingly, we discovered that none of the perforation phenomena common to natural languages occur in language models trained on those synthetic corpora.
This implies that perforation can be used as a basis for a natural language detector.
Here, the relationship between perforation changes during training for input and hidden state preresentations that holds for natural languages is no longer present.
The input embeddings exhibit trivial homological structure with perforation remaining at zero (plus negligible topological noise).
The hidden state perforation is also near zero, dropping slightly from a low value around 2 during initial epochs (while the natural language perforation always goes up significantly).
Figure \ref{combperf50} shows the hidden state plots of perforation values over 50 epochs of language model training.
Here the difference between natural and synthetic data is immediately visible.

\begin{figure}
	\centering
	\includegraphics[width=0.22\textwidth]{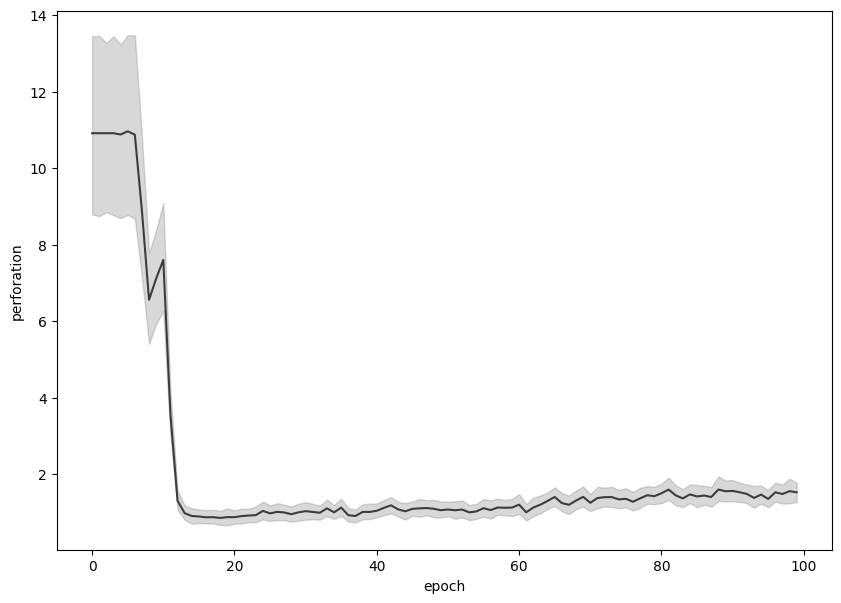}
	\includegraphics[width=0.22\textwidth]{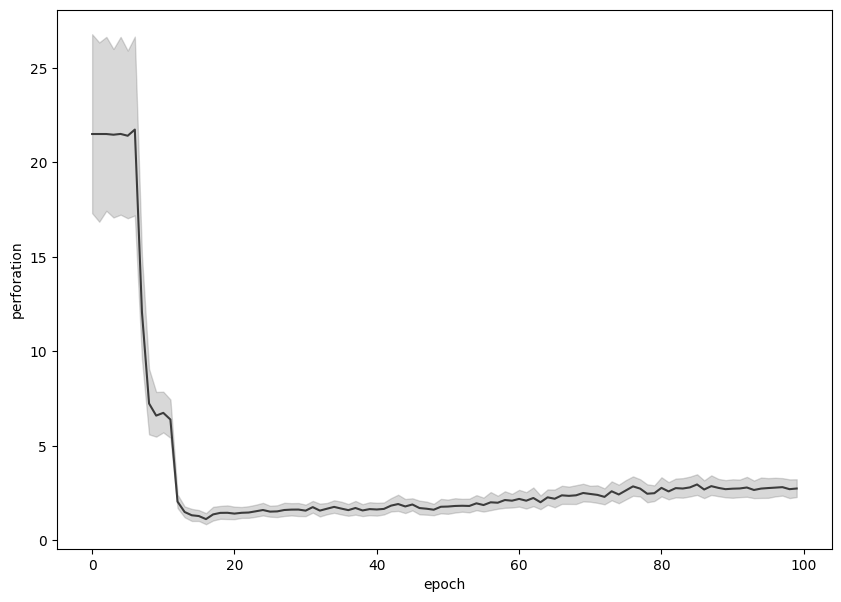}
	\includegraphics[width=0.22\textwidth]{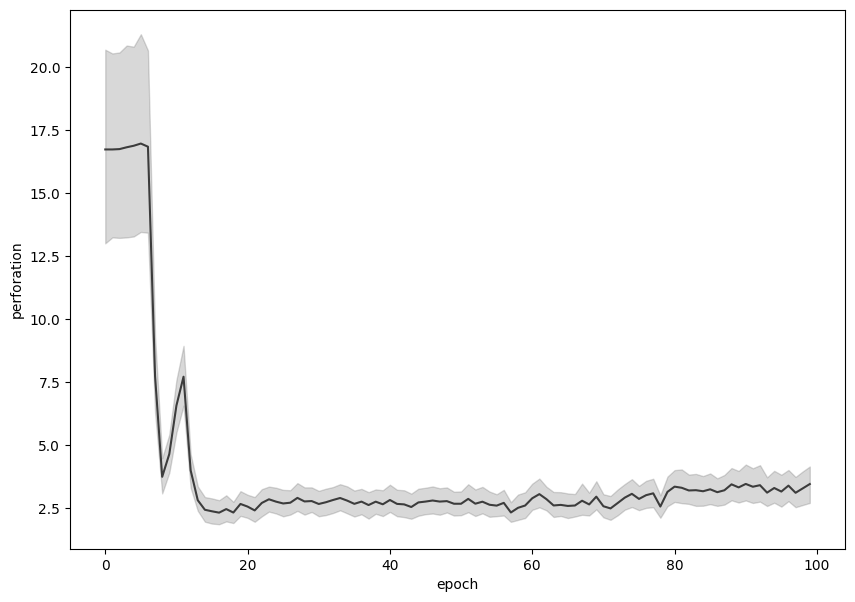}
	\includegraphics[width=0.22\textwidth]{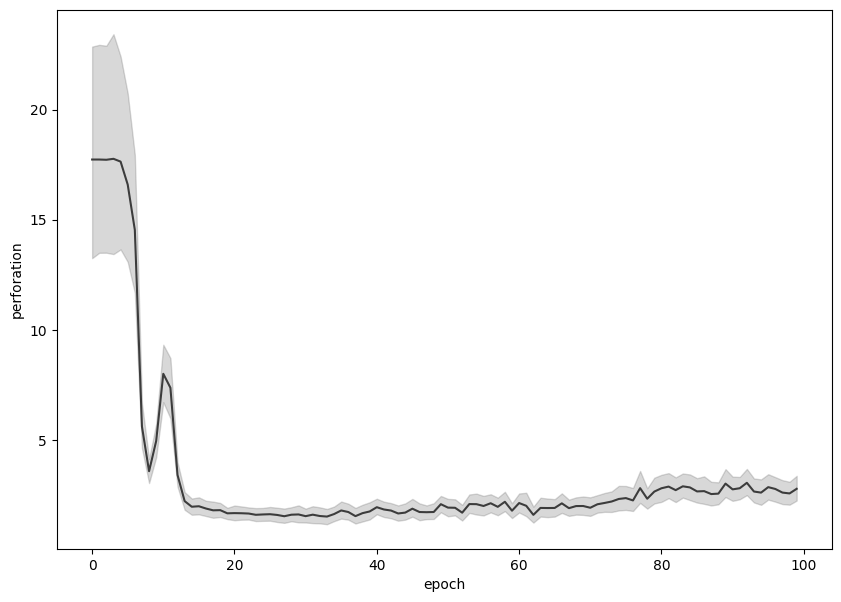}
	\includegraphics[width=0.22\textwidth]{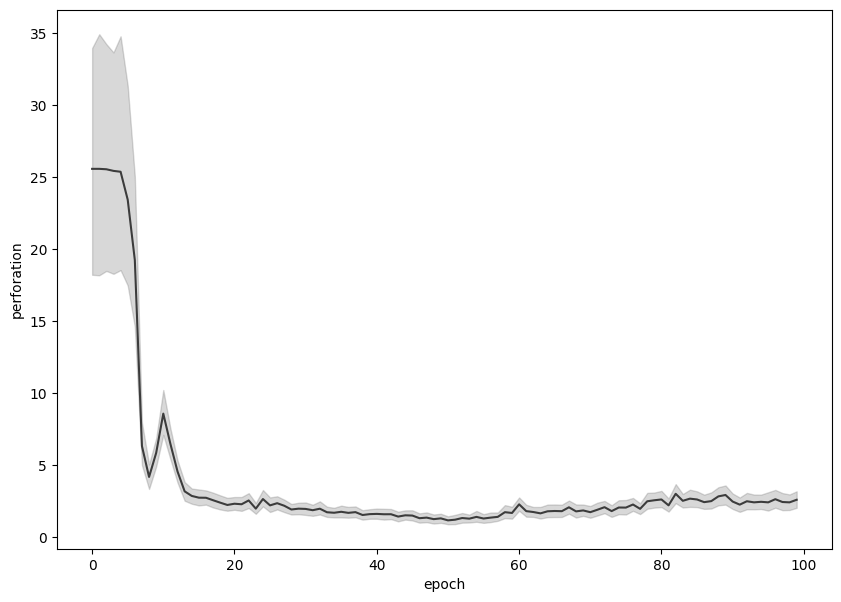}
	\includegraphics[width=0.22\textwidth]{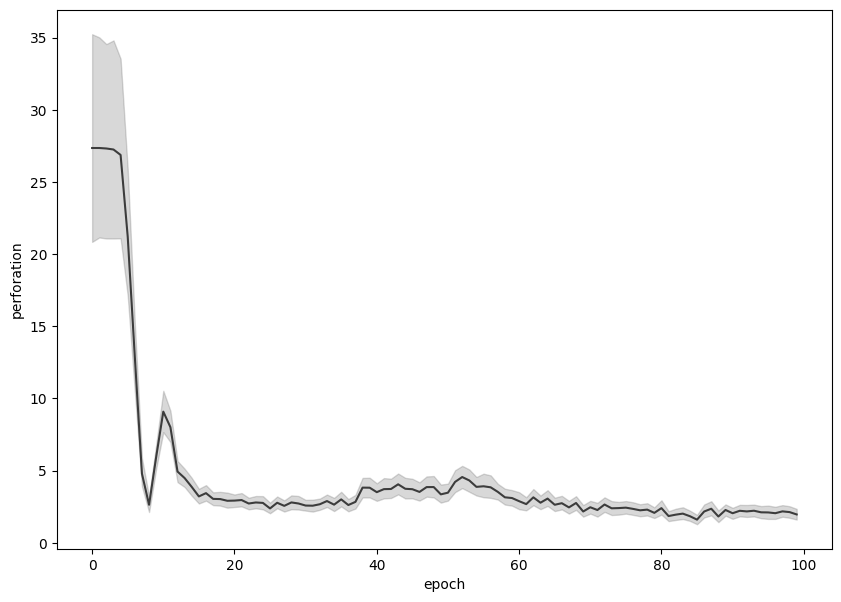}
  \caption{Perforation over epochs of training. All six transformer blocks (layer 1 is top left, layer 6 is bottom right) of a 70m GPT model trained on 800GB of diverse text.}
	\label{pythia_perf}
\end{figure}

Having established this pattern of increased topological complexity in the hidden layer of LSTM language models, we wanted to see if it holds for the transformer architecture.
For this purpose we used Pythia - a suite for analyzing large language models across training and scale \cite{biderman2023pythia}.
These are GPT based, decoder only, autoregressive transformer models ranging from 70m to 12B parameters, trained on the Pile dataset \cite{gao2020pile}.
Figure \ref{pythia_perf} shows the perforation plots for the transformer architecture.

\begin{figure}
	\centering
	\includegraphics[width=0.22\textwidth]{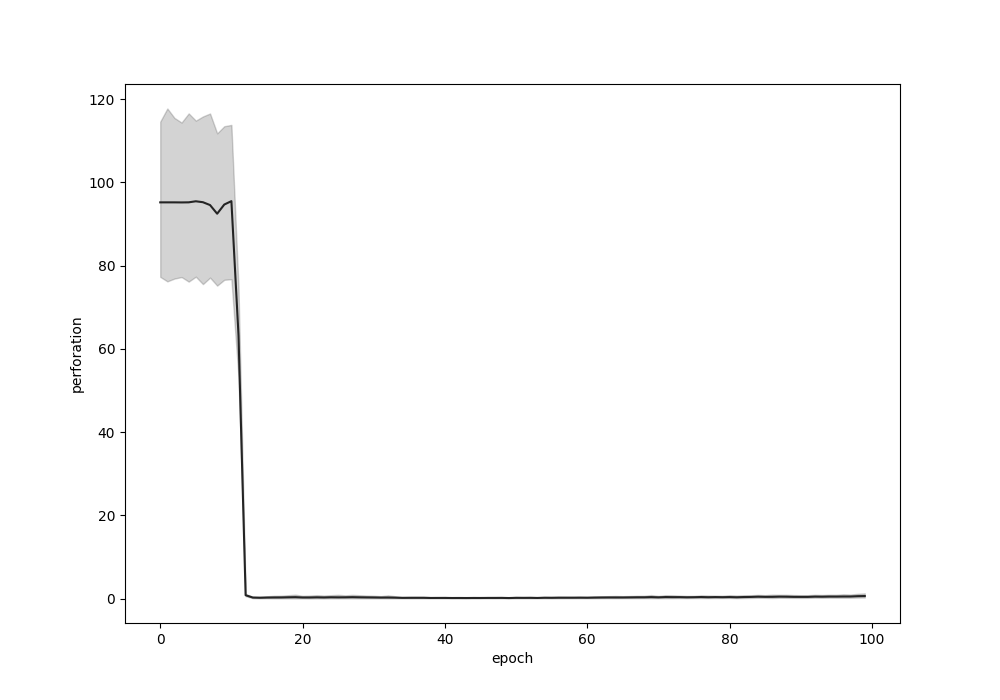}
	\includegraphics[width=0.22\textwidth]{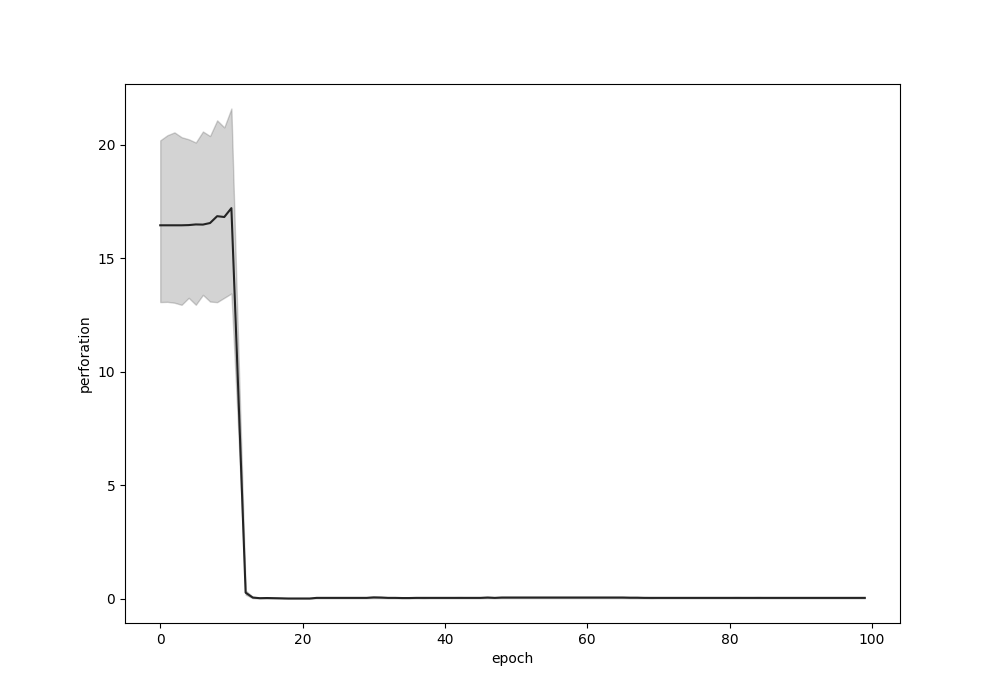}
	\includegraphics[width=0.22\textwidth]{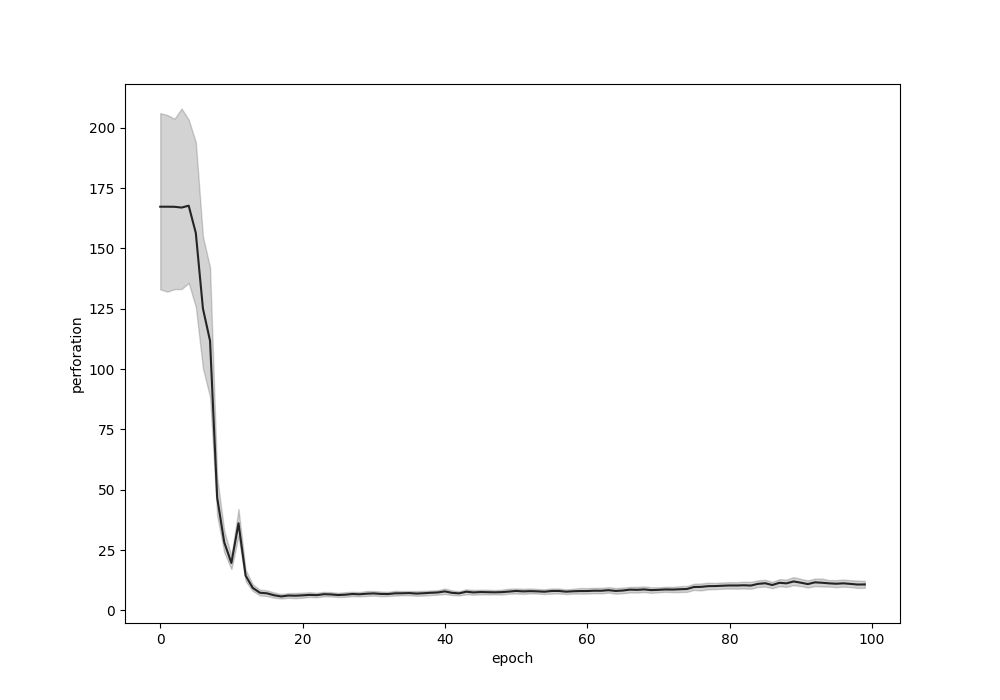}
	\includegraphics[width=0.22\textwidth]{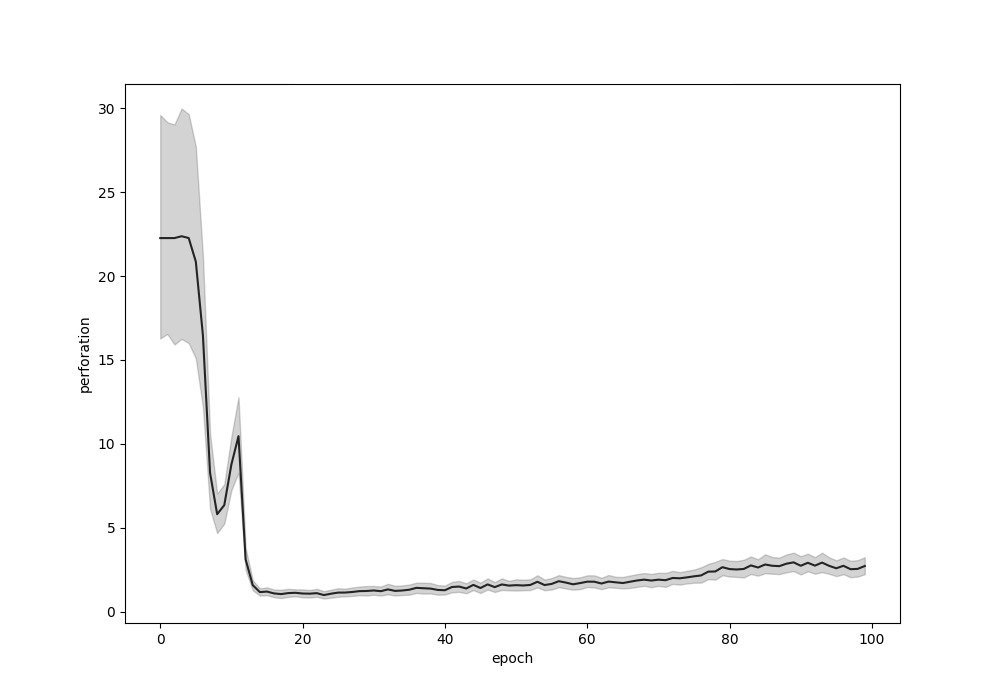}
	\includegraphics[width=0.22\textwidth]{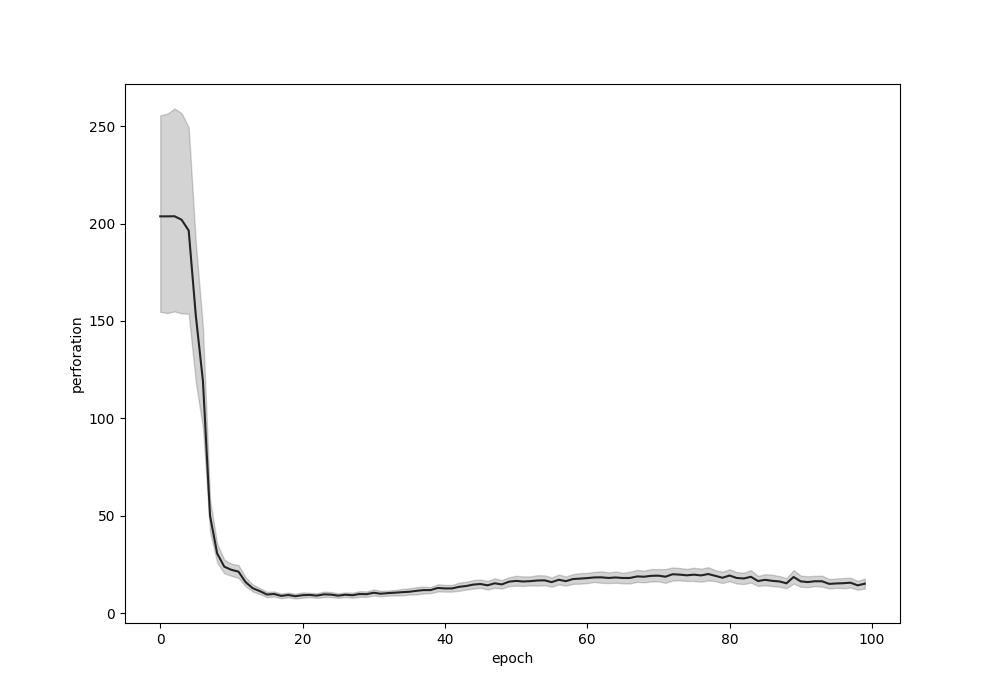}
	\includegraphics[width=0.22\textwidth]{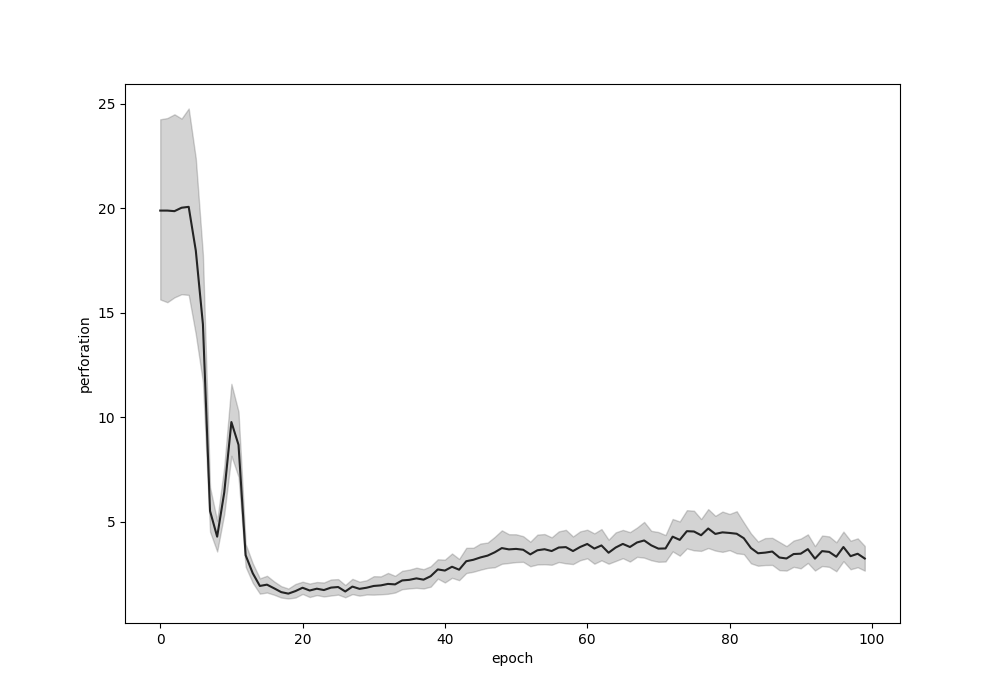}
	\includegraphics[width=0.22\textwidth]{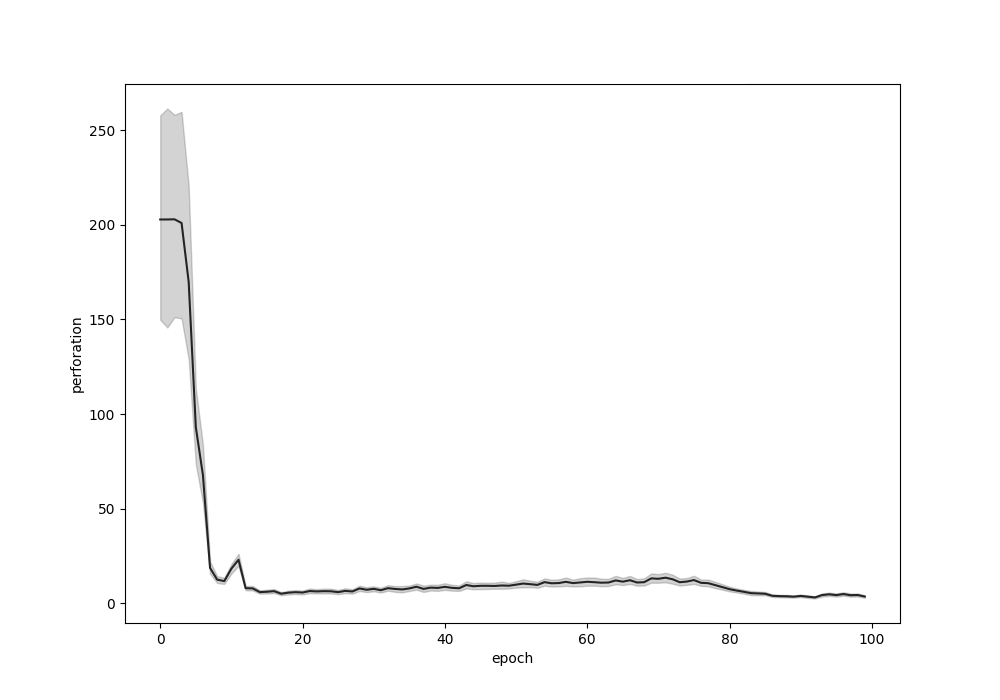}
	\includegraphics[width=0.22\textwidth]{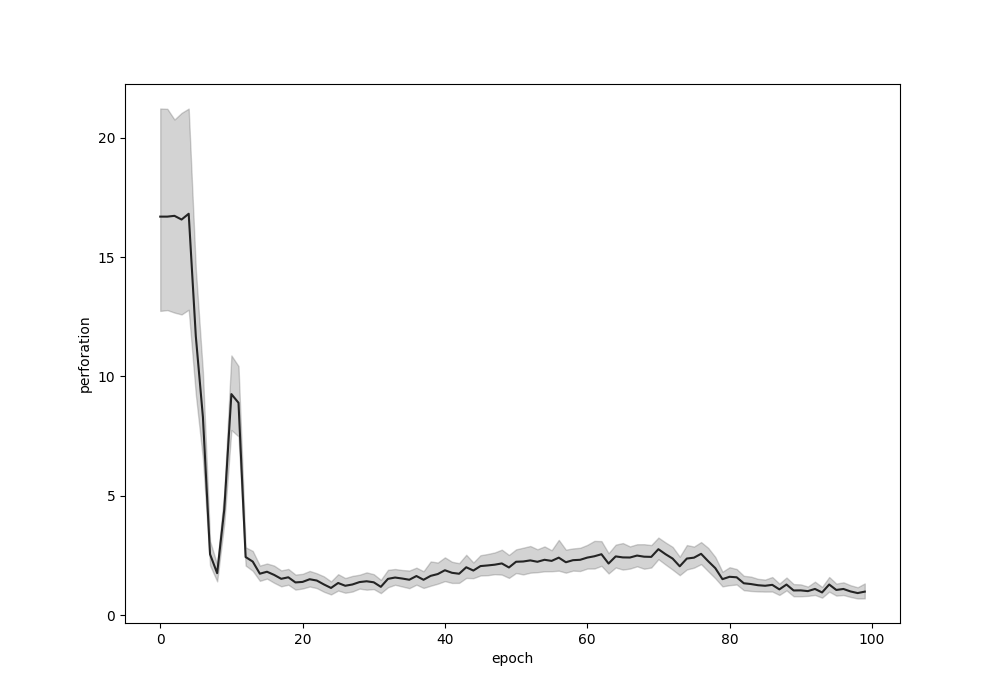}
  \caption{Perforation over epochs of training. The embedding layer and transformer blocks 4, 8, 12 of a 160m GPT model (Pythia) in order top to bottom (top row is the embedding and bottom is layer 12). The left column shows results for the "uniform corpus" obtained by sampling words with equal probability. The right column corresponds to the "Zipf corpus" which preserves unigram distribution of the natural corpora.}
	\label{pythia_perf_random}
\end{figure}

In contrast to the recurrent cell, we see initial sharp drop in perforation which then remains approximately constant around a low value below 2.
These plots look more like the token embedding layer of the LSTM model, although they are the deep representations computed by the hidden layers.
In the recurrent case, the deep representations exhibited a clear rise in perforation.
This means that the hidden state manifold of the LSTM cell develops complex topological structure during training, while the manifolds induced by the transformer layers do not.

Figure \ref{pythia_perf_random} shows perforation plots for the larger 160m parameter Pythia model on two synthetic corpora.
These corpora preserve the sentence length distribution of the English corpus, but are composed of meaningless sentences produced from two unigram distributions: uniform and Zipf.
We see that the plots are similar to GPT plots for natural English corpus in figure \ref{pythia_perf}, except that the uniform plots start much higher.
The high initial perforation values for the uniform corpus are expected because in this corpus every token has an approximately equal chance of appearing in the context of any other tokens, which means many random connections will be formed when a limited sample is chosen within a sentence.
This leads to random graphs with many loops.
However, all perforation vanishes as the model is trained, same as for the natural corpora.
This shows that transformer models do not exhibit the topological structure that allowed recurrent models to distinguish between natural and synthetic data purely based on the homology of their representation manifolds.

\begin{figure}
	\centering
	\includegraphics[width=0.22\textwidth]{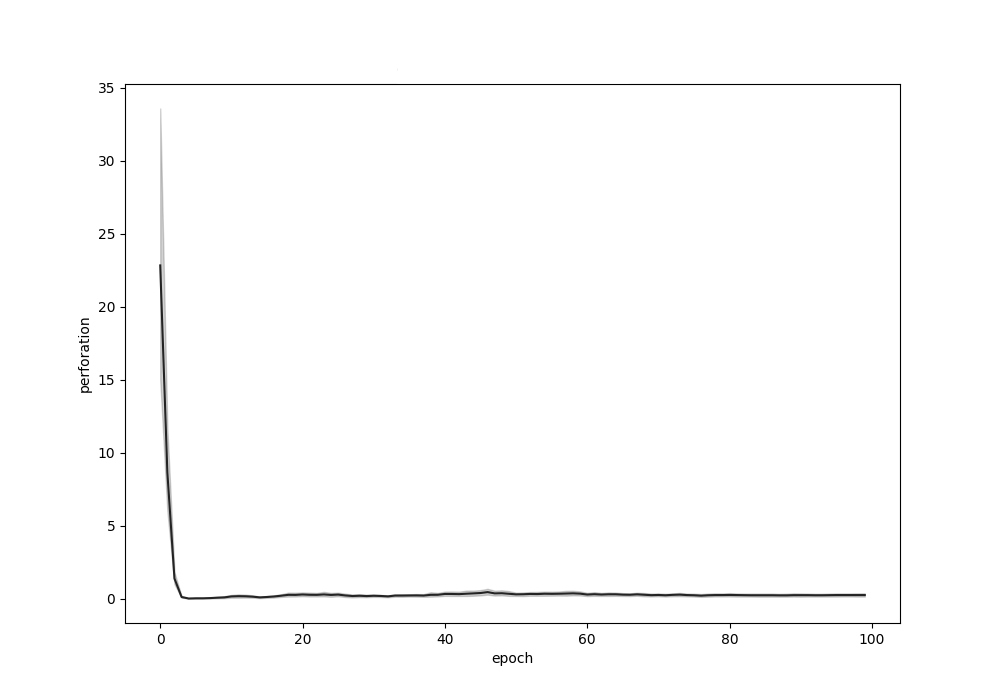}
	\includegraphics[width=0.22\textwidth]{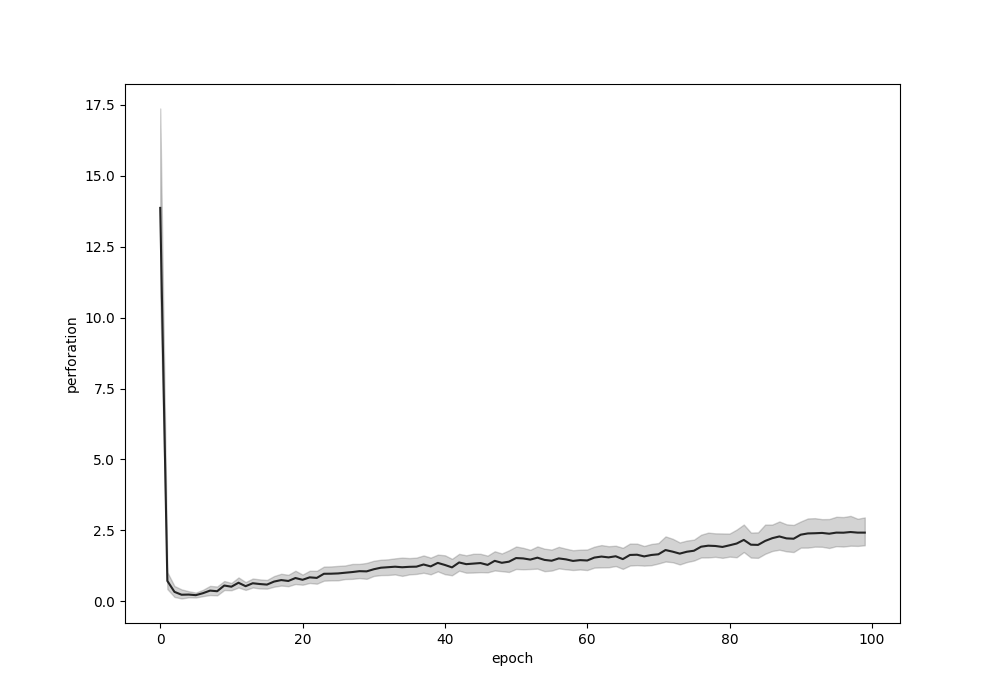}
	\includegraphics[width=0.22\textwidth]{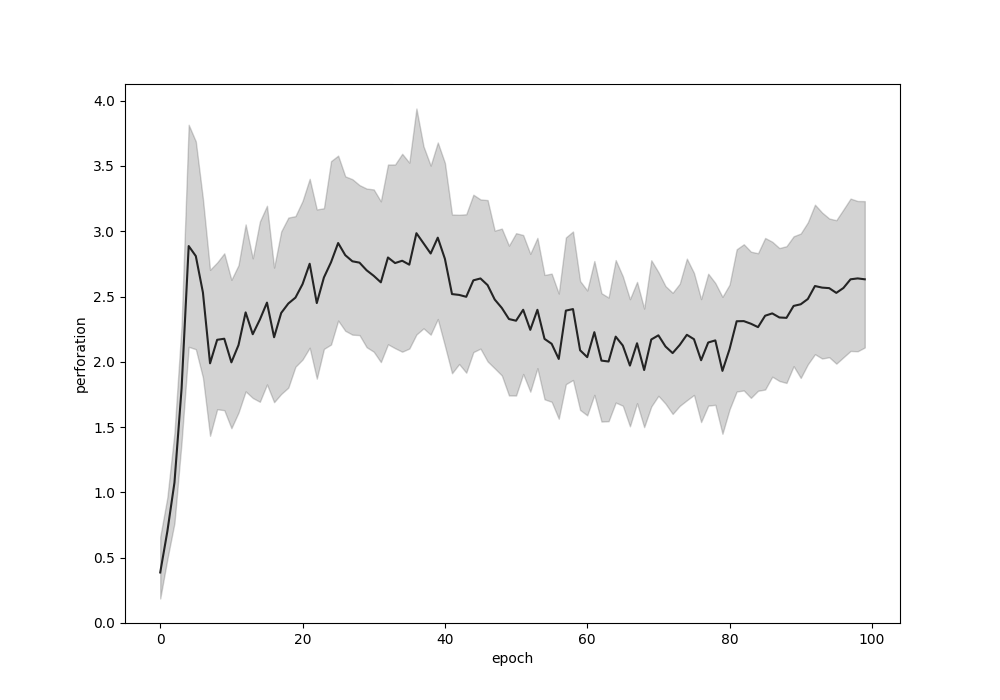}
	\includegraphics[width=0.22\textwidth]{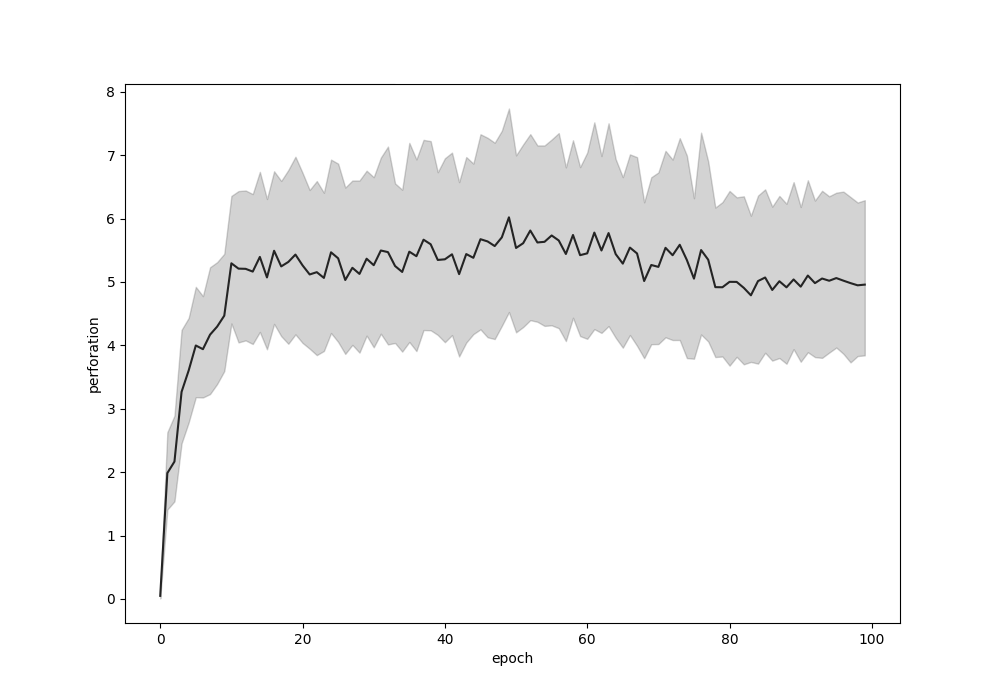}
  \caption{Perforation over epochs of training for a convolutional language model with 4 CNN layers. The top row shows the first and last layers (from left to right) and the bottom row shows hidden layers (second and third from left to right). These plots are for Greek, but similar pattern was exhibited by LMs trained on all natural corpora. See the appendix for additional plots from other languages.}
	\label{pythia_cnn}
\end{figure}

We wanted to test whether the vanishing of perforation in GPT models had something to do with the fact that transformer layers are fully connected.
For that purpose, we implemented a third language model architecture fully based on convolutional neural networks and trained it on a variety of source languages.
The CNN language model is not fully connected, while also not involving recurrence, so it is a good third option allowing us to ablate those two factors.
Figure \ref{pythia_cnn} shows results for one of the languages (additional plots are given in the appendix).
We observed transformer like vanishing of perforation at the input and output layers.
However, the hidden middle layers of the CNN LM showed a consistent pattern of slight grow in topological complexity that stabilizes in mean within the range of 3-6 and shows increased variance.
This is interesting because the inner layers of the CNN model show some similarity in topological changes during training to the hidden state of the LSTM, while the input and output layers look similar to GPT.
The input layer is the token embedding, while the output layer is a linear projection and softmax.
The hidden layers have nontrivial connection structure, because they are the convolutions, which introduce sparsity.
In case of the LSTM, the nontrivial topology in terms of computation graph was introduced by the hidden state bottleneck.
Because of these observations we believe that ther relative lack of topological complexity in GPT models is caused by the fully connected nature of tranformer blocks.

\section{Conclusion}

We presented a novel method of analyzing the topological structure of representation manifolds induced by neural language models.
Our method produces human readable visualizations of the changes in high dimensional topological structure of holes in the hidden layers of neural network models.

Using our method, we showed that the hidden state representations of LSTM language models exhibit complex topological structure, while the input embeddings do not.
Furthermore, we showed that this pattern is consistent across different natural languages, and is not present in synthetic data.
Surprisingly, transformers do not exhibit this pattern, and their hidden state representations remain topologically simple throughout training.
In order to suggest a possible reason for this lack of "holes" in transformer manifolds, we implement convolutional language models, and show that sparsity of the convolutional layers introduces holes into their emergent representations.
This suggests that fully connected computational graphs, such as those in the GPT models, induce solid ball representation manifolds.
In contrast, nontrivial topologies of the recurrent and convolutional architectures lead to complex topologies in the representations such models induce.

In order to explain this further we suggest two paths.

The empirical path would involve evolving (e.g. using algorithms such as NeuroEvolution of Augmenting Topologies \cite{stanley2002evolving}) nontrivial neural network topologies with the goal of increasing perforation.

The theoretical path would be proving theorems relating the topology of the computation graph to the topology of the corresponding representation manifolds.



\bibliography{main}

\begin{thebibliography}{23}
\providecommand{\natexlab}[1]{#1}
\providecommand{\url}[1]{\texttt{#1}}
\expandafter\ifx\csname urlstyle\endcsname\relax
  \providecommand{\doi}[1]{doi: #1}\else
  \providecommand{\doi}{doi: \begingroup \urlstyle{rm}\Url}\fi

\bibitem[Akiba et~al.(2019)Akiba, Sano, Yanase, Ohta, and
  Koyama]{akiba2019optuna}
Akiba, T., Sano, S., Yanase, T., Ohta, T., and Koyama, M.
\newblock Optuna: A next-generation hyperparameter optimization framework.
\newblock In \emph{Proceedings of the 25th ACM SIGKDD international conference
  on knowledge discovery \& data mining}, pp.\  2623--2631, 2019.

\bibitem[Biderman et~al.(2023)Biderman, Schoelkopf, Anthony, Bradley,
  O’Brien, Hallahan, Khan, Purohit, Prashanth, Raff,
  et~al.]{biderman2023pythia}
Biderman, S., Schoelkopf, H., Anthony, Q.~G., Bradley, H., O’Brien, K.,
  Hallahan, E., Khan, M.~A., Purohit, S., Prashanth, U.~S., Raff, E., et~al.
\newblock Pythia: A suite for analyzing large language models across training
  and scaling.
\newblock In \emph{International Conference on Machine Learning}, pp.\
  2397--2430. PMLR, 2023.

\bibitem[Bulatov et~al.(2023)Bulatov, Kuratov, and Burtsev]{bulatov2023scaling}
Bulatov, A., Kuratov, Y., and Burtsev, M.~S.
\newblock Scaling transformer to 1m tokens and beyond with rmt.
\newblock \emph{arXiv preprint arXiv:2304.11062}, 2023.

\bibitem[Carlsson et~al.(2008)Carlsson, Ishkhanov, De~Silva, and
  Zomorodian]{carlsson2008local}
Carlsson, G., Ishkhanov, T., De~Silva, V., and Zomorodian, A.
\newblock On the local behavior of spaces of natural images.
\newblock \emph{International journal of computer vision}, 2008.

\bibitem[Dauphin et~al.(2017)Dauphin, Fan, Auli, and
  Grangier]{dauphin2017language}
Dauphin, Y.~N., Fan, A., Auli, M., and Grangier, D.
\newblock Language modeling with gated convolutional networks.
\newblock In \emph{International conference on machine learning}, pp.\
  933--941. PMLR, 2017.

\bibitem[Edelsbrunner et~al.(2008)Edelsbrunner, Harer,
  et~al.]{edelsbrunner2008persistent}
Edelsbrunner, H., Harer, J., et~al.
\newblock Persistent homology-a survey.
\newblock \emph{Contemporary mathematics}, 453\penalty0 (26):\penalty0
  257--282, 2008.

\bibitem[Fitz(2022)]{fitz2022shape}
Fitz, S.
\newblock The shape of words-topological structure in natural language data.
\newblock PMLR, 2022.

\bibitem[Franco(2016)]{franco2016amazing}
Franco, M.
\newblock Amazing star wars shadow art will awaken your wonder, 2016.
\newblock URL \url{http://tinyurl.com/43th5jre}.

\bibitem[Fuchs et~al.(2020)Fuchs, Worrall, Fischer, and Welling]{fuchs2020se}
Fuchs, F.~B., Worrall, D.~E., Fischer, V., and Welling, M.
\newblock Se (3)-transformers: 3d roto-translation equivariant attention
  networks.
\newblock \emph{arXiv preprint arXiv:2006.10503}, 2020.

\bibitem[Gao et~al.(2020)Gao, Biderman, Black, Golding, Hoppe, Foster, Phang,
  He, Thite, Nabeshima, et~al.]{gao2020pile}
Gao, L., Biderman, S., Black, S., Golding, L., Hoppe, T., Foster, C., Phang,
  J., He, H., Thite, A., Nabeshima, N., et~al.
\newblock The pile: An 800gb dataset of diverse text for language modeling.
\newblock \emph{arXiv preprint arXiv:2101.00027}, 2020.

\bibitem[Ghrist(2008)]{ghrist2008barcodes}
Ghrist, R.
\newblock Barcodes: the persistent topology of data.
\newblock \emph{Bulletin of the American Mathematical Society}, 45\penalty0
  (1):\penalty0 61--75, 2008.

\bibitem[Gu \& Dao(2023)Gu and Dao]{gu2023mamba}
Gu, A. and Dao, T.
\newblock Mamba: Linear-time sequence modeling with selective state spaces.
\newblock \emph{arXiv preprint arXiv:2312.00752}, 2023.

\bibitem[Hatcher(2001)]{hatcher2001algebraic}
Hatcher, A.
\newblock \emph{Algebraic topology}.
\newblock Cambridge University Press, 2001.

\bibitem[Hinton et~al.(2011)Hinton, Krizhevsky, and
  Wang]{hinton2011transforming}
Hinton, G.~E., Krizhevsky, A., and Wang, S.~D.
\newblock Transforming auto-encoders.
\newblock In \emph{International conference on artificial neural networks},
  pp.\  44--51. Springer, 2011.

\bibitem[Jumper et~al.(2021)Jumper, Evans, Pritzel, Green, Figurnov,
  Ronneberger, Tunyasuvunakool, Bates, {\v{Z}}{\'\i}dek, Potapenko,
  et~al.]{jumper2021highly}
Jumper, J., Evans, R., Pritzel, A., Green, T., Figurnov, M., Ronneberger, O.,
  Tunyasuvunakool, K., Bates, R., {\v{Z}}{\'\i}dek, A., Potapenko, A., et~al.
\newblock Highly accurate protein structure prediction with alphafold.
\newblock \emph{Nature}, pp.\  1--11, 2021.

\bibitem[Lee et~al.(2021)Lee, Ippolito, Nystrom, Zhang, Eck, Callison-Burch,
  and Carlini]{lee2021deduplicating}
Lee, K., Ippolito, D., Nystrom, A., Zhang, C., Eck, D., Callison-Burch, C., and
  Carlini, N.
\newblock Deduplicating training data makes language models better.
\newblock \emph{arXiv preprint arXiv:2107.06499}, 2021.

\bibitem[Merity et~al.(2017)Merity, Keskar, and Socher]{merity2017regularizing}
Merity, S., Keskar, N.~S., and Socher, R.
\newblock Regularizing and optimizing lstm language models.
\newblock \emph{arXiv preprint arXiv:1708.02182}, 2017.

\bibitem[Perea \& Harer(2015)Perea and Harer]{perea2015sliding}
Perea, J.~A. and Harer, J.
\newblock Sliding windows and persistence: An application of topological
  methods to signal analysis.
\newblock \emph{Foundations of Computational Mathematics}, 15\penalty0
  (3):\penalty0 799--838, 2015.

\bibitem[Singh et~al.(2007)Singh, M{\'e}moli, Carlsson,
  et~al.]{singh2007topological}
Singh, G., M{\'e}moli, F., Carlsson, G.~E., et~al.
\newblock Topological methods for the analysis of high dimensional data sets
  and 3d object recognition.
\newblock \emph{PBG@ Eurographics}, 2, 2007.

\bibitem[Stanley \& Miikkulainen(2002)Stanley and
  Miikkulainen]{stanley2002evolving}
Stanley, K.~O. and Miikkulainen, R.
\newblock Evolving neural networks through augmenting topologies.
\newblock \emph{Evolutionary computation}, 10\penalty0 (2):\penalty0 99--127,
  2002.

\bibitem[Touvron et~al.(2023)Touvron, Lavril, Izacard, Martinet, Lachaux,
  Lacroix, Rozi{\`e}re, Goyal, Hambro, Azhar, et~al.]{touvron2023llama}
Touvron, H., Lavril, T., Izacard, G., Martinet, X., Lachaux, M.-A., Lacroix,
  T., Rozi{\`e}re, B., Goyal, N., Hambro, E., Azhar, F., et~al.
\newblock Llama: Open and efficient foundation language models.
\newblock \emph{arXiv preprint arXiv:2302.13971}, 2023.

\bibitem[Wolf et~al.(2020)Wolf, Debut, Sanh, Chaumond, Delangue, Moi, Cistac,
  Rault, Louf, Funtowicz, Davison, Shleifer, von Platen, Ma, Jernite, Plu, Xu,
  Scao, Gugger, Drame, Lhoest, and Rush]{wolf-etal-2020-transformers}
Wolf, T., Debut, L., Sanh, V., Chaumond, J., Delangue, C., Moi, A., Cistac, P.,
  Rault, T., Louf, R., Funtowicz, M., Davison, J., Shleifer, S., von Platen,
  P., Ma, C., Jernite, Y., Plu, J., Xu, C., Scao, T.~L., Gugger, S., Drame, M.,
  Lhoest, Q., and Rush, A.~M.
\newblock Transformers: State-of-the-art natural language processing.
\newblock In \emph{Proceedings of the 2020 Conference on Empirical Methods in
  Natural Language Processing: System Demonstrations}, pp.\  38--45, Online,
  October 2020. Association for Computational Linguistics.
\newblock URL \url{https://www.aclweb.org/anthology/2020.emnlp-demos.6}.

\bibitem[Zomorodian \& Carlsson(2004)Zomorodian and
  Carlsson]{zomorodian2004computing}
Zomorodian, A. and Carlsson, G.
\newblock Computing persistent homology.
\newblock In \emph{Proceedings of the twentieth annual symposium on
  Computational geometry}, pp.\  347--356, 2004.

\end{thebibliography}
\bibliographystyle{icml2024}

\newpage
\appendix
\onecolumn

\section{Model Details}

\subsection{AWD-LSTM}

We used a SoTA LSTM model design for our recurrent LMs.
We briefly describe features of this model and the optimization scheme.
The Averaged Stochastic Gradient Descent Weight-Dropped Long Short-Term Memory model (AWD-LSTM) is an enhancement over traditional LSTM models. 
Designed to address overfitting and improve regularisation in sequence modeling tasks, especially in natural language processing, it includes the following key-features: 
DropConnect on LSTM Weights: Randomly drops connections in the LSTM layers; a variant of dropout applied to recurrent networks.
DropConnect is a variation of the dropout technique, specifically applied to the recurrent connections in LSTM layers. 
Unlike standard dropout that randomly zeroes out activations, DropConnect zeroes out a random subset of the weights in the weight matrices. 
This randomness in dropping connections helps prevent overfitting by ensuring that the model does not rely too heavily on any single connection, thus improving the model's ability to generalise to unseen data.
Weight Dropping: Applies dropout to the weights of the recurrent connections, enhancing the model's ability to generalise.
This is a regularisation strategy where dropout is applied directly to the weights of the hidden-to-hidden recurrent connections within the LSTM units. 
By randomly setting a fraction of these weights to zero, the network is forced to learn redundant representations, making it more robust to the loss of specific connections. 
This technique is particularly effective for sequential data like natural language, where dependencies span over long sequences, as it encourages the model to capture and preserve information over longer time steps.
Non-monotonically Triggered Averaged Stochastic Gradient Descent (NT-ASGD): An optimiser that transitions from SGD to ASGD when a specified trigger in the validation loss is observed, promoting convergence stability.
In NT-ASGD, the transition from standard SGD to ASGD is triggered based on the model’s performance on a validation set. 
Specifically, ASGD begins when the validation loss fails to improve monotonically. 
ASGD averages the model parameters over time, which often leads to better generalisation and more stable convergence, especially in the latter stages of training.
Customisable Embedding and Dropout Layers: Allows different embedding sizes and dropout rates for each layer, providing flexibility in model architecture.
Embedding size determines the dimensionality of the vector space in which words or other inputs are represented. 
Different sizes can capture varying levels of semantic information. 
Similarly, allowing different dropout rates for each layer gives more control over how much regularisation is applied at different stages in the network. 
This customisation enables the model to be more finely tuned to the specific characteristics of the dataset it is trained on.
These features collectively enhance the model's performance in handling long-range dependencies and complex patterns in sequential data like natural language processing \cite{merity2017regularizing}. 

\subsection{Transformer}

\subsubsection{Pythia} 

Pythia is a comprehensive and extensible analytical suite designed for in-depth study and analysis of LLMs across various stages of training and scaling. 
In essence, it provides a series of autoregressive, decoder-only, GPT-based transformer models of 70m to 12B parameters, which are trained on the Pile dataset \cite{gao2020pile}.
Furthermore, it offers advanced tools for assessing the performance of LLM, their accuracy, efficiency, and throughput.
These include tools for analysing the training process, identifying patterns, bottlenecks, and monitoring and recording metrics like loss function values, gradient norms, and other pertinent indicators of model learning and convergence, which are essential for diagnosing training issues and optimising model architecture and hyperparameters.
Also, it allows to evaluate scaling and its impact on model performance, including memory management, processing speed, and scalability limits in terms of parameters, data size, and computational resources, enabling research on trade-offs and benefits of scaling.
Since it includes a set of benchmarks and evaluation metrics to assess the performance of LLMs on various tasks, it allows for comparing different models and understanding their strengths and weaknesses in specific domains or tasks.
Beyond those tools and methodologies for assessing complexities and bottlenecks in LLM development, our foremost interest is in their provision of checkpoints that allow us to understand information encoding throughout training and analyze the evolution of topological patterns at each layer of the GPT stack \cite{biderman2023pythia}.

\subsubsection{LLaMA} 

Linguistic LAnguage Model Attention (LLaMA) is a set of SOTA foundation language models between 7B and 65B parameters, trained on trillions of tokens, using publicly available datasets exclusively.
LLaMA-13B outperforms GPT-3 (175B) on most benchmarks, and LLaMA-65B is competitive with other equally sized models like Chinchilla-70B and PaLM-540B. 
After an initial leak that caused some outrage, all models were released to the research community.
LLaMA models distinguish themselves through novel attention mechanis and architectural design. 
They adopt a transformer-based architecture, similar to GPT-3 or BERT, but incorporate modifications in their attention mechanisms by employing a multi-head attention structure that allows for more efficient and effective processing of linguistic inputs. 
Each attention head is designed to capture different aspects of language like syntax, semantics, and context, which enables the model to generate more coherent and contextually relevant text.
LLaMA are trained on large datasets of about 4.75 TB, comprising diverse textual sources, including books (e.g., ArXiv, Gutenberg Project), articles (e.g. StackExchange), Code (e.g., GitHub) and crawled websites (e.g., Wikipedia), which ensures the models' proficiency across a wide range of topics and over 20 languages. 
The training process utilizes techniques like gradient checkpointing and mixed-precision training to optimise computational efficiency and reduce memory requirements. 
LLaMA achieved SoTA performance in various natural language processing domains, such as language understanding, generation, translation, fluency, and contextual relevance in generated text. 
The models are particularly strong at tasks requiring deep linguistic understanding, such as sentiment analysis, summarization, and question-answering \cite{touvron2023llama}.

\subsection{GCN}

Finally, we deploy Gated Convolutional Networks (GCN), a variant of Convolutional Neural Networks (CNN) that integrate gated linear units (GLU).
These GLU are instrumental in controlling the flow of information, similar to the gating mechanisms in LSTM, but optimised for convolutional structures.
Through convolution, inputs can be handled in a parallel manner, thus significantly enhancing computational efficiency compared to the sequential processing inherent in RNN.
This parallelization is particularly beneficial for large-scale language modeling tasks, where large amounts of data need to be processed efficiently. 
Another aspect crucial to language modelling is that Gated Convolutional Networks exhibit proficiency in capturing long-range dependencies within sequences.
This proficiency is primarily attained through the use of gated mechanisms and the inherent design of CNN that facilitates modeling of hierarchical structures and dependencies in data.
This is crucial for understanding contextual relationships in language, which is vital for tasks such as text generation and machine translation.
GCN outperform LSTM-based models on several language modeling benchmarks, which mainly is attributed to their ability to efficiently process large sequences of data while effectively capturing long-range dependencies that are prevalent in natural language \cite{dauphin2017language}.

\section{Data Preparation}

\subsection{AWD LSTM}

Based on previous research \cite{merity2017regularizing}, we implement a simple stacked LSTM based on a standard PyTorch LSTM model, surrounded by an embedding and decoding layer, as well as a number of hidden dropout layers that helps regularising the model.
Dropout probabilities for each layer and types of activation are tuned hyperparmeters. 
More concretely, we create two module lists for the rnns and their dropouts and loop each pair through the previously computed embeddings, which then is passed through the linear decoder for the next prediction.

We use custom corpora of text from several languages based on crawling news articles and open domain books, as well as synthetic randomly generated data, and tokenise it by adding BOS and EOS tokens, replacing words that occur less than 3 times with UNK tokens, transforming every word to lowercase, retaining digits, and stripping leading and trailing whitespace.
Finally, we split sentences into training and validation set 8:2 without the use of a test set.

We perform Bayesian hyperparameter tuning with Optuna \cite{akiba2019optuna} by first deciding a for a prior of each hyperparameter based on mean observations in the training data, then defining an objective function to return the validation loss of the model. 
Table \ref{tab_hyp_tune_lstm} explains the hyperparameters and priors.

\begin{table}[htbp]
\label{tab_hyp_tune_lstm}
\centering
\begin{tabular}{r|l|p{7cm}l}
Hyperparameter & Prior & Explanation\\
\hline
\texttt{n\_hid} & UniformInt(50, 1000) & Number of hidden states to use. When using multiple LSTM layers, this is the total size of all hidden states\\
\texttt{n\_layers} & UniformInt(2, 5) & Number of LSTM layers. Fixed to 1 when using single layer LSTM\\
\texttt{emb\_sz} & UniformInt(10, 500) & Embedding size dimension. Fixed to 300, when using Glove\\
\texttt{tie\_weights} & UniformInt(0, 1) & Whether to set same weights for decoder and encoder\\
\texttt{out\_bias} & UniformInt(0, 1) & Whether to use a bias parameter in the decoder\\
\texttt{output\_p} & UniformFloat(0, 0.5) & Dropout probability of output\\
\texttt{input\_p} & UniformFloat(0, 0.5) & Dropout probability of input\\
\texttt{hidden\_p} & UniformFloat(0, 0.5) & Dropout probability hidden states\\
\texttt{embed\_p} & UniformFloat(0, 0.5) & Dropout probability embeddings\\
\texttt{weight\_p} & UniformFloat(0, 0.5) & Dropout probability of weights\\
\texttt{lr} & UniformFloat(1e-4, 0.1) & Learning rate to use\\
\texttt{wd} & UniformFloat(0, 0.5) & Weight decay\\
\texttt{freeze\_epochs} & UniformInt(0, 95) & Number of epochs to freeze pretrained embeddings for. Fixed to 0, when not using pretrained embeddings\\
\end{tabular}
\caption{Hyperparameters and priors of the LSTM model}
\end{table}

Those choices are constrained by computational capabilities, especially when using glove embeddings in combination with multiple LSTM layers.
For example, we need to limit the number of total hidden states to about 1,000 during hyperparameter tuning due to GPU memory limits.
Table \ref{tab_hyp_res_lstm} displays the results of the hyperparameter tuning for the following types of LSTM: multilayer, single layers, pretrained embeddings and non-pretrained embeddings.

\begin{table}[htbp]
\label{tab_hyp_res_lstm}
\begin{tabular}{l|lllll}
\toprule
Name & Multi Pretrained & Multi Random & Single Pretrained & Single Random \\
  \midrule
    Perplexity & 163.56 & 179.79 & 167.10 & 179.39 \\
    emb\_sz & nan & 768.00 & nan & 445.00 \\
    freeze\_epochs & 78.00 & nan & 71.00 & nan \\
    n\_hid & 349 & 979 & 762 & 327 \\
    n\_layers & 2.00 & 3.00 & nan & nan \\
    embed\_p & 0.14 & 0.36 & 0.00 & 0.40 \\
    hidden\_p & 0.22 & 0.41 & 0.24 & 0.42 \\
    input\_p & 0.07 & 0.20 & 0.37 & 0.36 \\
    lr & 0.02 & 0.01 & 0.04 & 0.04 \\
    out\_bias & True & True & False & True \\
    output\_p & 0.48 & 0.44 & 0.44 & 0.47 \\
    weight\_p & 0.41 & 0.48 & 0.17 & 0.34 \\
    wd & 0.08 & 0.25 & 0.17 & 0.20 \\
    \bottomrule
  \end{tabular}
\caption{Hyperparameter settings and their results. *~emb\_sz is shown as NA, because it isn't optimised in hyperparameter tuning, but the actual emb\_sz of the architecture is 300, which is the size of the glove embeddings. freeze\_epochs value shown as NaN means that freezing isn't implemented at the time. n\_layers being Nan means single LSTM layer. wd being NAN means the default value of 0.01.}
\end{table}

We retrieve hidden states over a random sampling of 2000 input sentences from the training corpus (by performing inference) at 100 evenly space checkpoints during training.  
Before retrieval, we set all hidden states to 0, put the model into evaluation mode, and turn off dropout and gradient calculations.
We use the HDF5 binary data format, which allows efficient manipulation of large data tensors stored on the hard drive.
With every new sentence used for analysis, we allocate a tensor with the dimensions (number of tokens) x (hidden state dimension) x (number of epochs). 
In order to retrieve those tensors from tokens of each sentence, we loop over all tokens while recording the corresponding embedding and hidden state vectors at each layer.

\subsection{Transformer}

\subsubsection{Pythia}

Since Pythia is pretrained mostly on English test, we perform inference on a test corpus of English sentences, as well as three variants of random corpora generated using a subset of English tokens recognized by the Pythia models.
The rendom samples are the following.
A "permutation" corpus - obtained by randomly shuffling the order of words within a subset of English sentences.
A "Zipf" corpus - obtained by generating random sequnces of words according to the unigram distribution matching that present in the natural corpus.
A "uniform" corpus - where words are generated with equal probability.
All these synthetic corpora preserve sentence length distribution of the natural corpus.
Corpora are tokenised by model-default tokeniser, lower-cased, transformed into lists of strings and inference is conducted on sentence level. 
We conduct inference on the 70m and 160m deduped \cite{lee2021deduplicating} Pythia models \cite{biderman2023pythia} with these corpora, and record embeddings and hidden states as numpy arrays with size (sequence length x hidden size) for 100 training checkpoints.
The chosen training checkpoints were: the initial checkpoint (0), ten log-spaced steps (1,2,4,...512), and 89 linearly spaced (using numpy linspace method) steps in range (1000 ... 143000) .

Model selection was limited by computational constraints.
We select all layers for the 70m model, and three layers for the 160m model, as summarised in table \ref{table:model_selection_pythia}.

\begin{table}[h]
\centering
\begin{tabular}{c|c|c}
 & \textbf{70m model} & \textbf{160m model} \\
\hline
\textbf{Hidden size} & 512 & 768 \\
\hline
\textbf{Total number of layers} & 6 & 12 \\
\hline
\textbf{Chosen layers} & all & 4, 8, 12 \\
\hline
\textbf{File size per step} & ca. 170MB & ca. 375MB \\
\end{tabular}
\caption{Overview Pythia models chosen}
\label{table:model_selection_pythia}
\end{table}

We deploy the standard Pythia tokeniser \cite{biderman2023pythia}, using their training corpus that has a vocabulary size of 50,254 tokens. 
Since it treats spaces as prefix of tokens and does not feature a dedicated beginning-of-sentence token, we adjusted the settings to the input format of list of strings.

\subsubsection{LLAMA}

Due to restrictions in computational resources, we conduct sentence-level inference on thse same custom-made research corpora, using the smallest 7B parameter LLaMA model \cite{touvron2023llama}, of which we use the Int8 quantized version from the Hugging Face transformer library, which deploys mixed-precision decomposition for maintaining model performance \cite{wolf-etal-2020-transformers}. 
We use the provided byte-pair-encoding tokeniser based on sentencepiece that is trained on a trillion tokens from publicly available data of the top 20 most spoken languages.
Since it does not prepend a prefix space if the first token is the start of a word and there is no padding token in the original model, we unset the default padding token. 
In absence of time steps or checkpoints, we collect input embeddings and hidden states of the output layer only by performing inference on the fully trained model.
Due to the lack of training checkpoints we can not produce a regular "perforation over epochs" plot as for the other models, but instead produce a bar-chart visualization by randomly sampling 2000 sentences and summarizing results with a histogram of perforation values (quantized to fall between different intervals - that is the height of the bar corresponds to the fraction of sentences in the sample that had perforation with a given interval of values).

\subsection{GCN}
We implement a Gated Convolutional Neural Network (CNN) \cite{dauphin2017language} model in PyTorch, and train it on the same data as the LSTM network.
For that, we first construct and initialise a CNN Module for sequential data processing, which uses GloVe embeddings to map vocabulary indices to high-dimensional vectors of fixed size (300). 
The CNN architecture is comprised of an embedding layer, multiple convolutional layers, and padding strategies to maintain uniform input sizes for convolutional operations.
We further integrate dropout layers (0.75 and 0.90) for regularisation and to mitigate overfitting, and a fully connected linear layer for output dimensionality mapping to the vocabulary size.

Second, to extract and process the initial embeddings and hidden states, we first tokenise and the input sentences.
Then, we set the model to evaluation mode, to affect layers like dropout and batch normalisation differently than in training mode, and detach the extracted hidden states and embeddings from the computational graph turning off gradient accumulation.
Finally, we extract outputs for each layer from the CNN, converting and reorganizing these hidden states into numpy arrays, and store those in an HDF5 database for further processing.

\section{Methods of Analysis}
\subsection{Persistence Modules}

In order to associate topological spaces to the sentences of the corpus, we compute the Vietoris-Rips Complex (VRC).
It can be defined for sets of vectors using a technique inspired by hierarchical clustering methods.

Given a set of data points \( \mathbb{X} = \{x_1, \dots, x_n\} \) embedded in a vector space with metric $d$, and real number \( \epsilon \geq 0 \), we define the Vietoris-Rips complex \( \text{VRC}_{\epsilon}(\mathbb{X}) \) as the set of simplices \( \sigma = [x_0, \dots, x_k] \) s.t. \( \forall i,j \in \{0, \dots, k\}, d(x_i, x_j) \leq \epsilon \).
This definition produces an abstract simplicial complex \( \mathbb{K} \) with vertex set \( \mathbb{X} \) and set \( \Sigma \) of subsets of \( \mathbb{X} \) (the simplices), with the property that for any \( \sigma \in \Sigma \), all elements of the power set of \( \sigma \) belong to the complex \( \Sigma \).
A good way of thinking about this process is as sliding an $\epsilon$ ball across the ambient space, while recording simplices spanned by vertices that are captured inside the boundary of the ball.
Note that as $\epsilon$ grows, the number of simplices always increases, and VRCs generated by increasing values of $\epsilon$ form a filtration of abstract simplicial complexes containing each other.

A major difference between topological language representations such as the word manifold, introduced in \cite{fitz2022shape}, and the representation manifolds in neural language models is that the former arises from discrete data without a canonical choice for topology. In contrast, linguistic unit representations exist in a metric space, which naturally comes equipped with the open $\epsilon$-ball topology.
However, it is not obvious what value of $\epsilon$ we should choose for the construction of the VRC.
Furthermore, even small amount of noise can alter the homotopy type of a VRC associated to a random sample of points, by generating simplices that alter topological information in significant ways.
These issues are main reasons why, until recently, topology was not very useful in dealing with real world data.

The solution to both these issues resulted from work of Edelsbrunner et al. and was further developed by Zomordian and Carlsson \cite{zomorodian2004computing} in form of \emph{persistent homology}.
For each fixed value of $\epsilon$ in a Vietoris-Rips Complex filtration, we can compute homology groups of the resulting complex and depict each generator as an interval with endpoints corresponding to the \emph{birth} and \emph{death} $\epsilon$ values of the homology class generator that it represents.
Note that as $\epsilon$ value grows, higher dimensional simplices appear in the VRC filtration, and these simplices will have boundaries that were cycles generating homology classes for earlier values of $\epsilon$.
Because of this, as $\epsilon$ grows we will see some bars appear and disappear, until all bars end, and only a single bar remains in dimension zero, which corresponds to the connected component of the entire point cloud, when it is finally enclosed by a large enough $\epsilon$ ball (see figure \ref{persistence}).
The interval representation is often referred to as a persistence \emph{bar code}, and was initially inspired by the \emph{quiver representation} of a sequence of vector spaces.
This construction solves the first issue by taking all values of $\epsilon > 0$ into account.
It also solves the second issue, by interpreting the longest bars as representing informative topological features and the shortest bars as topological noise.


Because the VR complexes for larger $\epsilon$ values contain those with smaller values, we have a nested family of simplicial complexes.
That means that there is a simplicial map from the set of simplices for a given $\epsilon$ value into the set of simplices generated by any larger value.
Simplicial maps induce homomorphisms on homology groups.
If we compute homology with field coefficients, we can think of the object generated by all possible $\epsilon$ values as a sequence of vector spaces with linear maps between them.
There is a similar visualization technique in linear algebra known as \emph{quiver representation}.
The barcode diagram for that quiver representation corresponds to the persistence diagram generated by the given data sample.
Eventually as $\epsilon$ increases the number of bars decreases in each dimension, as homology classes disappear.
With a finite number of points there are finite number of bars, and finite number of merge events.

We do not actually need a metric, but only a nested family of spaces (a filtration) where inclusion maps induce simplicial maps on the corresponding abstract simplicial complexes associated to each space.
Then, in a fixed dimension, we have a quiver representation of homology groups with the associated induced maps on homology.

The bar codes generated from a point cloud can be interpreted as measuring the shape of the space in the following way.
In dimension zero the barcode is equivalent to a dendrogram of a hierarchical clustering of the points in the data set.
In dimension one, bars represent loops.
Dimension two shows spherical cavities.
Higher dimensions correspond to $( n-1 )$-dimensional spheres wrapped around $n$-dimensional cavities.

Formally, topological persistence can be computed using tools from commutative algebra of modules over principal ideal domains and methods such as Smith Normal Form decomposition of the boundary map matrices.
The computation is more intricate than the process used in simplicial homology theory, because we have to keep track of the extra scale parameter.
The result is a significantly more efficient algorithm for computing persistence intervals from a filtration of simplicial complexes \cite{zomorodian2004computing}, than performing a separate homology computation for each space in a filtration.
It is informative to see an example of such computation done directly.

Suppose we generated a filtration of abstract simplicial complexes from a point cloud by considering a range of parameter values.
For a finite set of initial points, there will be only a finite number of simplices that appear, and we end up with a finite sequence of simplicial complexes that contain each other.

$$
X_0 \subseteq X_1 \subseteq X_2 \dots
$$

\begin{figure}
  \centering
  \includegraphics[width=0.5\textwidth]{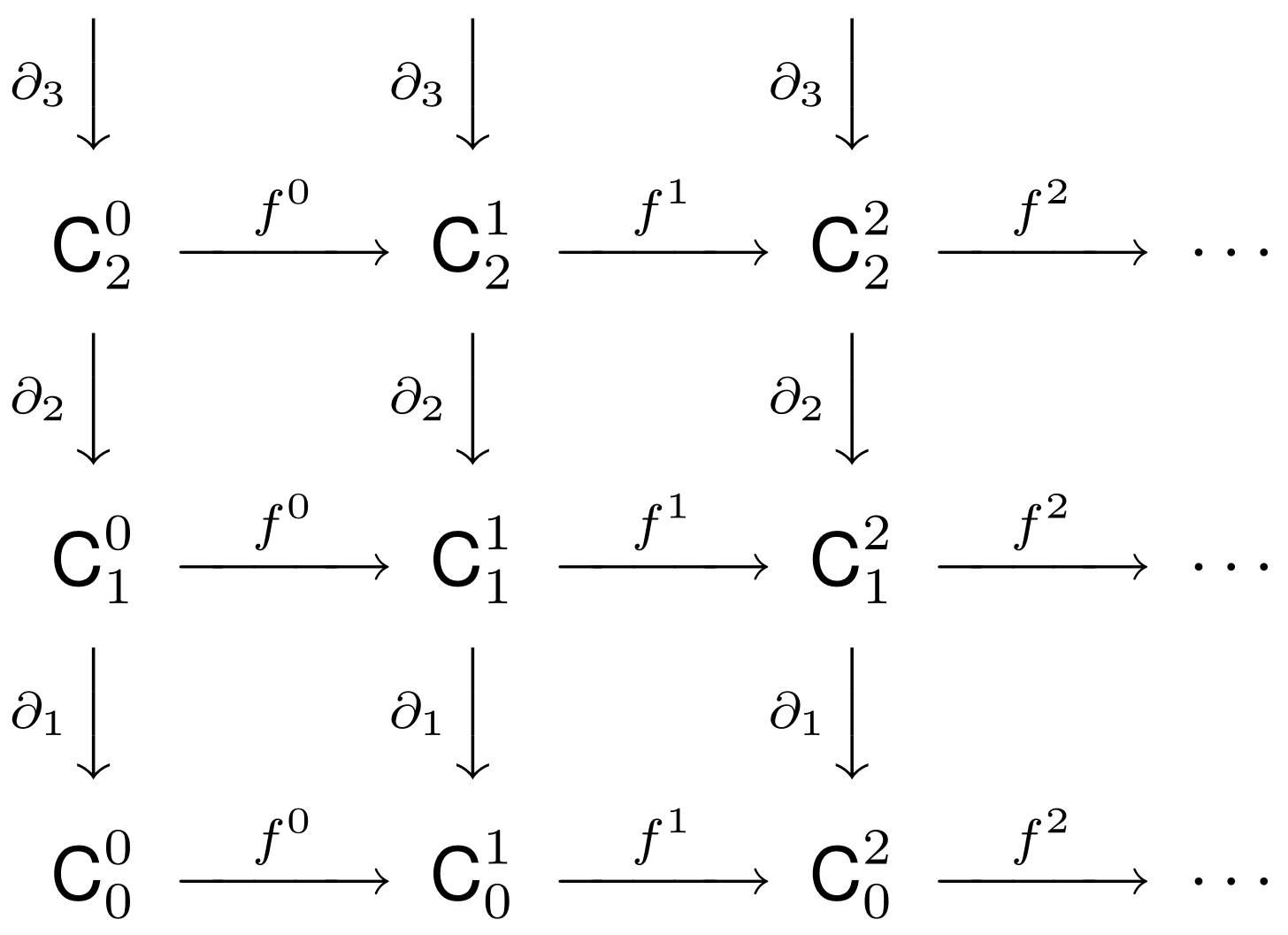}
  \caption{Chain groups of a filtration of spaces. The columns show boundary chain complex for each complex in a sequence. The rows show maps induced by the inclusion of each subcomplex within a larger complex corresponding to higher parameter values. The lower index marks the dimension within a complex, and the upper index specifies the ordering within a filtration.}
  \label{persistence_chain_groups_diagram}
\end{figure}

Every simplex has a definite birth time $t$ which is the index of the complex within the filtration such that $\sigma \in X_t$ and $\sigma \notin X_{t-1}$.
We can arrange the boundary maps as columns, and the inclusion induced homomorphisms between chain complexes of each simplicial complex in the sequence as rows, into a commutative diagram as in figure \ref{persistence_chain_groups_diagram}.

The goal of the persistence algorithm is to compute homology once, simultaneously for all time steps, while also reducing the sizes of matrices involved.
In order to do that, we compute homology of the \emph{persistence module}, which will encode information about homology of all the subcomplexes in the sequence simultaneously.
The idea behind this construction is to label the simplices within the filtration with their corresponding birth times.
Algebraically, we convert the boundary matrices, such as those used in simplicial homology, from integers to polynomials in the \emph{birth time} variable $t$, which we adjoin to the ring of integers $\mathbb{Z}$.
Let $X$ be the whole complex, and $X_k$ be the subcomplex in the filtration where a simplex $\sigma$ appears for the first time.
The strategy of the persistence algorithm is to track birth times algebraically.
Rather than letting $C_n(X)$ contain combinations of simplices with coefficients in $\mathbb{Z}$ we upgrade the coefficients to polynomials in $\mathbb{Z}[t]$.
Given $n$-simplex $\sigma$ with birth time $k$, we represent it by $t^k \sigma$.
If the simplex existed from the start (time index 0), we just get $\sigma$ as usual.
However, if a simplex shows up at the first time step, we represent it by $tk$.
If it showed up in the second complex within the filtration, we would represent it as $t^2k$, and so forth.
With this convention, our boundary maps now contain polynomials in $t$.
Reduction of matrices over $\mathbb{Z}[t]$ involves polynomial division, and comes with issues as discussed in \cite{zomorodian2004computing}.
However, we could also perform this computation over polynomials with field coefficients, which makes calculation more straightforward.
This hides torsion related phenomena, but if we are just interested in computing the Betti numbers, we can transition to vector spaces with coefficients in $\mathbb{Q}[t]$, and leverage existing linear algebra packages such as numpy to perform reduction operations.

\begin{figure}
  \centering
  \includegraphics[width=0.75\textwidth]{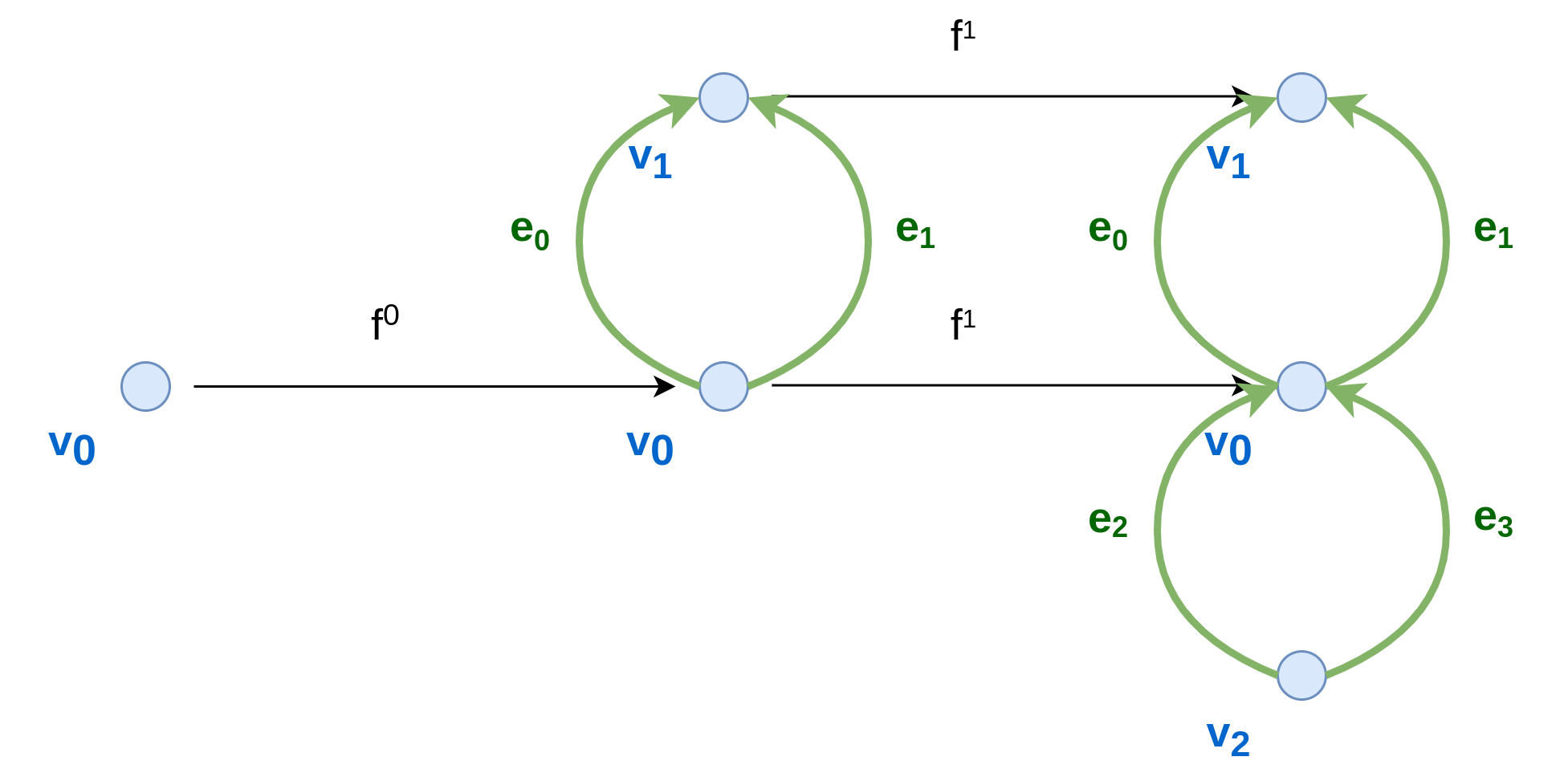}
  \caption{A filtration of 3 spaces. Note that the vertices $v_0$ and $v_1$ forming the boundary of edges $e_0$ and $e_1$ have different birth times, which is encoded by the exponents of $t$ in the entries of the boundary operator.}
  \label{persistence_filtration}
\end{figure}

For the remainder of the discussion, we define persistent chain groups $PC_k$ to be free modules over the polynomial ring $\mathbb{Q}[t]$, generated by the simplices of $X$.
In order to make this computation work, we need to define the boundary maps in a way that that satisfies the chain complex condition necessary to define homology (c.f. \cite{hatcher2001algebraic}).
The following definition of boundary matrices works.
The $k$-th boundary map $\partial_k: PC_k \rightarrow PC_{k-1}$ is a linear map over $\mathbb{Q}[t]$.
In order to represent it as a matrix, we label rows and columns of boundary matrices with simplices of X.
The difference between it, and the boundary map used in simplicial homology, is that the entries now encode not only presence of simplices, but also their relative birth times (difference between the indices of spaces in the filtration at which they first appear).
Suppose that a column of the persistent boundary matrix corresponds to a $k+1$ simplex $[v_0, v_1, \dots, v_k]$, which first appeared in the filtration at index $B$ (for \emph{birth time}).
In this case, the $i$-th entry of this column will be of the form $(-1)^i t^{B-b}$, which corresponds to the face $[v_0, \dots, \hat{v}_i, \dots, v_k]$ with a birth time $b$.
Note that the birth times of simplices marked by columns must be greater than those that mark rows (since a simplex can not appear unless its boundary is already present), thus the exponent of $t$ is non-negative.
Zomordian et al. proved that this definition of a boundary operator induces a chain complex (hence we can take homology quotients) \cite{zomorodian2004computing}.
For instance, the persistent boundary operator in dimension 1 (that is a matrix representation of $\partial_1: PC_1 \rightarrow PC_0$) for the filtration in figure \ref{persistence_filtration} would be written as follows under this convention.

$$
\partial_1 =
\begin{array}{rcl}
  &\color{green}\begin{array}{c}\ e_0 \ \ \ \ \ e_1 \ \ \ \ \ e_2 \ \ \ \ \ \ e_3 \end{array} \\
    \color{blue}\begin{matrix}v_0 \\ v_1 \\ v_2 \end{matrix}\color{black}\hspace{-1em} &\begin{pmatrix}-t & -t & \ \ t^2 & \ \ t^2 \\ \ \ 1 & \ \ 1 & \ \ 0 & \ \ 0 \\ \ \ 0 & \ \ 0 & -1 & -1 \end{pmatrix}&\hspace{-1em}\\ \color{blue}\end{array}
$$

which can be column reduced to

$$
\begin{pmatrix}-t &  \ \ \ \ t^2 & \ 0 & 0 \\ \ \ 1 & \ \ 0 & \ 0 & 0 \\ \ \ 0 & -1 & \ 0 & 0 \end{pmatrix}
$$

and is hence of rank 2.

\begin{figure}
  \centering
  \includegraphics[width=0.42\textwidth]{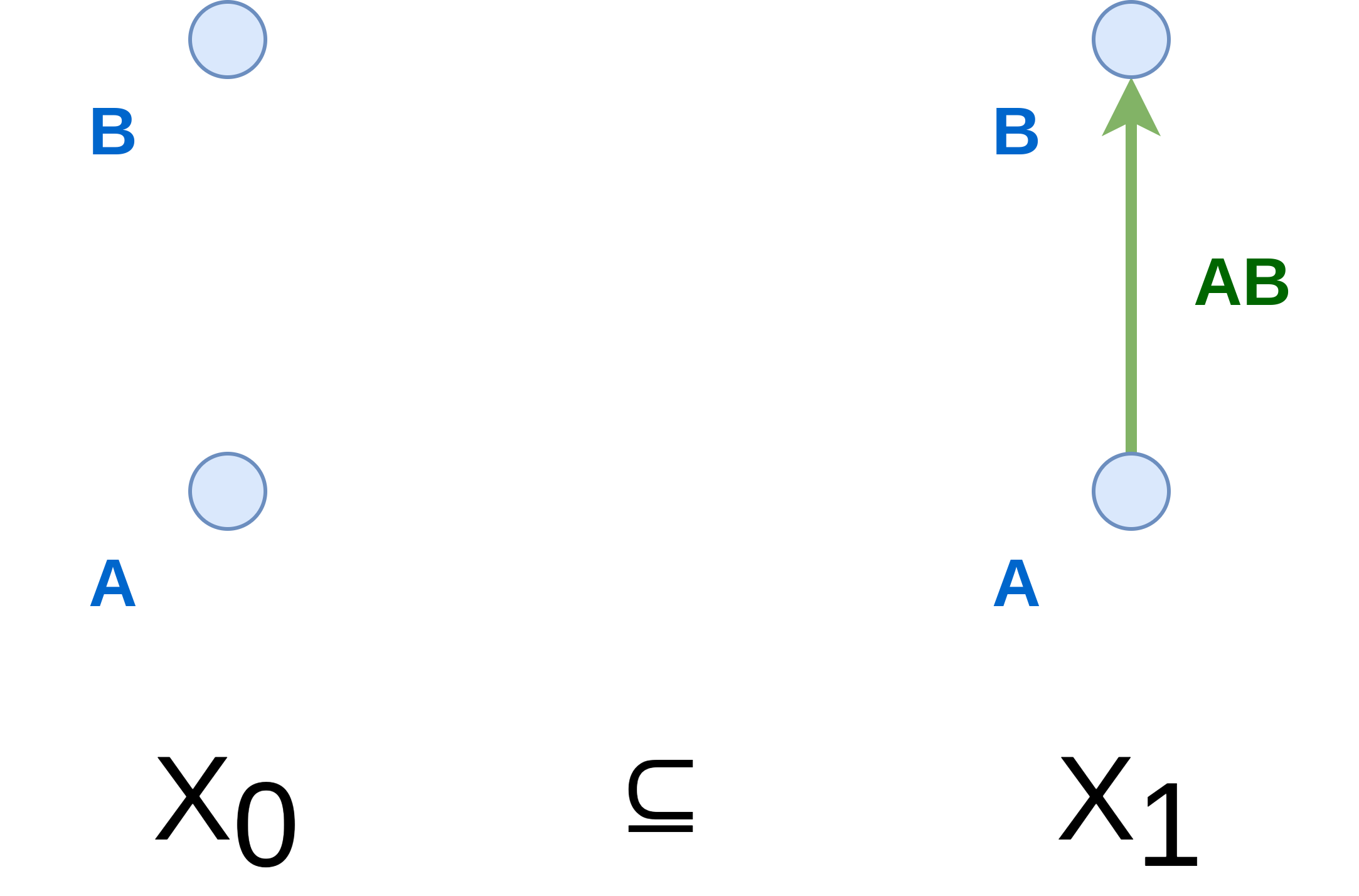}
  \caption{An edge appears at some value of $\epsilon$ between two vertices $A$ and $B$.}
  \label{minimal_persistence_example}
\end{figure}

It is informative to see persistent homology computation for a minimal example.
Consider an event of an edge appearing between two vertices as in figure \ref{minimal_persistence_example}.

The persistent boundary matrix in dimension 1 is of the form:

$$
\partial_1 =
\begin{array}{rcl}
  &\color{green}\begin{array}{c}AB\ \end{array} \\
  \color{blue}\begin{matrix}A \\ B\end{matrix}\color{black}\hspace{-1em} &\begin{pmatrix}-t\\t\end{pmatrix}&\hspace{-1em}\\ \color{blue}\end{array}
$$

Note that the exponents of $t$ in the entries above are the differences between birth times of the edge $AB$ and its boundary vertices $A$ and $B$ (both of which are $1$).
Thus the image of $\partial_1$ (i.e. the space of cycles) is rank one.

The kernel of persistent boundary in dimension 0 is is two dimensional, generated by the two vertices $A$ and $B$ with birth time $0$.

$$
\text{span}_{\mathbb{Q}[t]} \{ \begin{pmatrix} 0 \\ 1 \\ \end{pmatrix}, \begin{pmatrix} 1 \\ 0 \\ \end{pmatrix} \} = \{ \begin{pmatrix} q(t) \\ r(t) \\ \end{pmatrix} : q, r \in \mathbb{Q}[t] \}
$$

Since modules are not uniquely characterized by their dimension (as regular vector spaces are up to isomorphism), we can not simply do arithmetic on dimensions (which would wrongly suggest that the quotient is one dimensional), but instead need to work with the polynomial ring to figure out the rank of homology.

We need to compute the quotient:

$$
PH_0 = \text{ker} \partial_0 / \text{im} \partial_1 = \{ \begin{pmatrix} q(t) \\ r(t) \\ \end{pmatrix} + \{ \begin{pmatrix} -tp(t) \\ tp(t) \\ \end{pmatrix} : p \in \mathbb{Q}[t] \} : q, r \in \mathbb{Q}[t] \}
$$

Observe that we can always find a representative in each coset, where the first component is a constant, say $a$, simply by solving $q(t) - tp(t)$ for a given polynomial $q \in \mathbb{Q}[t]$ (just set the coefficients of $p$ so they cancel out all but the constant term of $q$ - we can not zero out the constant term because $tp(t)$ is at least of degree $1$).

$$
\begin{pmatrix} a \\ r(t) + t\tilde{p}(t) \\ \end{pmatrix}
$$

where $\tilde{p}$ has been fixed (from an arbitrary polynomial $p$, in order to make the first component constant), but $r$ is still an arbitrary polynomial.
This argument implies that the quotient is two dimensional. In fact we can now write:

$$
PH_0 = \{ \{ \begin{pmatrix} a \\ r(t) + t\tilde{p}(t) \\ \end{pmatrix} + \begin{pmatrix} -tp(t) \\ tp(t) \\ \end{pmatrix}: p \in \mathbb{Q}[t] \} : q, r \in \mathbb{Q}[t] \}
$$

and split these cosets into two summands of the form

$$
\{ \{ \begin{pmatrix} a \\ 0 \\ \end{pmatrix} + \begin{pmatrix} -tp(t) \\ tp(t) \\ \end{pmatrix} : p \in \mathbb{Q}[t] \}  + \{ \begin{pmatrix} 0 \\ r(t) + t\tilde{p}(t) \\ \end{pmatrix} + \begin{pmatrix} -ts(t) \\ ts(t) \\ \end{pmatrix}: s \in \mathbb{Q}[t] \} : q, r \in \mathbb{Q}[t] \}
$$

The above set is a sum of two spans (up to isomorphism):

$$
PH_0 \cong \mathbb{Q} \oplus \text{span}_{\mathbb{Q}[t]} \{ \begin{pmatrix} 0 \\ 1 \\ \end{pmatrix} \}
$$

where the fist summand comes from $a$ being an arbitrary constant term of the erased polynomial $q$, which gave $\mathbb{Q}$ as the coset represented by $\begin{pmatrix} a \\ 0 \\ \end{pmatrix}$ in the isomorphism above.

\begin{figure}
  \centering
  \includegraphics[width=0.5\textwidth]{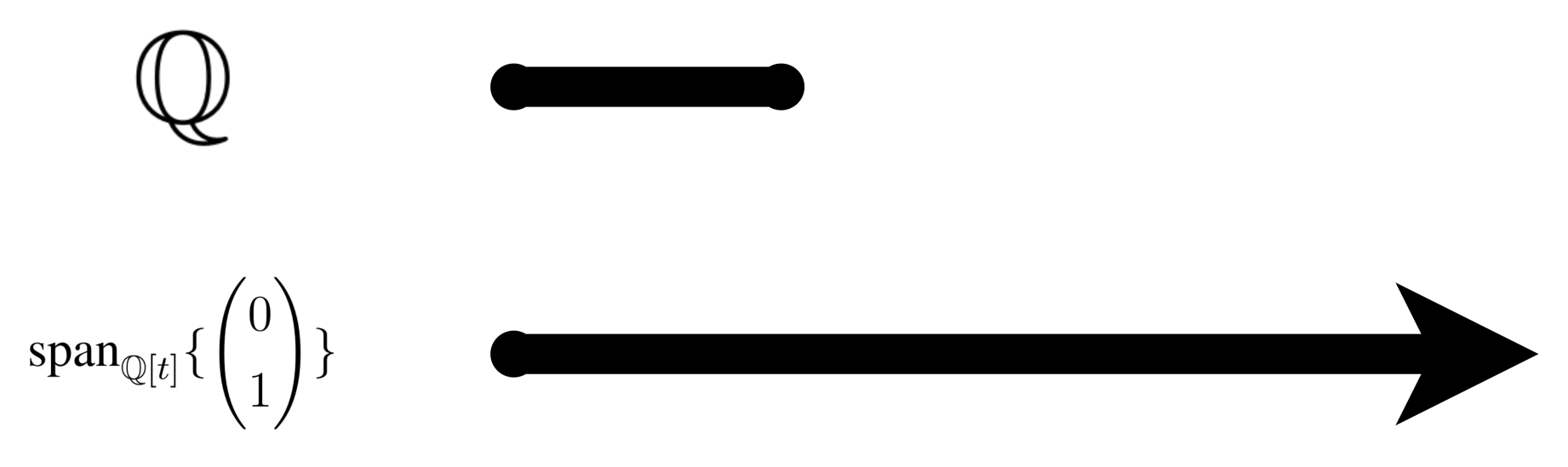}
  \caption{Bar code corresponding to an edge appearing between two vertices. The two bars are marked with direct summands of the homology group of the associated persistence module in dimension 1 as in the example calculation. The arrow head signifies that the corresponding homology class survives indefinitely.}
  \label{two_bars}
\end{figure}

This homology group encodes more data than the regular (i.e not persistent) homology we used previously.
In particular we can read Betti numbers over time.
They are usually represented as bar codes (such as those in figure \ref{persistence}).
From our computation we can conclude that there are two bars in dimension 0 (figure \ref{two_bars}).
The first term $\mathbb{Q}$ generates the first bar, which corresponds to a connected component that exists at time zero, but dies at time 1.
We can think of it as the connected component of vertex $A$ (or $B$).
The second term in the direct sum above is a longer bar, that represents the connected component which exists at time 1.
This component is the entire space, as $A$ and $B$ are joined by an edge.

In practice, there are many optimization that can be done while computing persistent homology.
In particular, the boundary matrices are very sparse, and algorithms used in practice take advantage of that fact.
There are also topological optimizations based on Morse theoretic ideas, that identify and remove contractible parts of the simplicial complex before computing homology (since the resulting spaces are homotopy equivalent and have isomorphic homology groups).

\subsection{Simplicial Mapping Approximation}

We produced graph visualizations of embedding and hidden state vectors corresponding to sentences sampled from the training corpora.
The technique we use here can be thought of as a topological dimensionality reduction method, where the goal is to summarize the shape of our representation space with a rough sketch in form of a low dimensional topological manifold.
This reduced representation can be thought of as a map approximating the shape of our embedding space.
Such description can be visually inspected by a human, while remaining more topologically informative than a naive projection.
Instead of growing $\epsilon$ balls around points directly, we can map them to a different space first, define an open cover, and then cluster the original points within the preimage of each cover set.
This produces a summary of the topological features in the embedding with a simplicial complex of a chosen dimension \cite{singh2007topological}.
In this set of experiments, we generated 1-dimensional simplicial complexes (i.e. graphs) from every sentence.
Figure \ref{mapper} shows a visualization of this process for a point cloud sampled from the circle ($\mathbb{S}^2$).
The general procedure can be summarized as follows.

\begin{center}
  \setlength{\fboxsep}{1em}
  \setlength{\fboxrule}{0.1em}
  \noindent\fcolorbox{black}{lightgray}{%
    \minipage[t]{\dimexpr0.98\linewidth-2\fboxsep-2\fboxrule\relax}

      Given data points \( \mathbb{X} = \{x_1, \dots, x_n\}, x_i \in \mathbb{R}^d \), a function \( f: \mathbb{R}^d \rightarrow \mathbb{R}^m, m < d \), and a cover \( \mathcal{U} = \bigcup_{i \in \mathcal{I}} U_i \) of the image \( f(\mathbb{X}) \) (where \( \mathcal{I} \) is some index set) we construct a simplicial complex as follows:

      \begin{enumerate}
        \item For each \( U_i \in \mathcal{U} \), cluster \( f^{-1}(U_i) \) into \( k_{U_i} \) clusters \( C_{U_{i,1}}, \dots, C_{U_i,k_{U_i}} \)
        \item \( \underset{U_i \in \mathcal{U}}{\bigsqcup} \{C_{U_{i,1}}, \dots, C_{U_i,k_{U_i}}\} \) now define a cover of \( \mathbb{X} \); calculate the nerve of this cover
      \end{enumerate}

      Nerve is defined in the following way. Given a cover \( \mathcal{U} = \bigcup_{i \in \mathcal{I}} U_i \), the nerve of \( \mathcal{U} \) is the simplicial complex \( \mathcal{C}(\mathcal{U}) \) where the 0-skeleton is formed by the sets in the cover (each \( U_i \) is a vertex) and \( \sigma =[U_{j_0}, \dots, U_{j_k}] \) is a k-simplex \( \iff \bigcap\limits_{l=0}^{k} U_{l_k} \neq 0 \)
    \endminipage
  }
\end{center}

\begin{figure}
	\centering
  \includegraphics[width=0.49\textwidth]{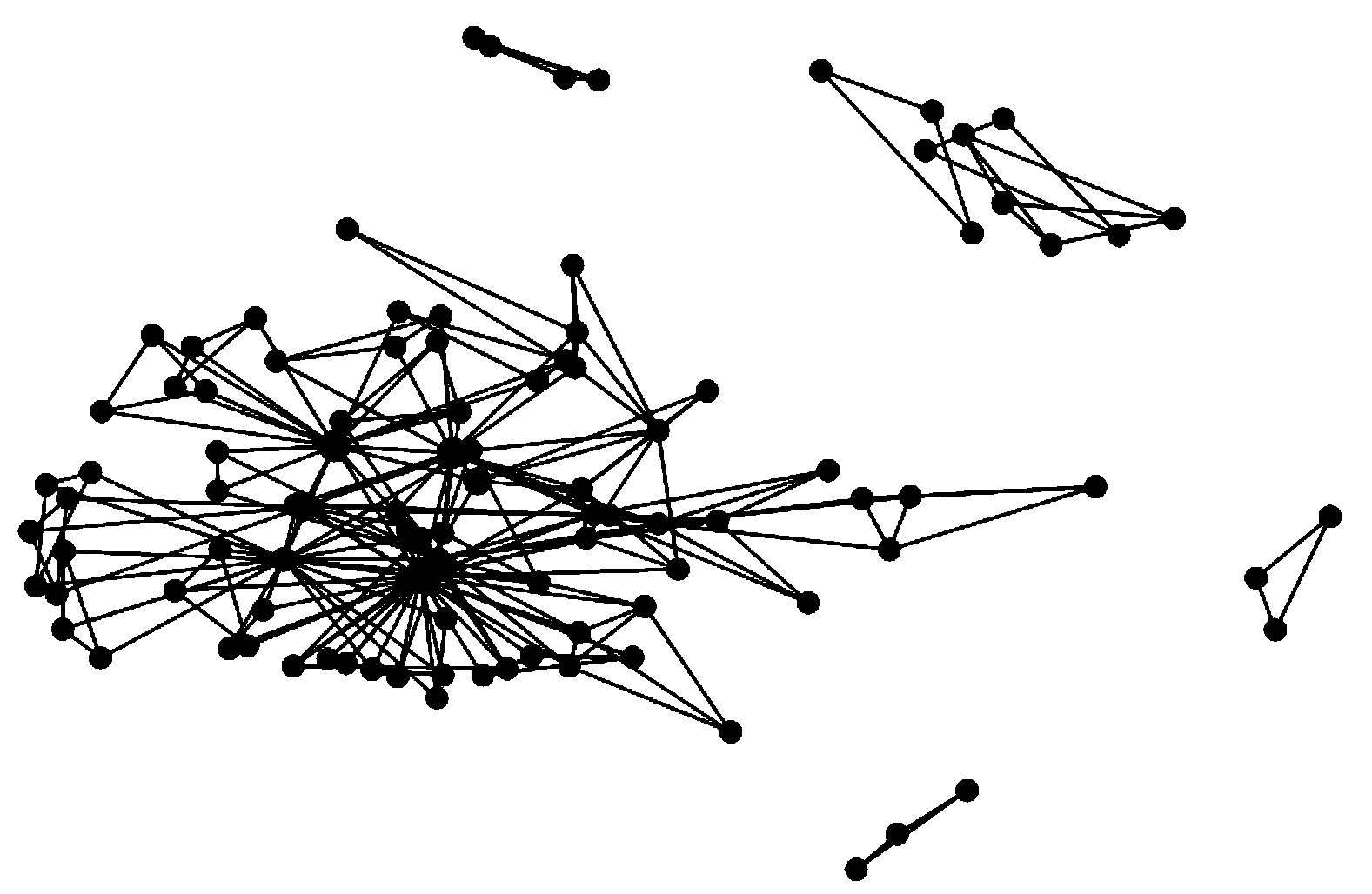}
	\includegraphics[width=0.49\textwidth]{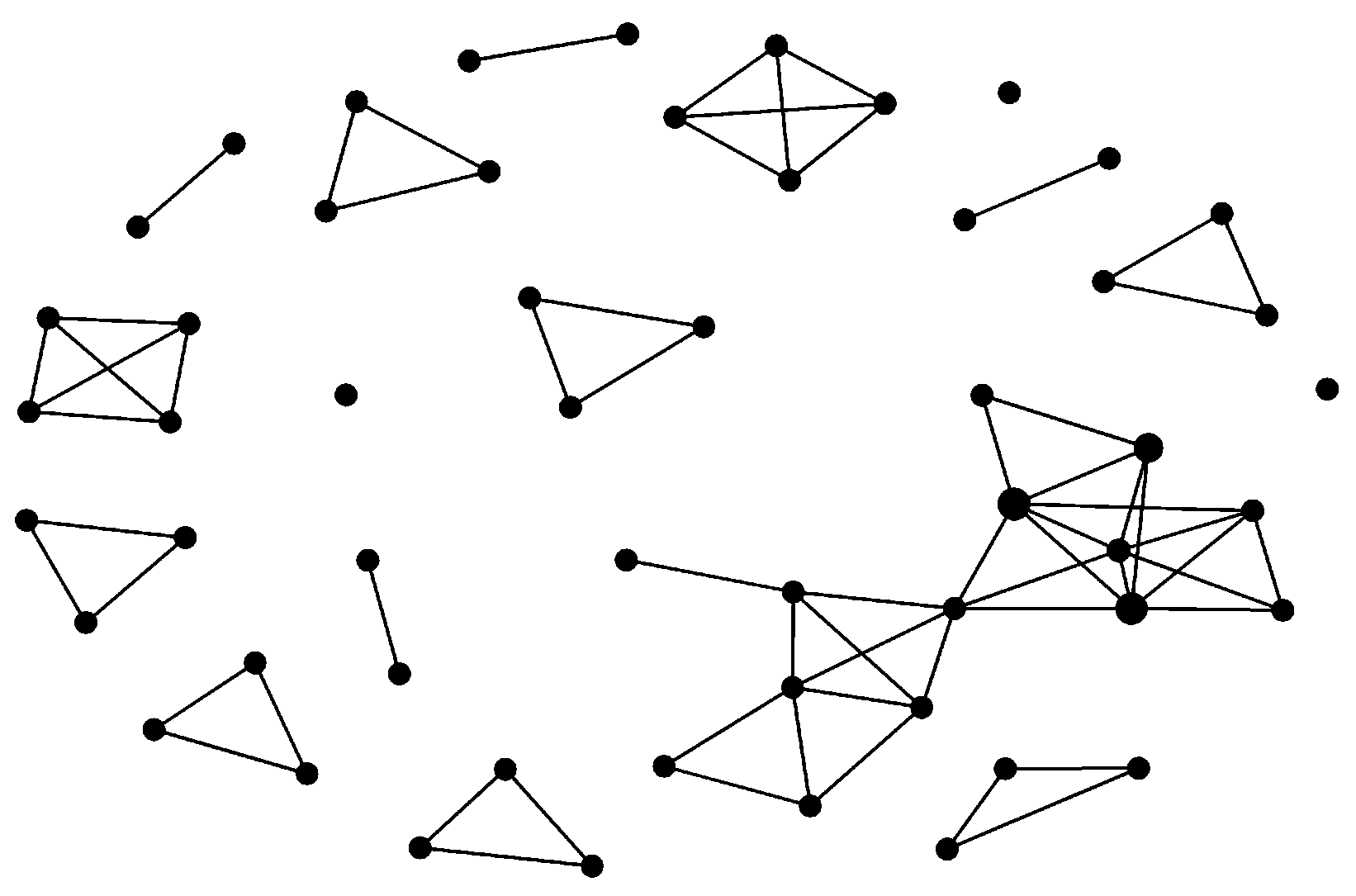}
  \caption{A random article composed of approximately 2600 words represented by the LSTM model (hidden state reset after each sentence). Hidden states are on the left side of the figure. Embeddings are on the right. Projection onto the first five principal components was performed before clustering.}
	\label{hidemb_100_mapper}
\end{figure}

We also attempted visualizations of larger sections of the corpus.
Generating simplicial mapping projections for multiple sentences in a single figure is computationally expensive.
In order to overcome that, we took steps to reduce the computational complexity while preserving as much of the overall shape of the point cloud as possible.
These included replacing ball like blobs of points with their center of mass, and performing a dimensionality reduction by PCA.
The first step does not affect homotopy type much in most cases, because dense, ball like, clusters of points that are distributed far from one another in the embedding space would be collapsed to single nodes in the mapping approximation anyways, and Gaussian blobs have no interesting topology (they are approximate to solid balls, which are contractible and have trivial homology groups).
This saves a lot of computation by reducing the number of vectors in the cloud.
Furthermore, since we are already projecting onto a graph, we can first project onto the top principal components first, without changing the topology of the graph significantly.
Due to these optimizations we were able to visualize entire articles from the corpus in a single picture.
Figure \ref{hidemb_100_mapper} shows a section of the corpus composed of more than 2600 words.
Again, we see that the hidden state projection is significantly more complex than the embeddings.

\begin{figure}
	\centering
	\includegraphics[width=0.19\textwidth]{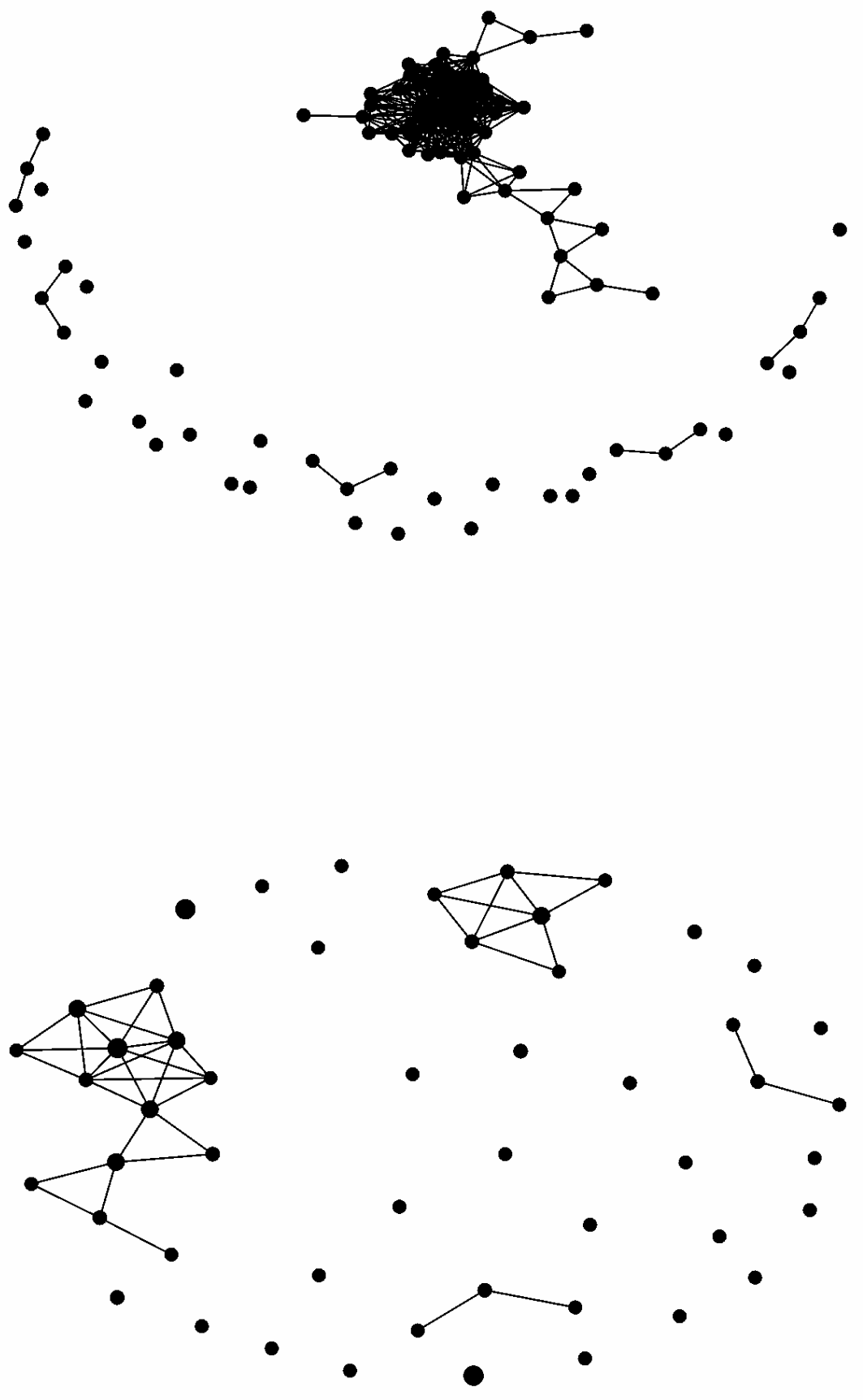}
	\includegraphics[width=0.19\textwidth]{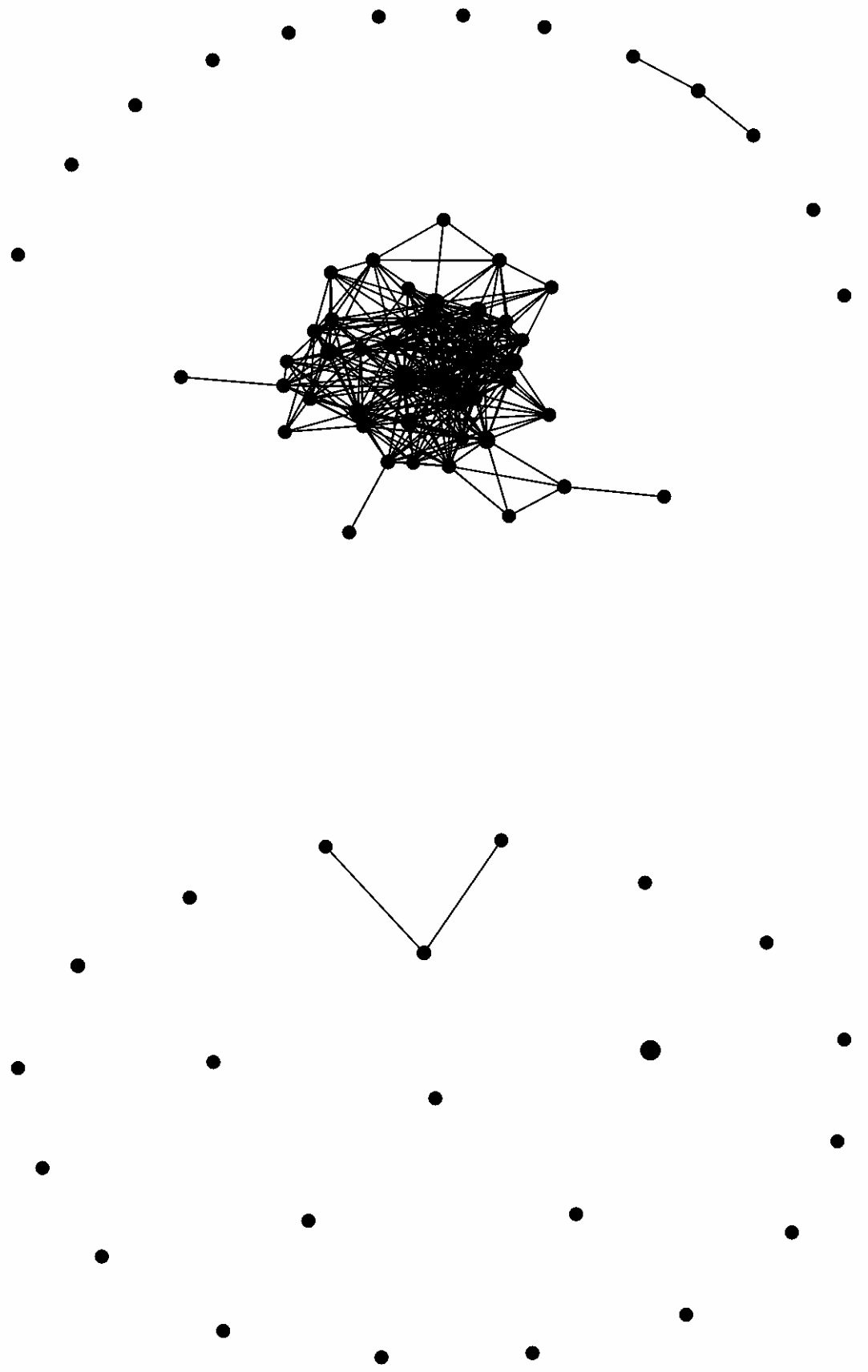}
	\includegraphics[width=0.19\textwidth]{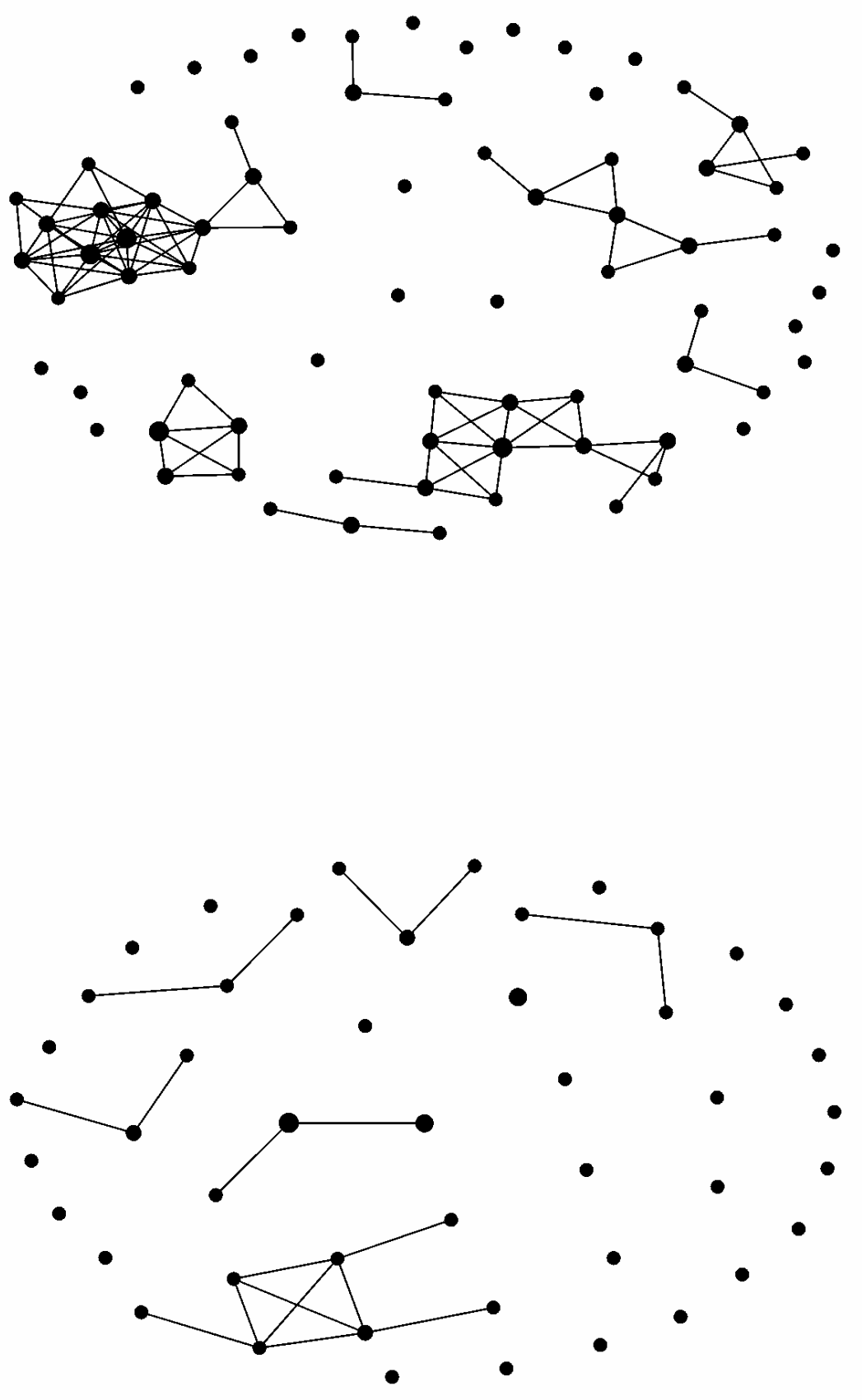}
	\includegraphics[width=0.19\textwidth]{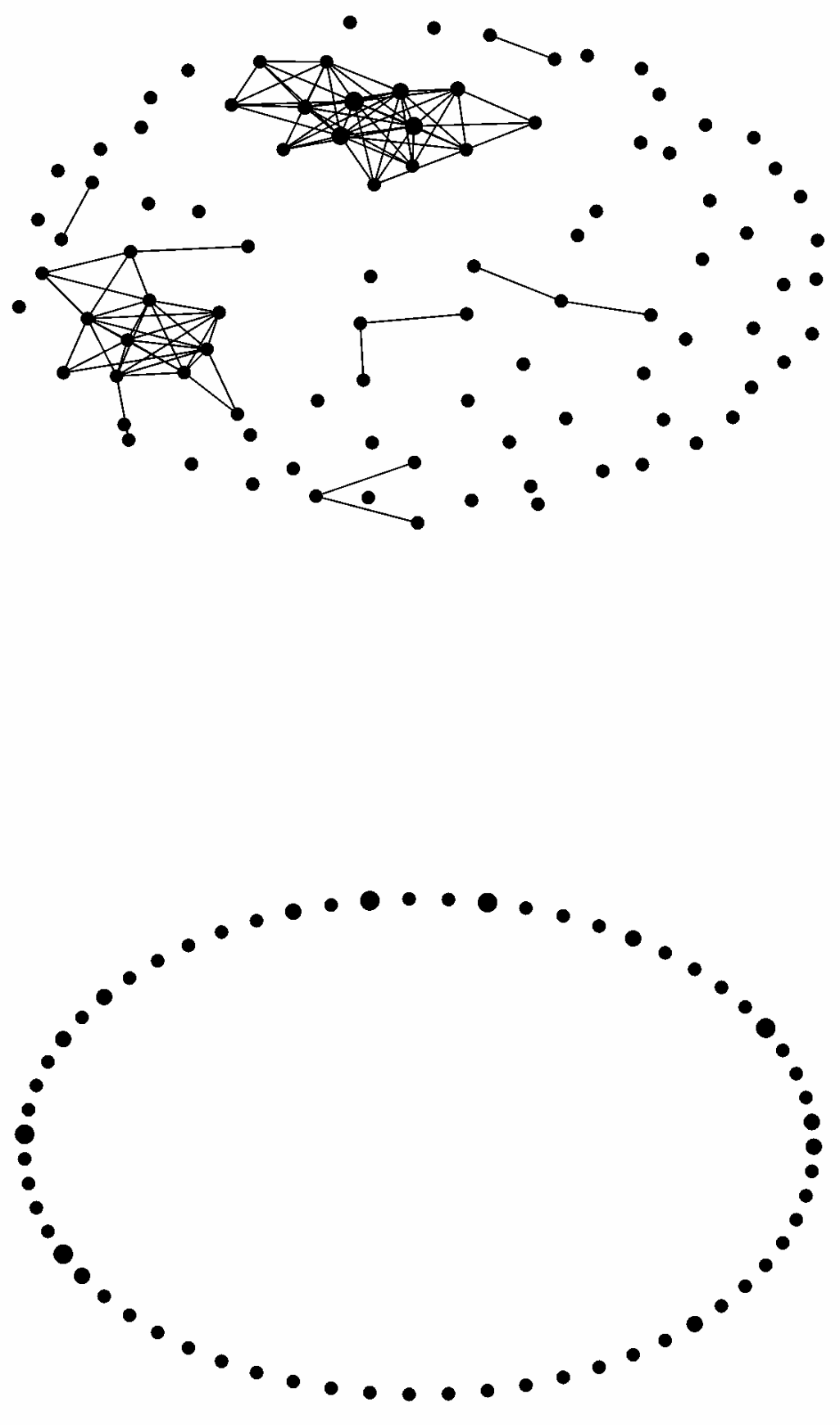}
	\includegraphics[width=0.19\textwidth]{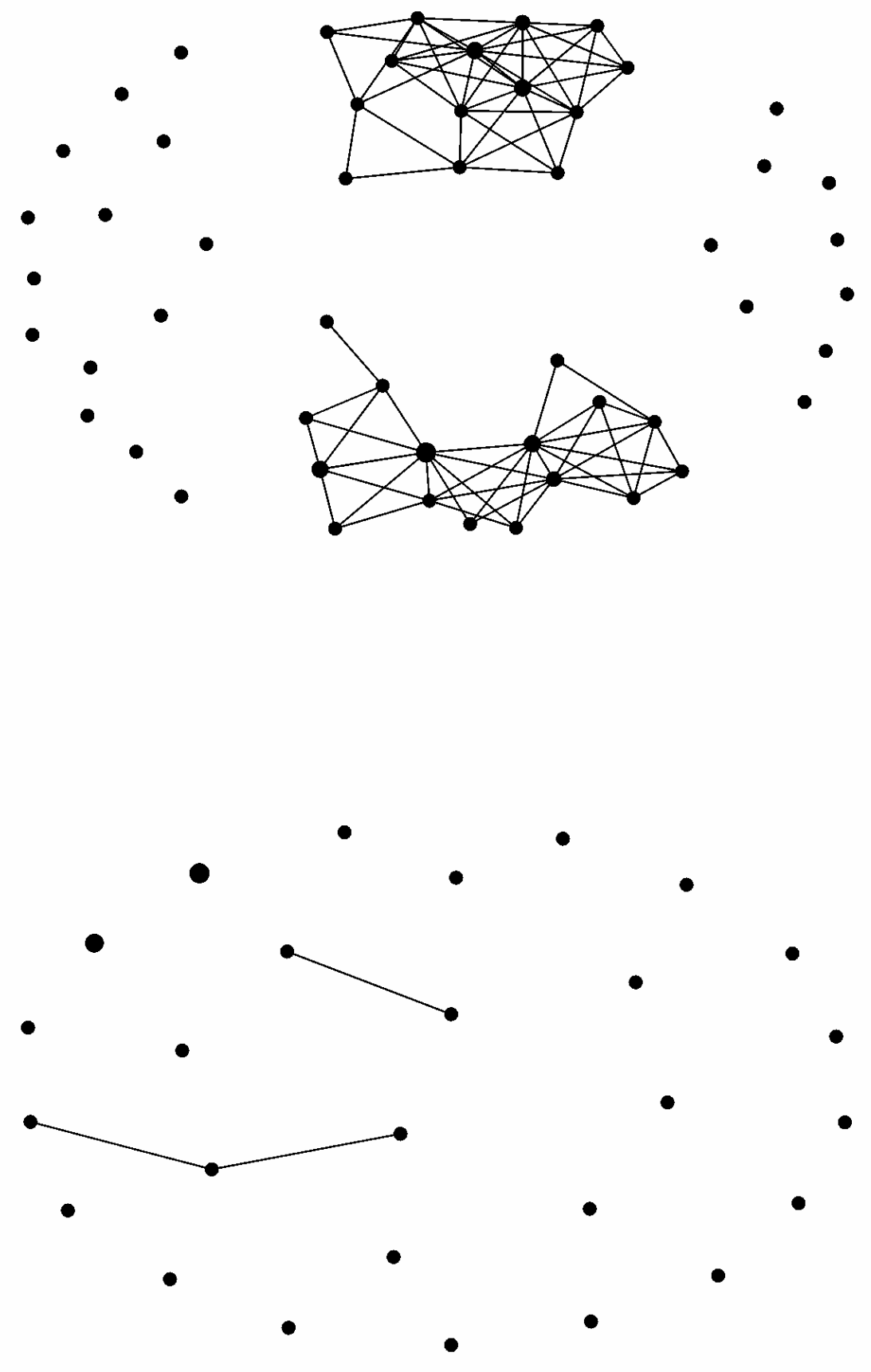}
  \caption{Five random paragraphs of English text represented by the LSTM model. Hidden states are shown above, embeddings are below.}
	\label{hidemb_mapper}
\end{figure}

\subsection{Sliding Window Embedding}

Finally, we take yet another look at our representations.
In the previous experiments, we looked at each sentence as a point cloud within the representation space of our neural language model.
Although, the order of words within each sentence is implicitly captured by the structure of the point cloud (because of the way word vectors are induced by the LM), we did not explicitly take it into consideration when computing topological features.
In this section we take the ordering of the embeddings directly into account by performing a re-representation step designed to model time series data.
This allows us to study homological properties of each dimension within the representation manifold of our language model.
When words from a corpus are fed into the neural network implementation of the language model, its hidden state vector traces out a path in the embedding space.
We can interpret topological properties of these paths, and their relationship to corpus data, by analyzing each dimension of the hidden state vector as a time series.
Every sentence of the corpus generates multiple sequences of floating point numbers - one in each dimension of the representation manifold.
We can transform those sequences, into topological objects, and study a notion of shape for each factor of the word embedding.
In order to do this, we slide a window over the time series of the hidden states associated to the LM, and compute topological invariants of the resulting point clouds (see figure \ref{sliding_window_embedding} for an illustration of the idea).


The first step is the construction of the \emph{sliding window embedding}.
This step depends on two parameters: $\tau$ for the delay and $d$ for the dimension.
Let $f_i(t)$ be the value of the $i$-th component of the hidden state vector in our language model, after $t$ words of the sentence being analyzed were consumed by it.
We collect the values $f_i(t), f_i(t+\tau), \cdots, f_i(t + (d - 1) \times \tau)$, which results in a vector of $d$ values.

$$
SW_{d,\tau}f_i(t) = \begin{bmatrix}
  f_i(t) \\
  f_i(t+\tau ) \\
  \vdots \\
  f_i(t+(d-1)\tau)
\end{bmatrix} \in \mathbb{R}^{d}
$$

The dimension $d$ in our case corresponds to the n-gram size chosen.
For instance, if we look at a 3-gram model, we would slide a window of 3 values over representations of each sentence.
We do this in each dimension separately.
We then analyze the collection of such vectors obtained from each sentence of the corpus.
Thus sentence with $w$ words, embedded into a $d$-dimensional manifold by the neural language model, analyzed using n-gram size of $n$, will produce $d$, $n$-dimensional point clouds.
We then analyze each one of these point clouds as a sample from an underlying topological manifold using techniques from the previous sections.
That is we compute Vietoris-Rips filtrations and their homology.
This produces topological summaries of each dimension in our representation space.

\begin{figure}
	\centering
  \includegraphics[width=0.6\textwidth]{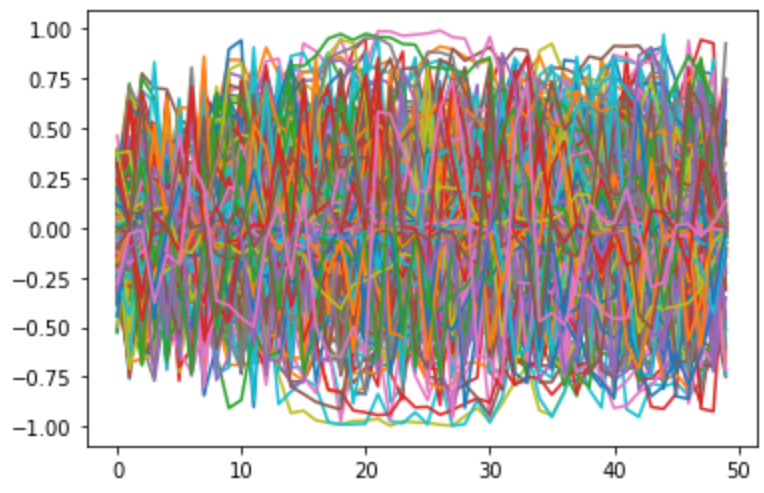}
  \caption{Trajectories of the hidden states in 450 dimensional representation manifold over a sentence composed of 50 words (LSTM model).}
	\label{swe_alldims}
\end{figure}

\begin{figure}
	\centering
  \includegraphics[width=0.34\textwidth]{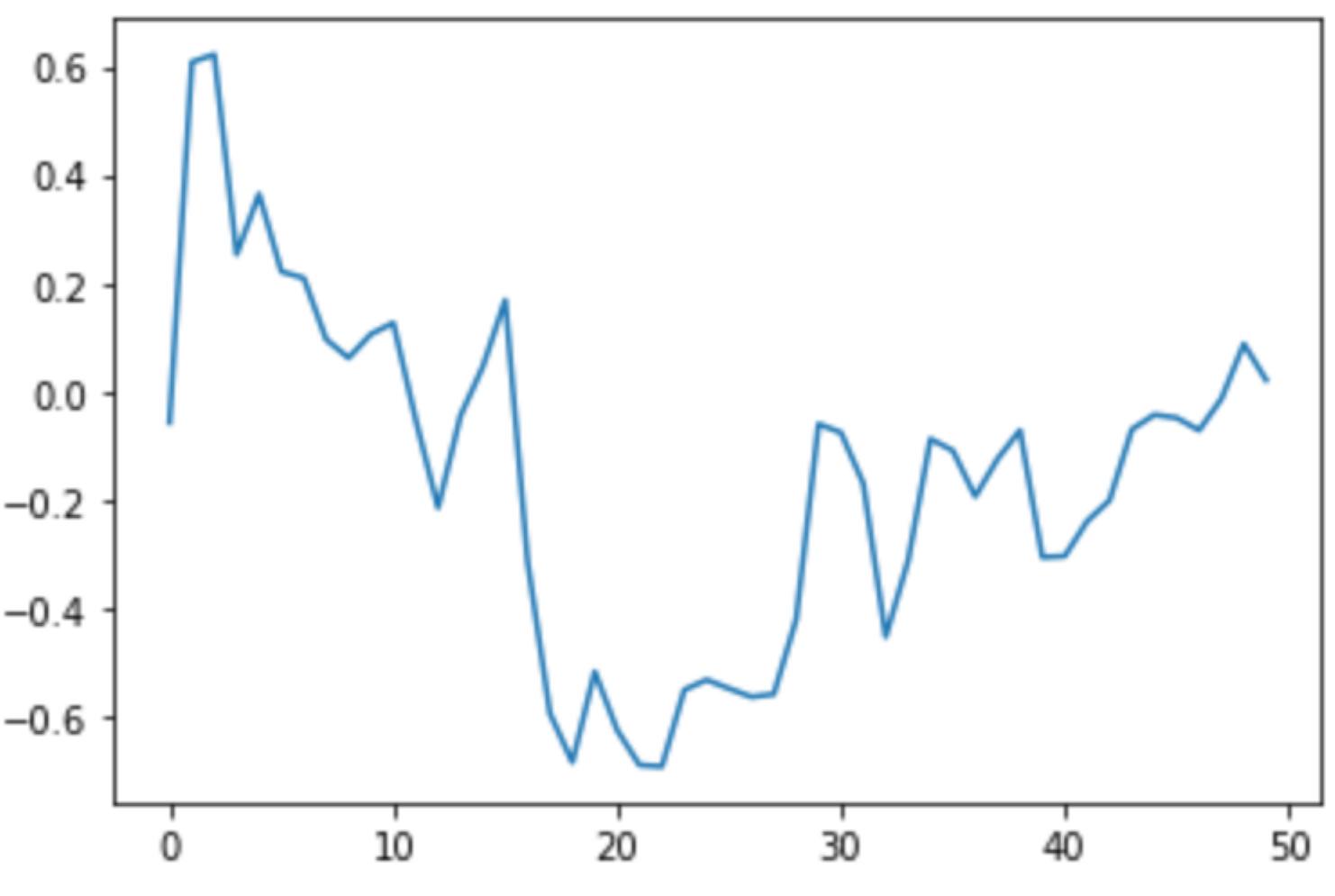}
	\includegraphics[width=0.35\textwidth]{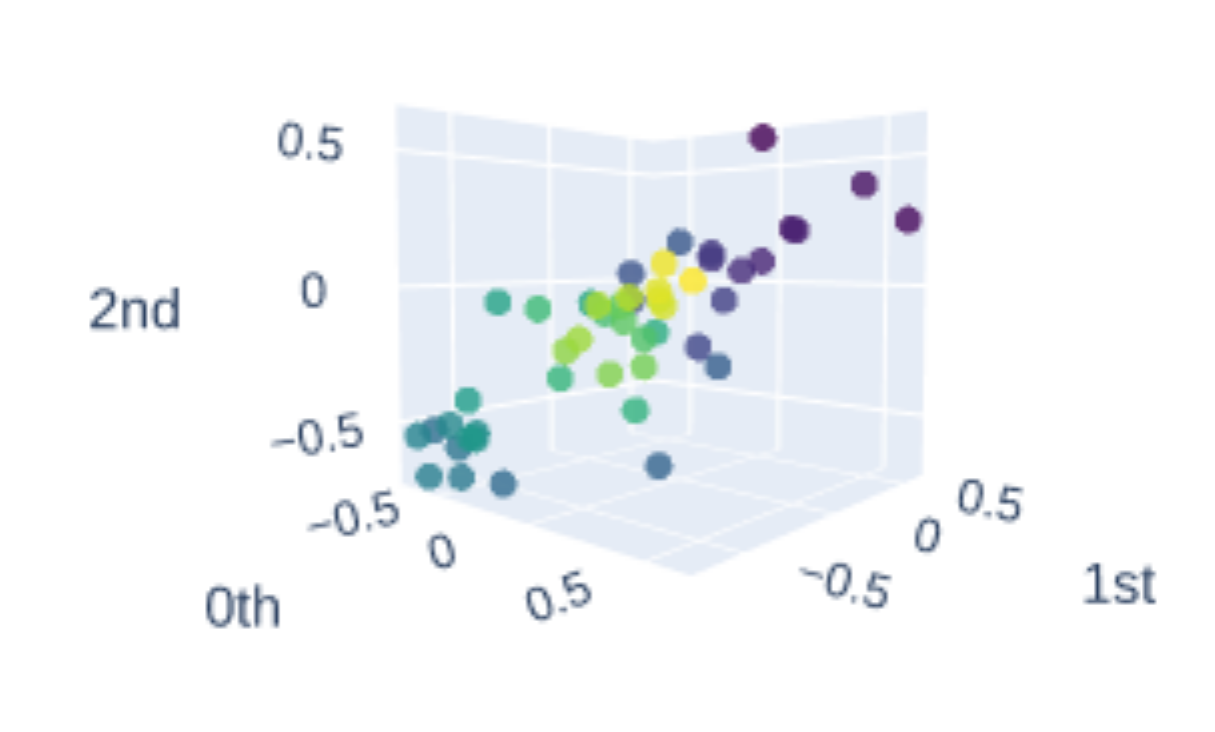}
	\includegraphics[width=0.29\textwidth]{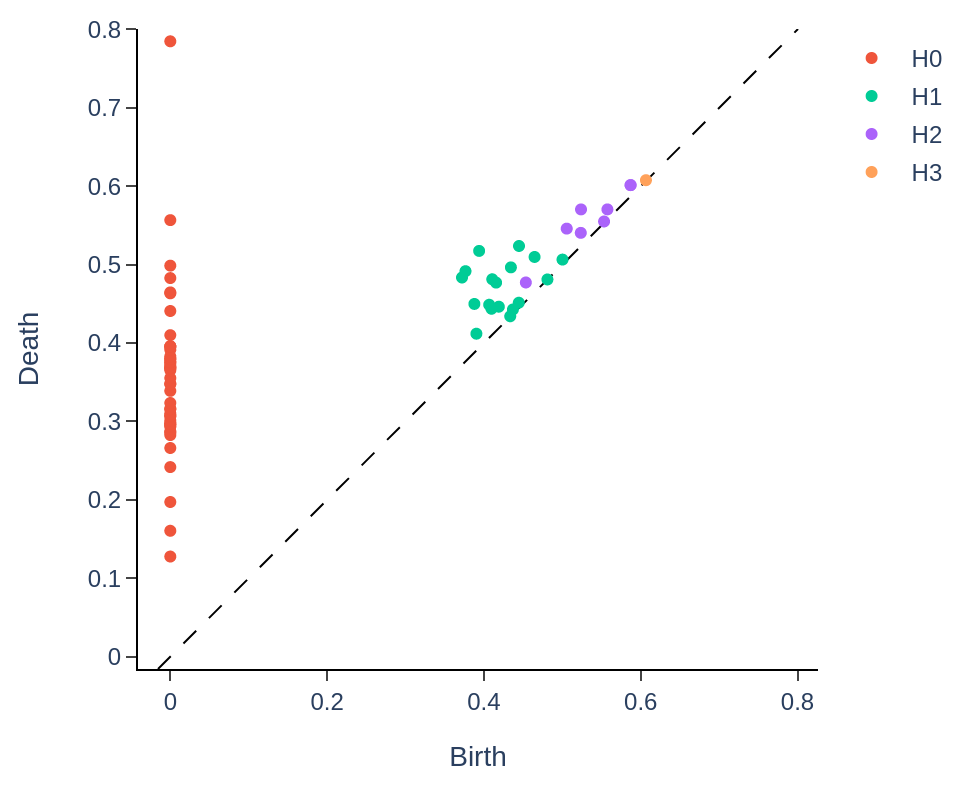}
  \caption{Left: visualization of the trajectory of hidden state of the LSTM in dimension 42 of the representation manifold over a sentence composed of 50 words. Center: sliding window embedding of that data using a 3-gram model of the sentence. Right: persistence diagram of this hidden state trajectory.}
	\label{swe_onedim}
\end{figure}

We analyzed the trajectories of the hidden states in each dimension of the representation manifold.
Looking at the raw data directly, it is hard to observe any patterns (see figure \ref{swe_alldims}).
However, topological representation allows us to summarize the behavior of these states in a way similar to the previously described approaches.
Figure \ref{swe_onedim} shows an example of a single dimension.

We applied this same process in all dimensions, and performed comparisons between global and contextualized representations, as well as analysis of state evolution over training epochs.
The results confirm the observations of the previous three sections. More precisely, in the LSTM models hidden states had significantly higher average topological complexity over all dimensions than that of the input embeddings. Furthermore, topological complexity increased during training with more topological features appearing with lower perplexity.

\newpage
\section{Additional Plots}

\begin{figure}[!h]
	\centering
	\includegraphics[width=0.22\textwidth]{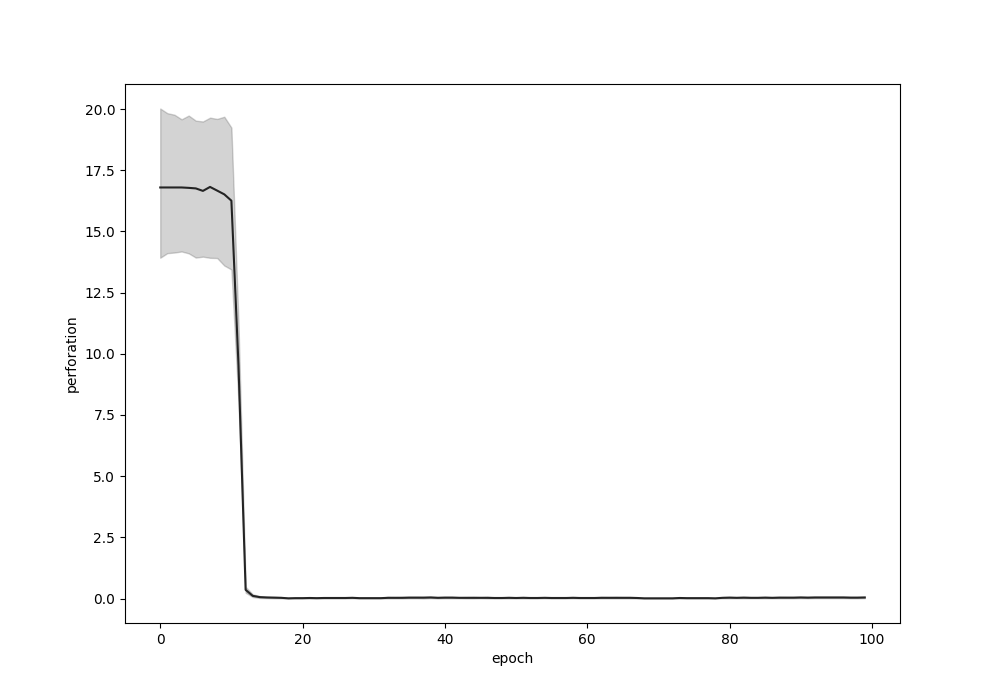}
	\includegraphics[width=0.22\textwidth]{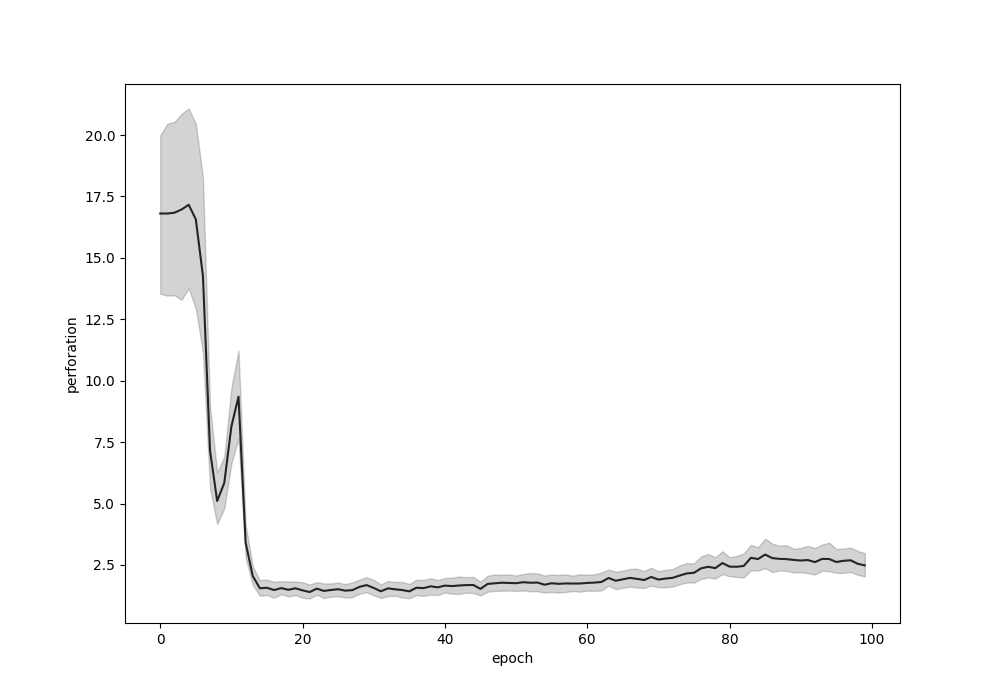}
	\includegraphics[width=0.22\textwidth]{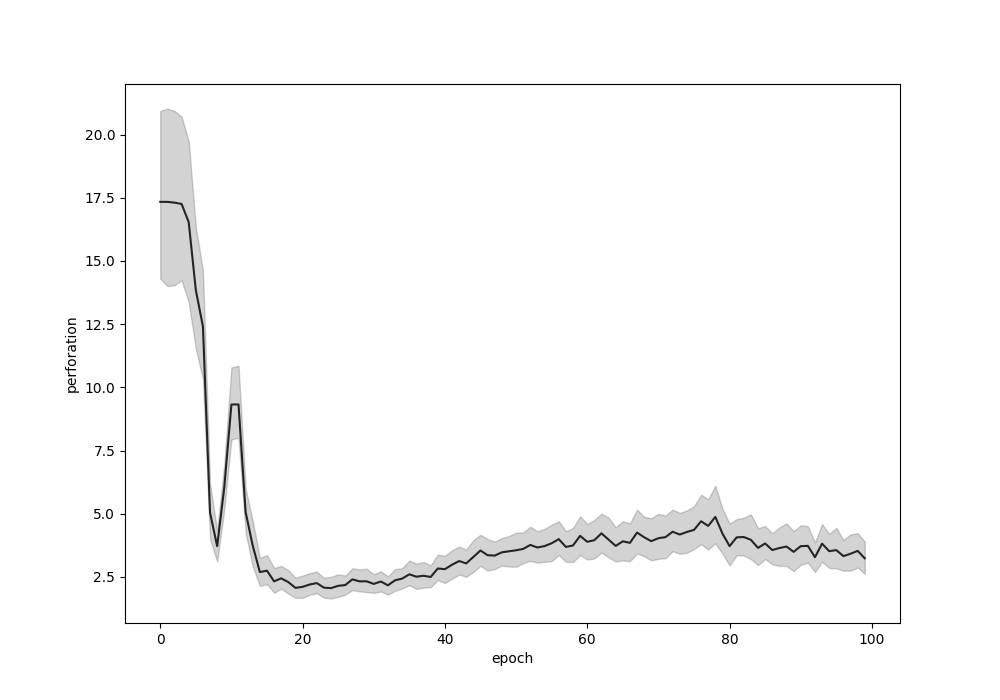}
	\includegraphics[width=0.22\textwidth]{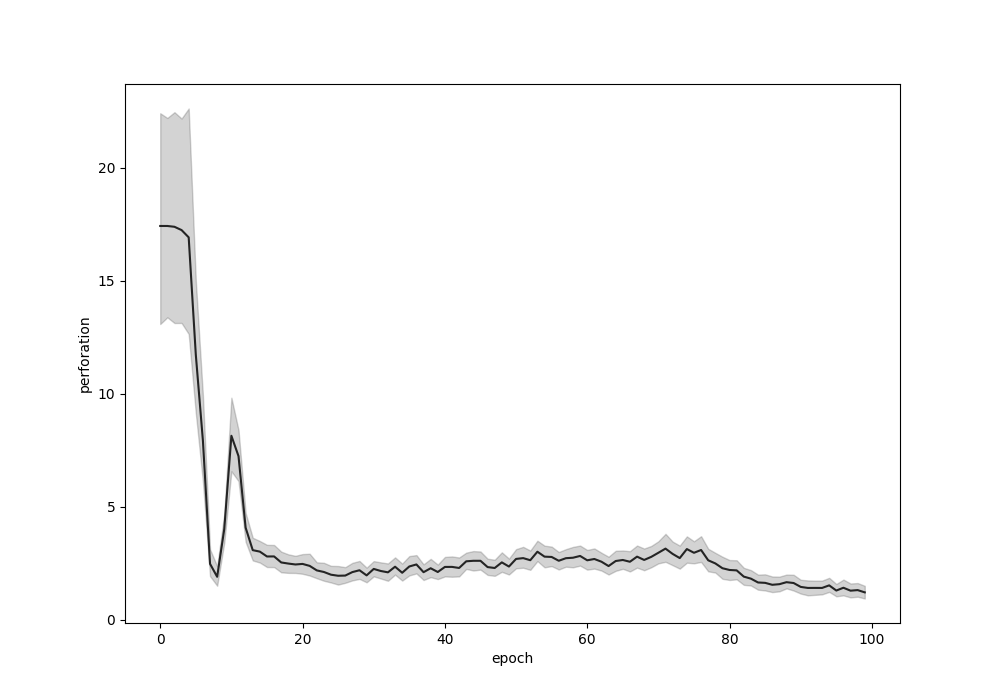}
  \caption{Perforation over epochs of training using a sample of English sentences. The embedding layer and transformer blocks 4, 8, 12 of a 160m GPT model (Pythia) (from left to right).}
	\label{pythia_perf_additional}
\end{figure}

\begin{figure}[!h]
	\centering
	\includegraphics[width=0.22\textwidth]{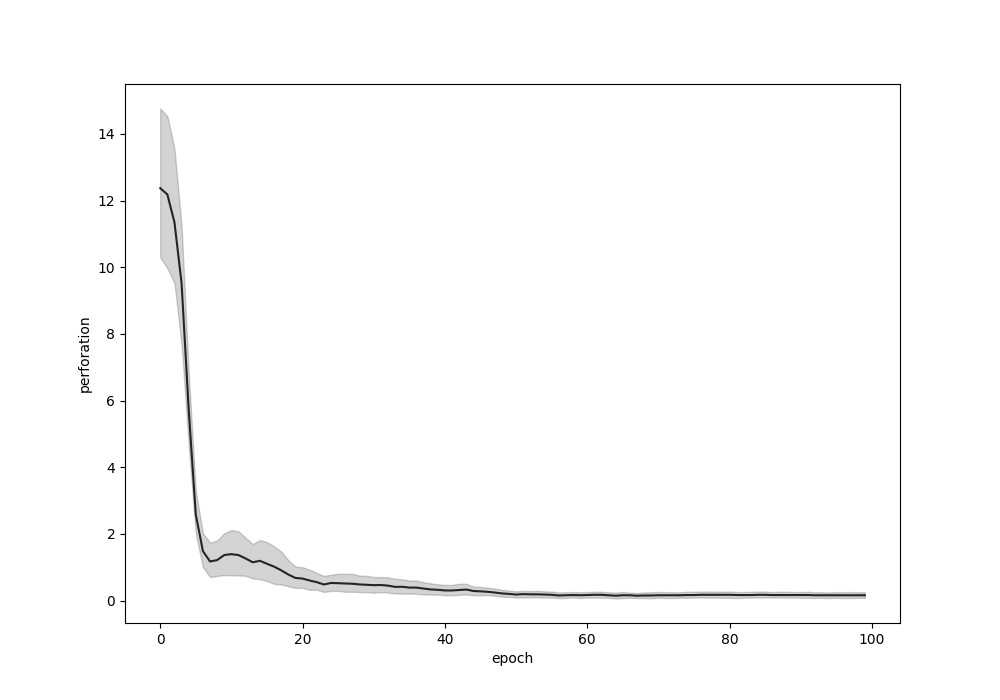}
	\includegraphics[width=0.22\textwidth]{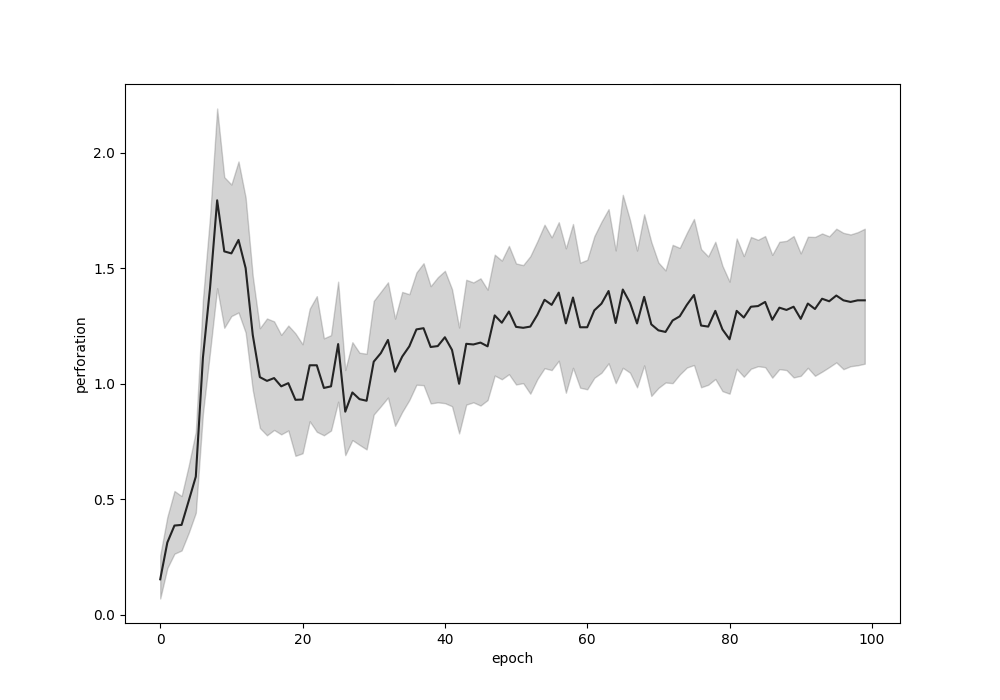}
	\includegraphics[width=0.22\textwidth]{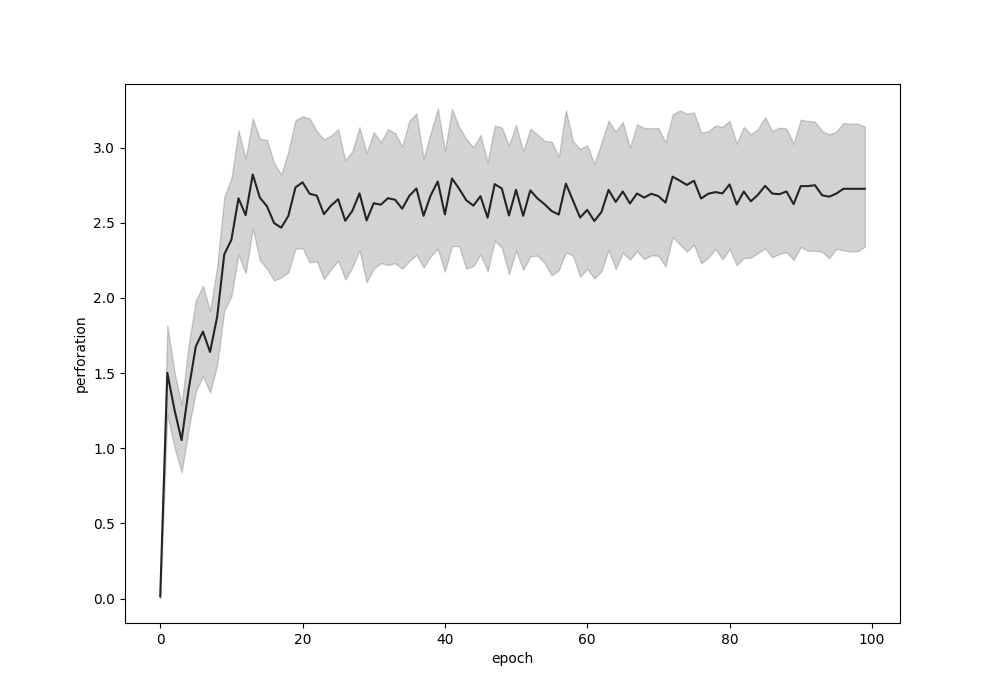}
	\includegraphics[width=0.22\textwidth]{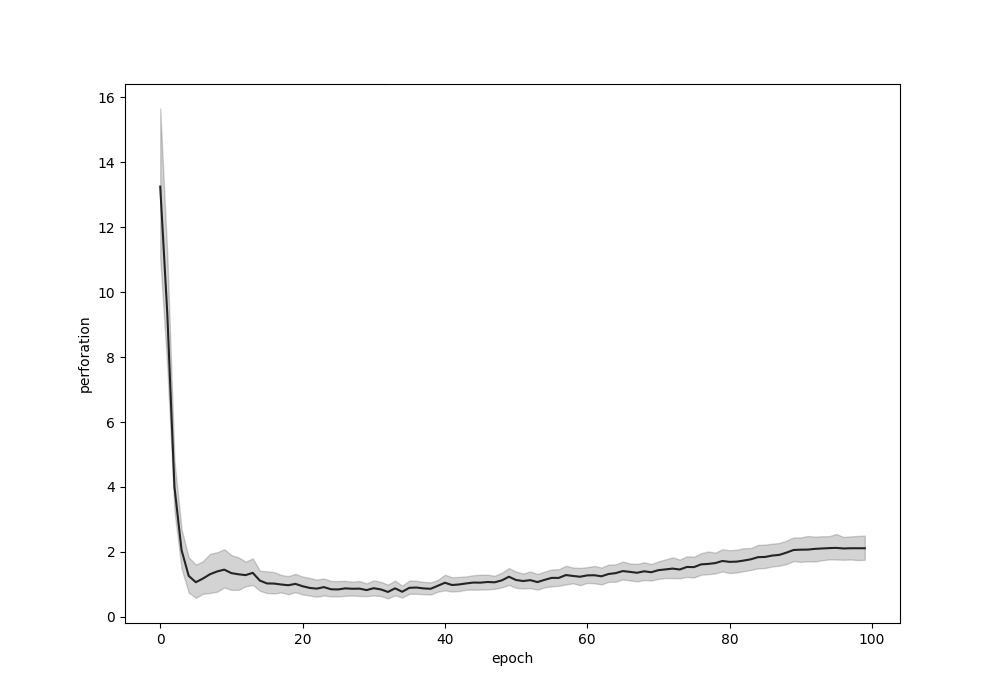}
	\includegraphics[width=0.22\textwidth]{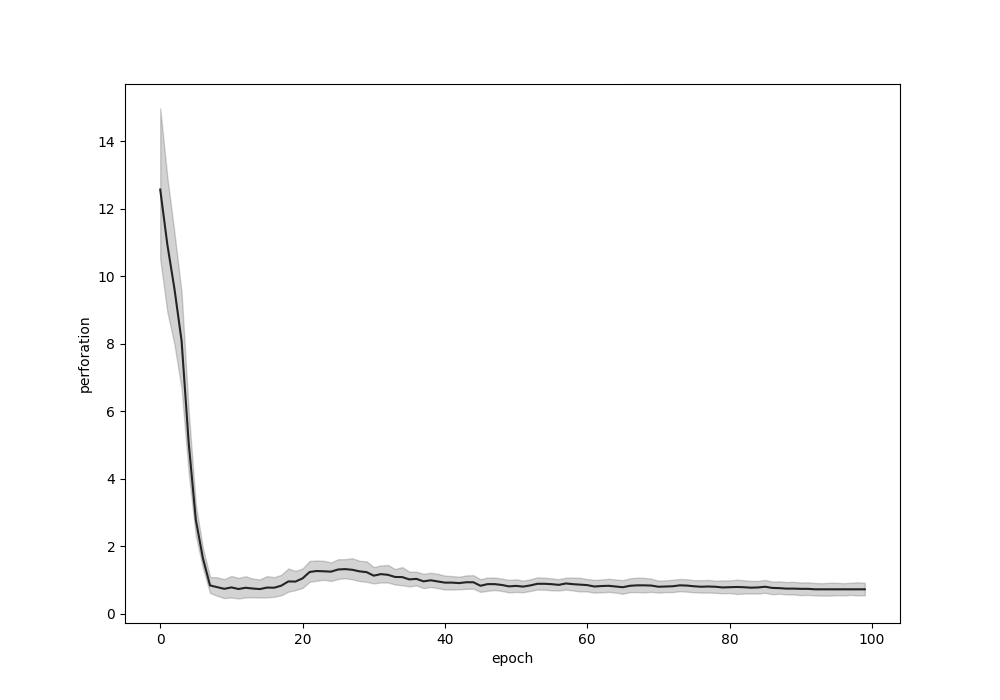}
	\includegraphics[width=0.22\textwidth]{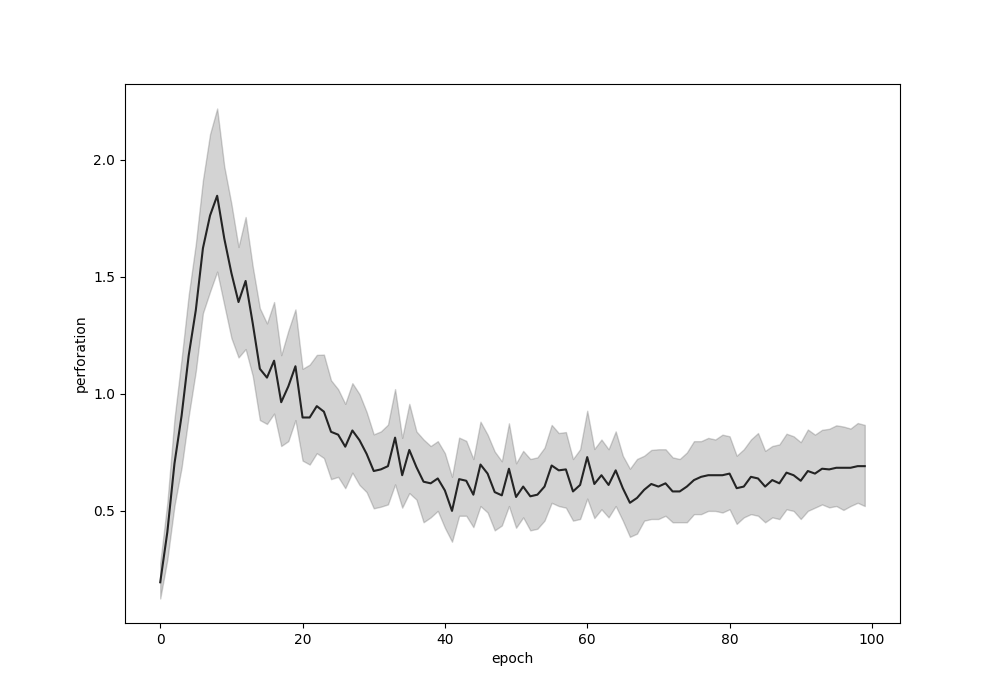}
	\includegraphics[width=0.22\textwidth]{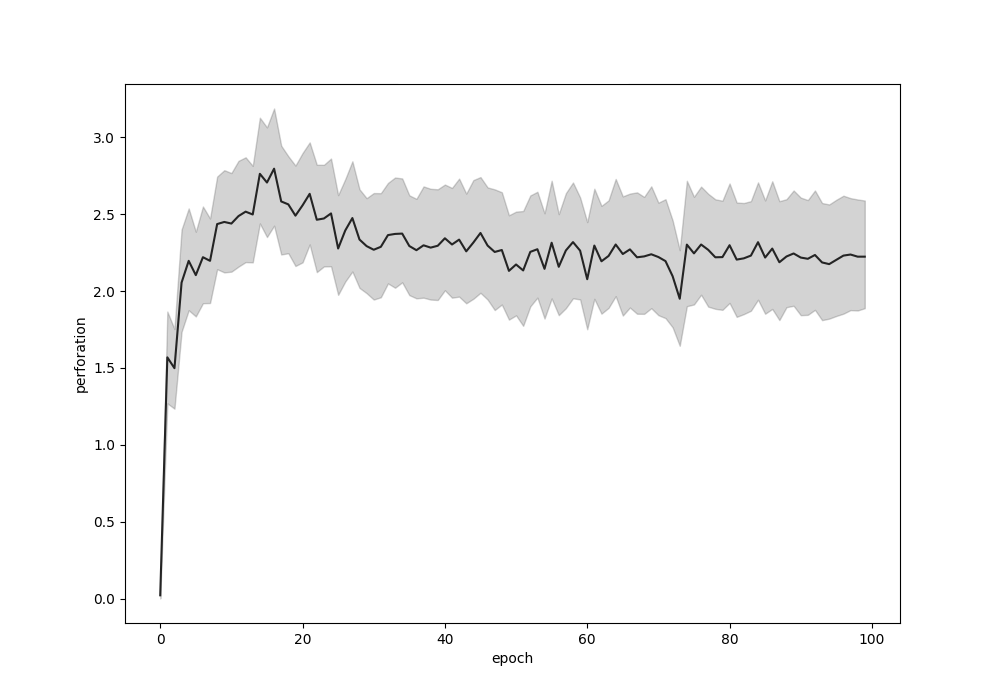}
	\includegraphics[width=0.22\textwidth]{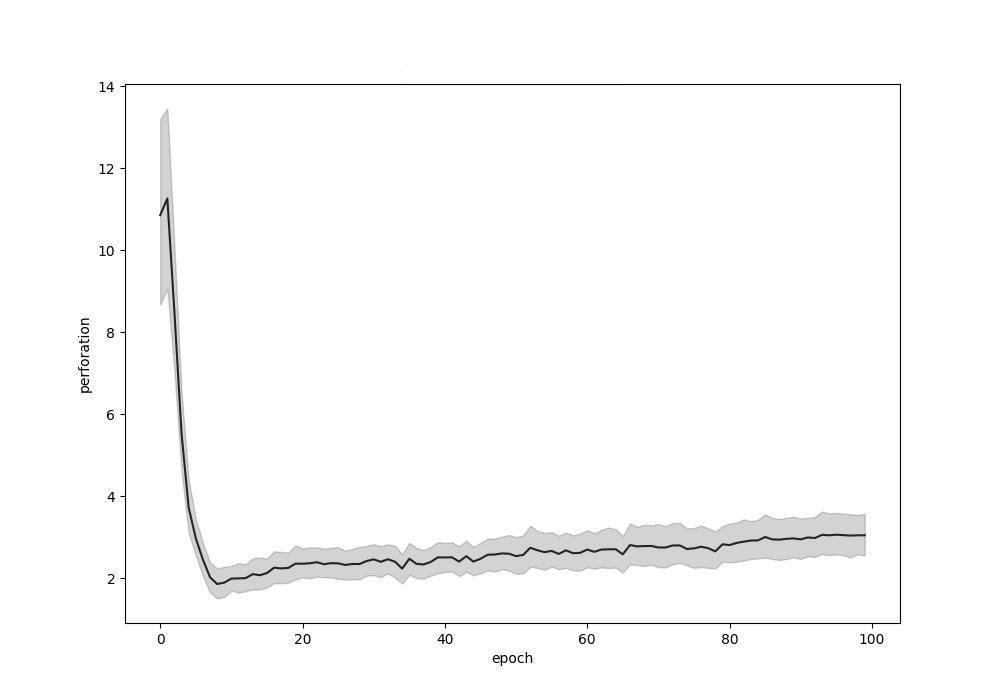}
	\includegraphics[width=0.22\textwidth]{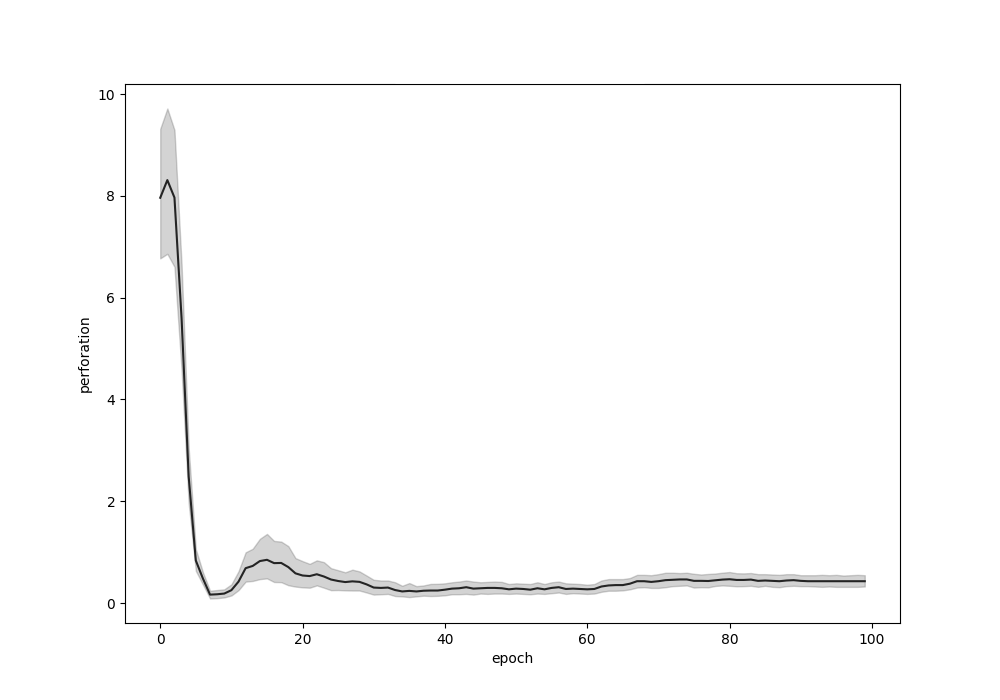}
	\includegraphics[width=0.22\textwidth]{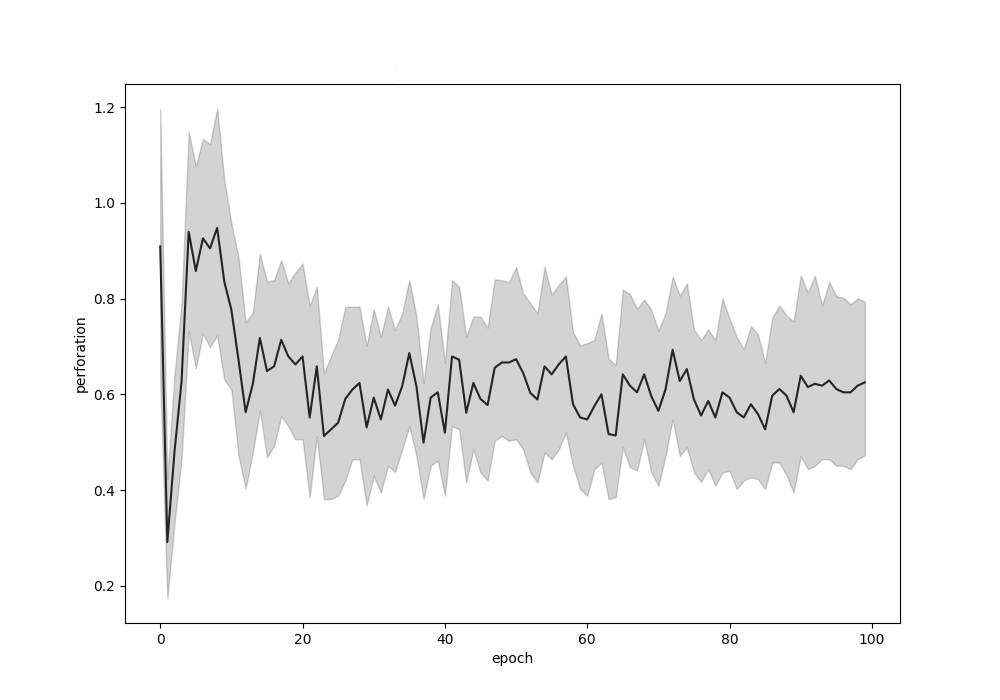}
	\includegraphics[width=0.22\textwidth]{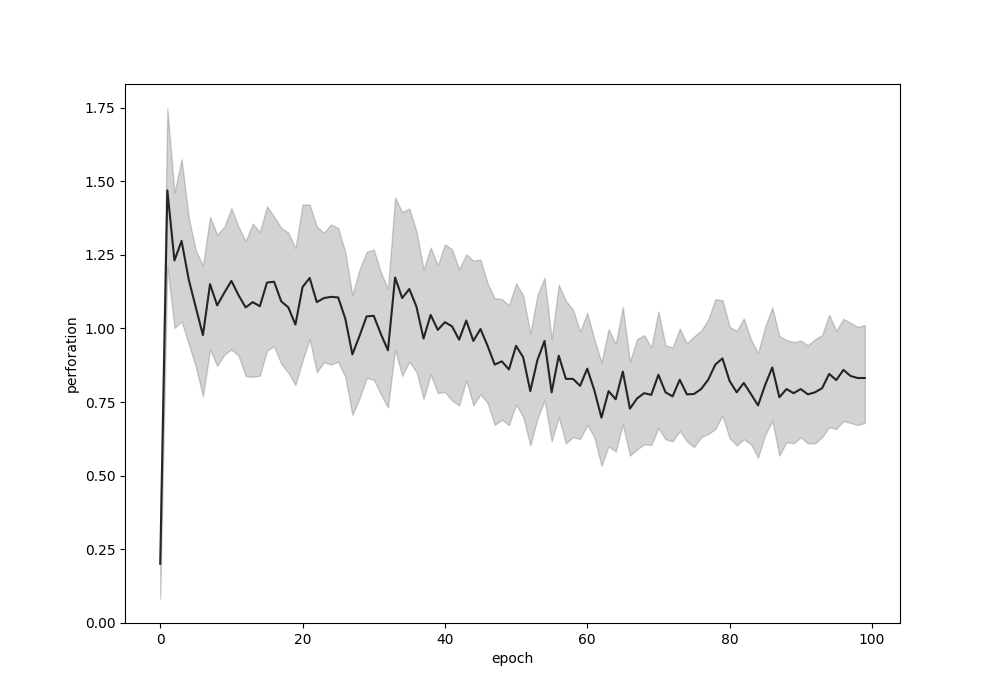}
	\includegraphics[width=0.22\textwidth]{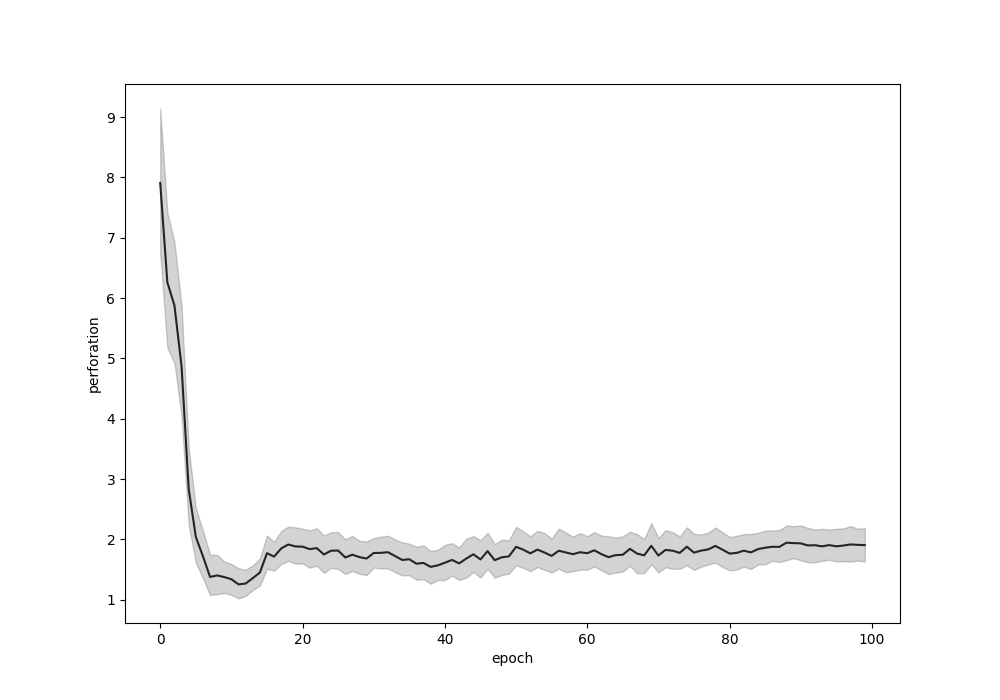}
	\includegraphics[width=0.22\textwidth]{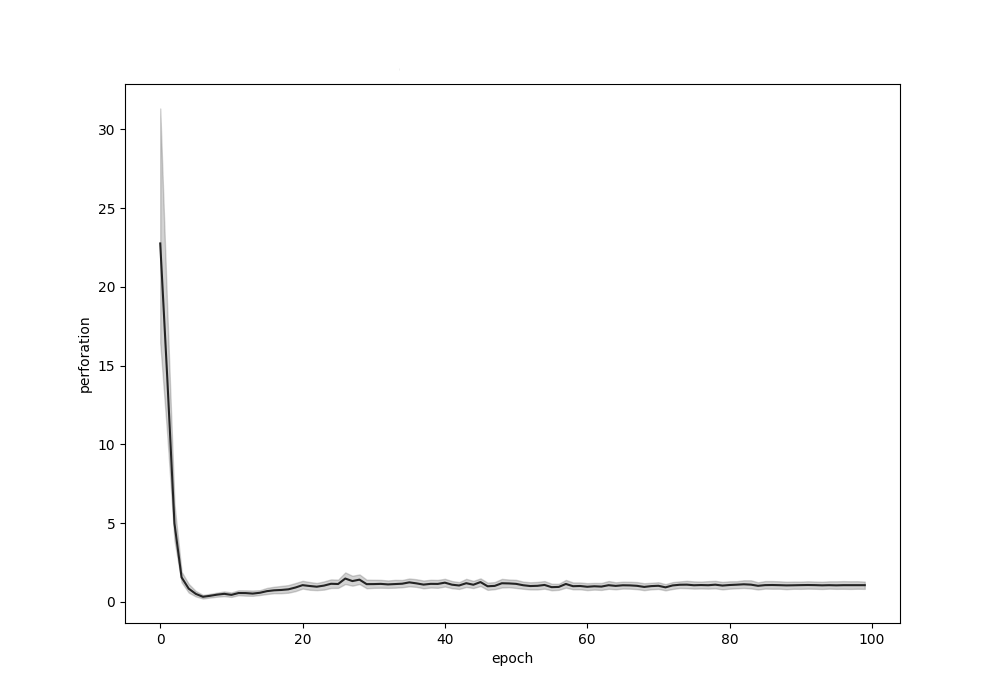}
	\includegraphics[width=0.22\textwidth]{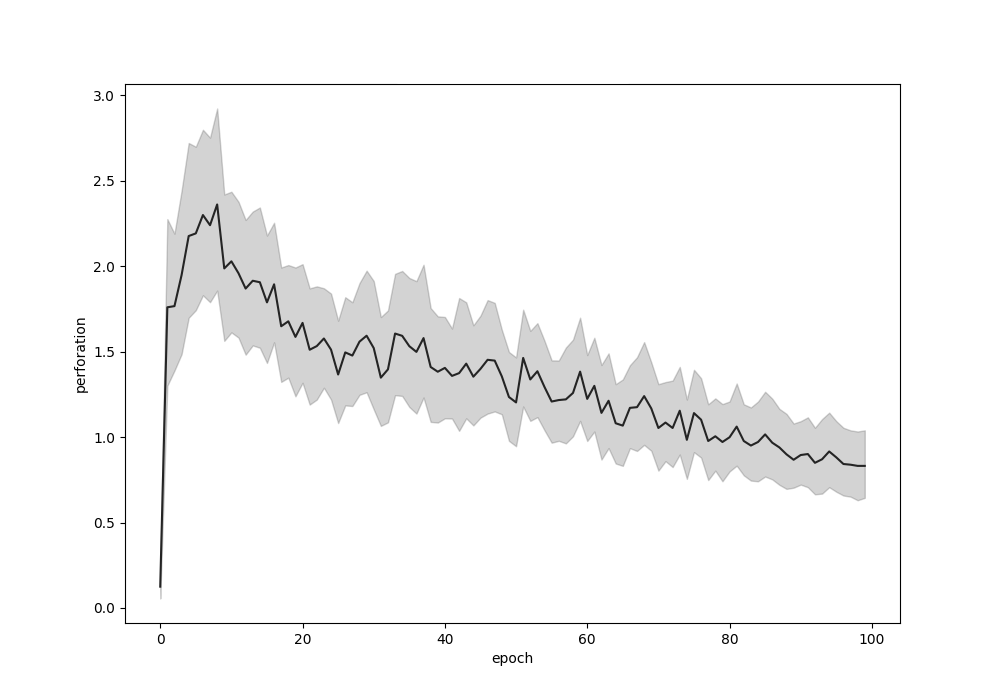}
	\includegraphics[width=0.22\textwidth]{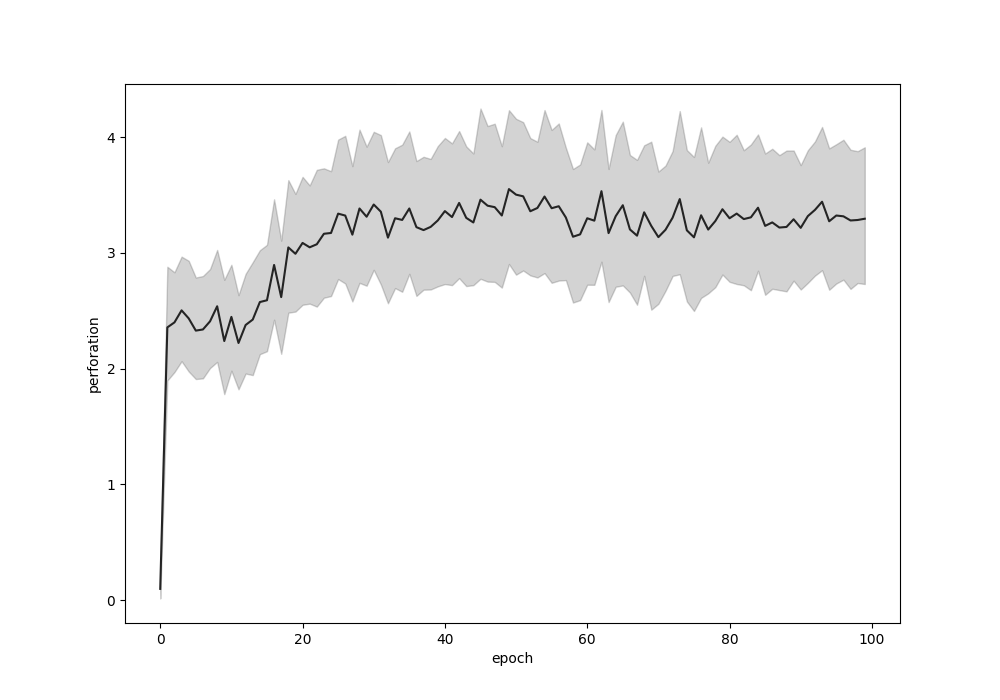}
	\includegraphics[width=0.22\textwidth]{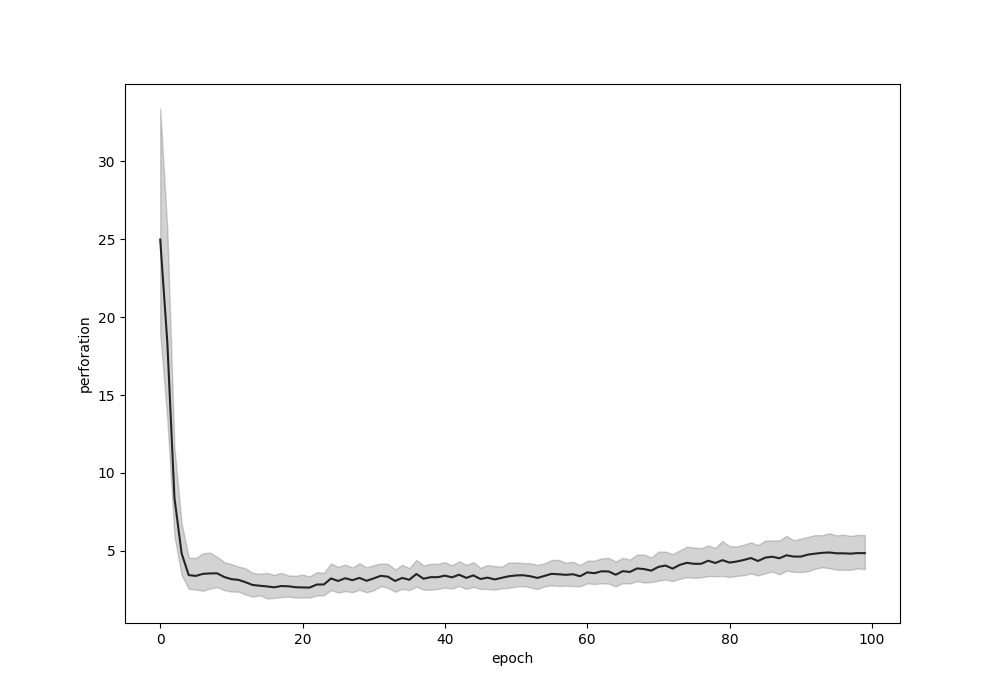}
	\includegraphics[width=0.22\textwidth]{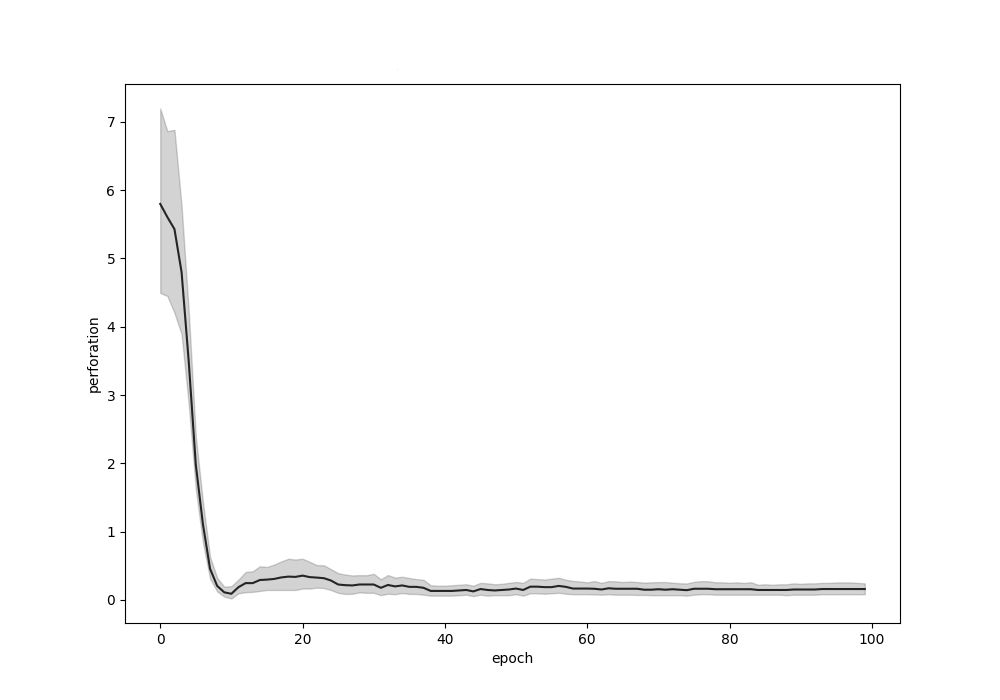}
	\includegraphics[width=0.22\textwidth]{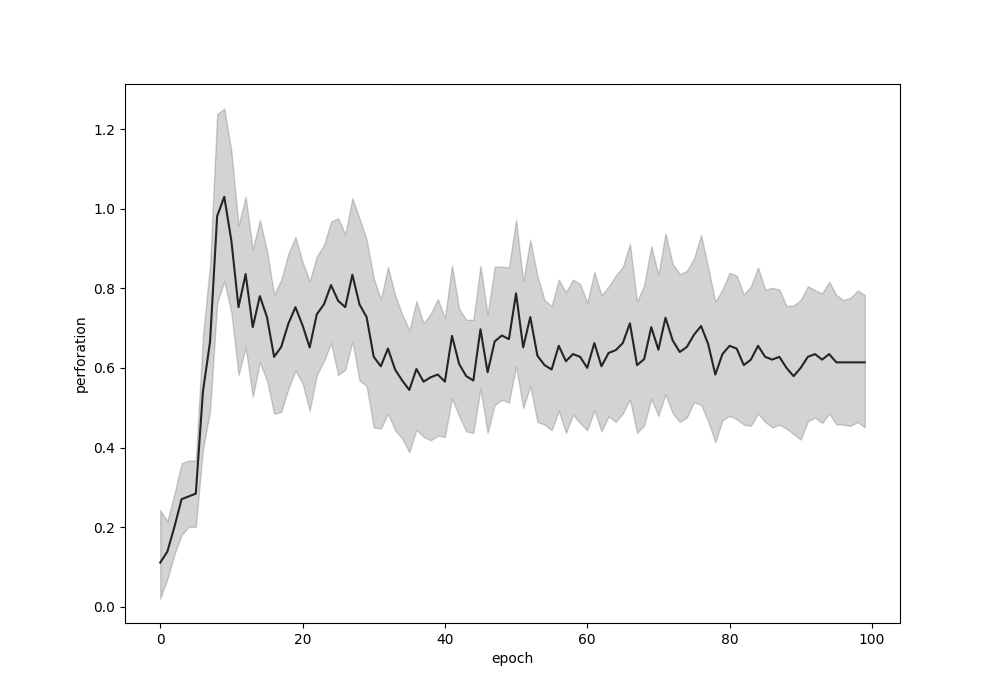}
	\includegraphics[width=0.22\textwidth]{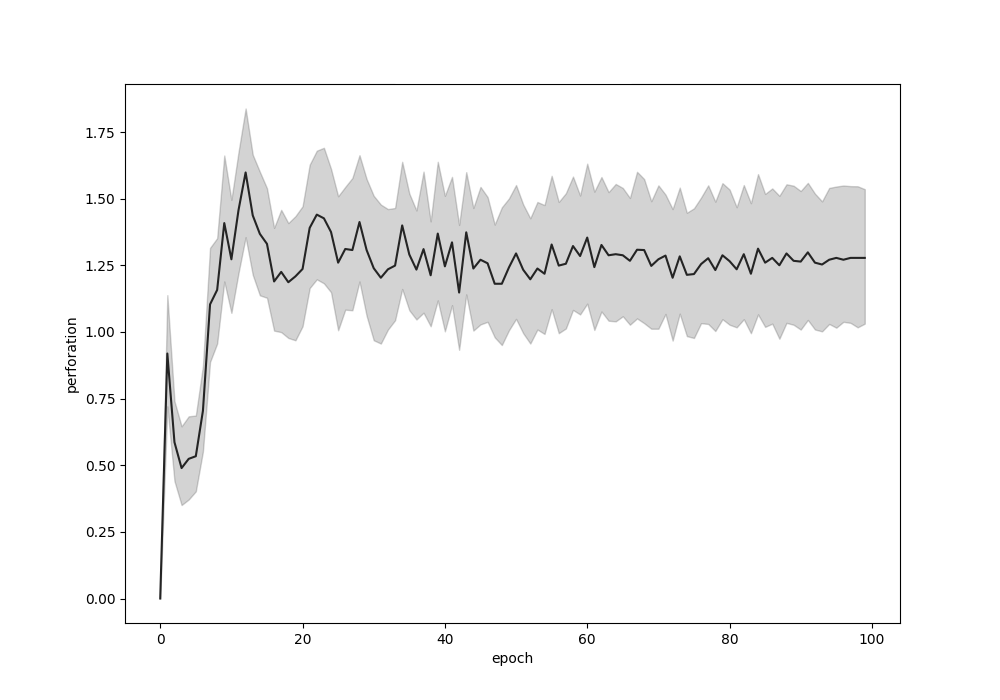}
	\includegraphics[width=0.22\textwidth]{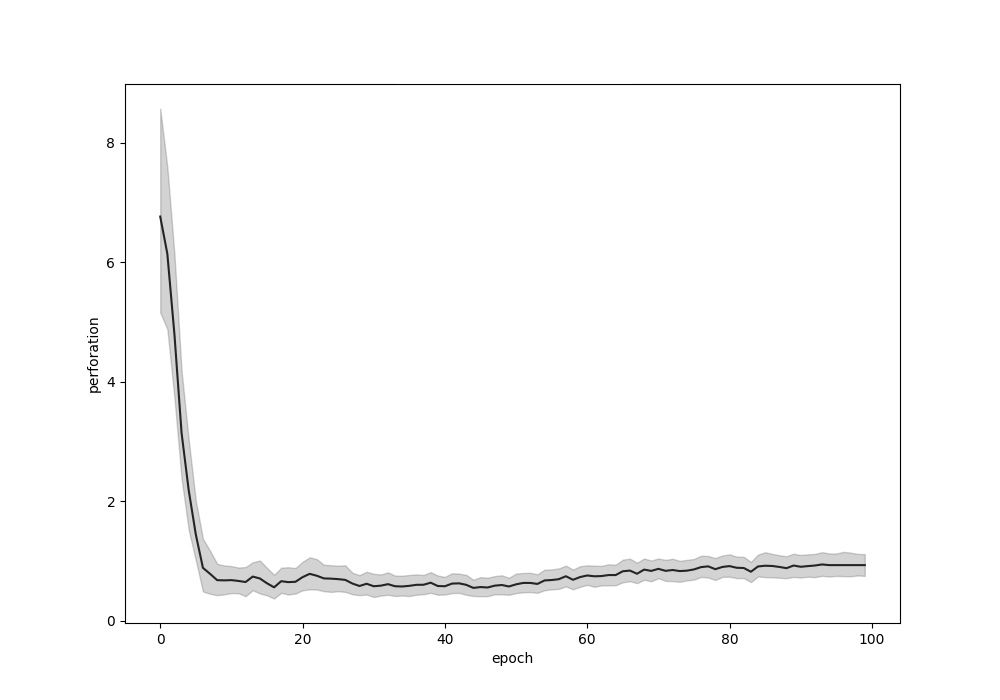}
  \caption{Perforation over epochs of training. Columns are the CNN model layers from embedding (left) to output (right). The rows from top to bottom are: Arabic, French, German, Japanese, Russian.}
	\label{cnn_perf_additional}
\end{figure}

\begin{figure}
	\centering
	\includegraphics[width=0.98\textwidth]{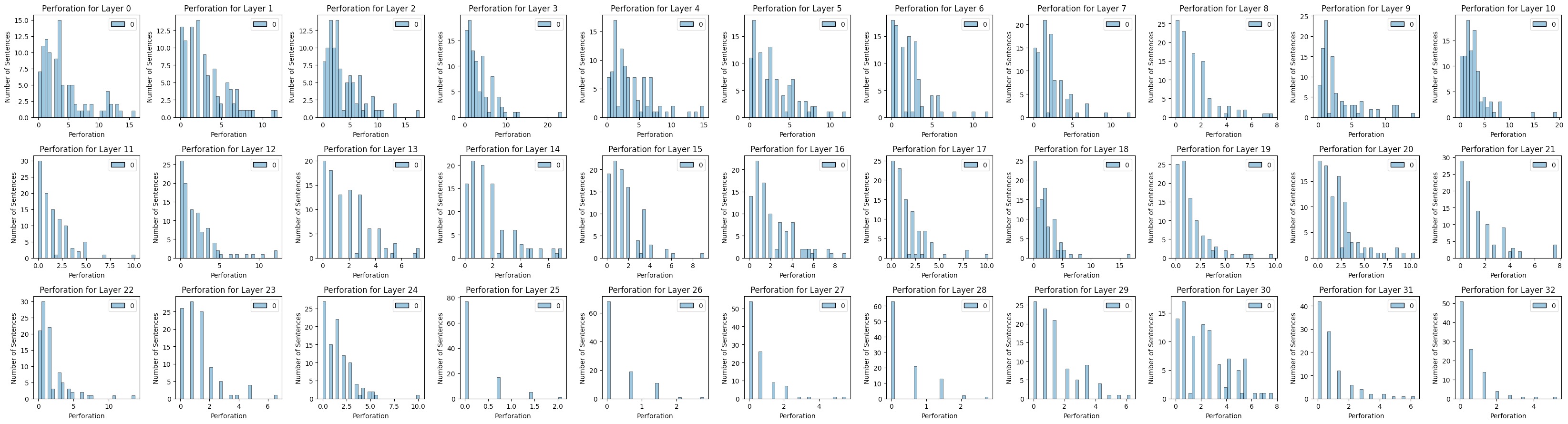}
  \caption{Perforation histograms for 33 transformer layers of LLaMA (fully trained). Synthetic Zipf corpus.}
	\label{llama_zipf}
\end{figure}

\begin{figure}
	\centering
	\includegraphics[width=0.98\textwidth]{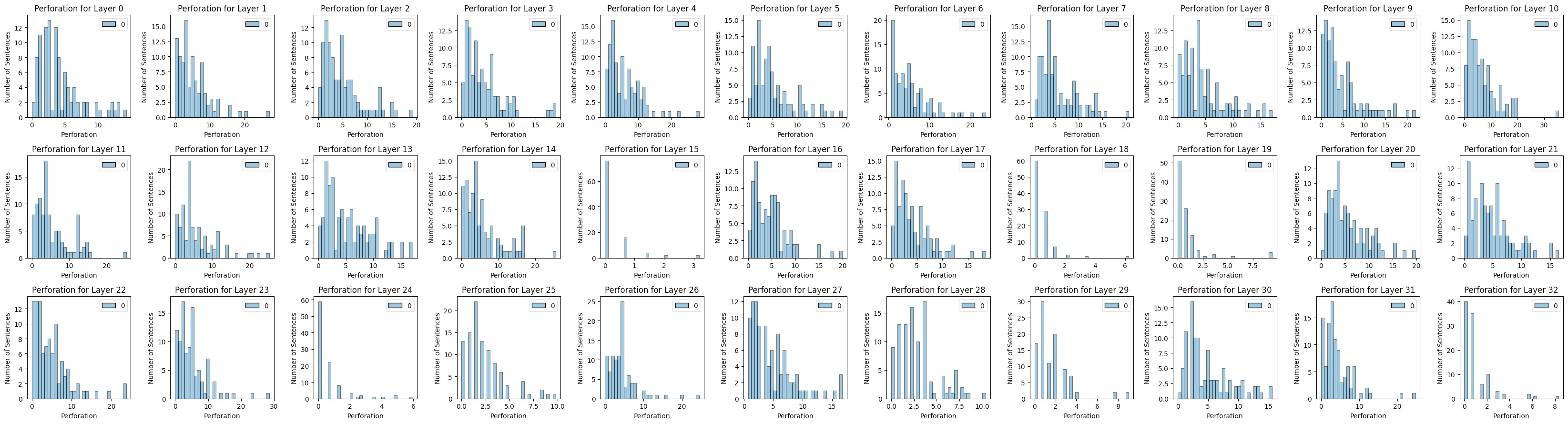}
  \caption{Perforation histograms for 33 transformer layers of LLaMA (fully trained). Natural English corpus.}
	\label{llama_eng}
\end{figure}

\begin{figure}
	\centering
	\includegraphics[width=0.27\textwidth]{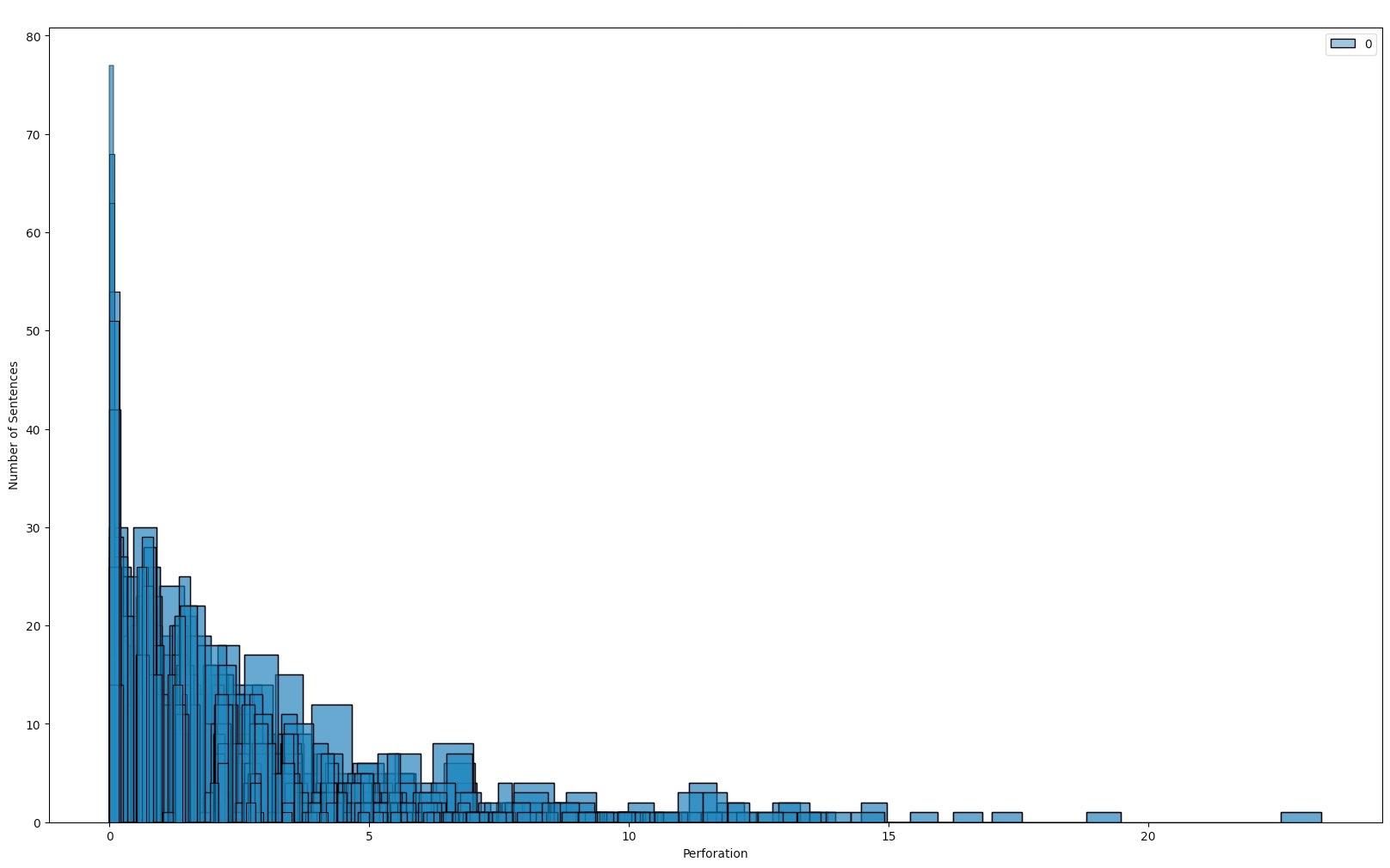}
	\includegraphics[width=0.22\textwidth]{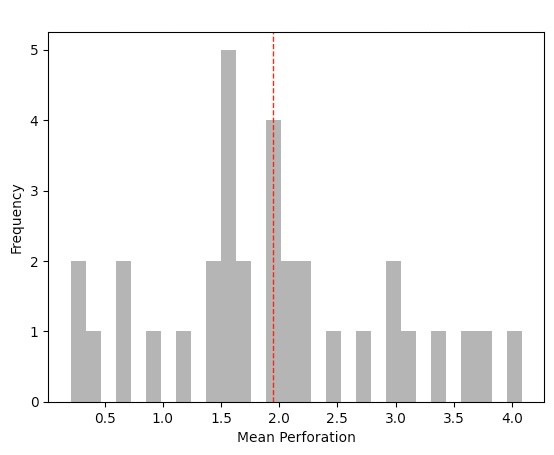}
	\includegraphics[width=0.27\textwidth]{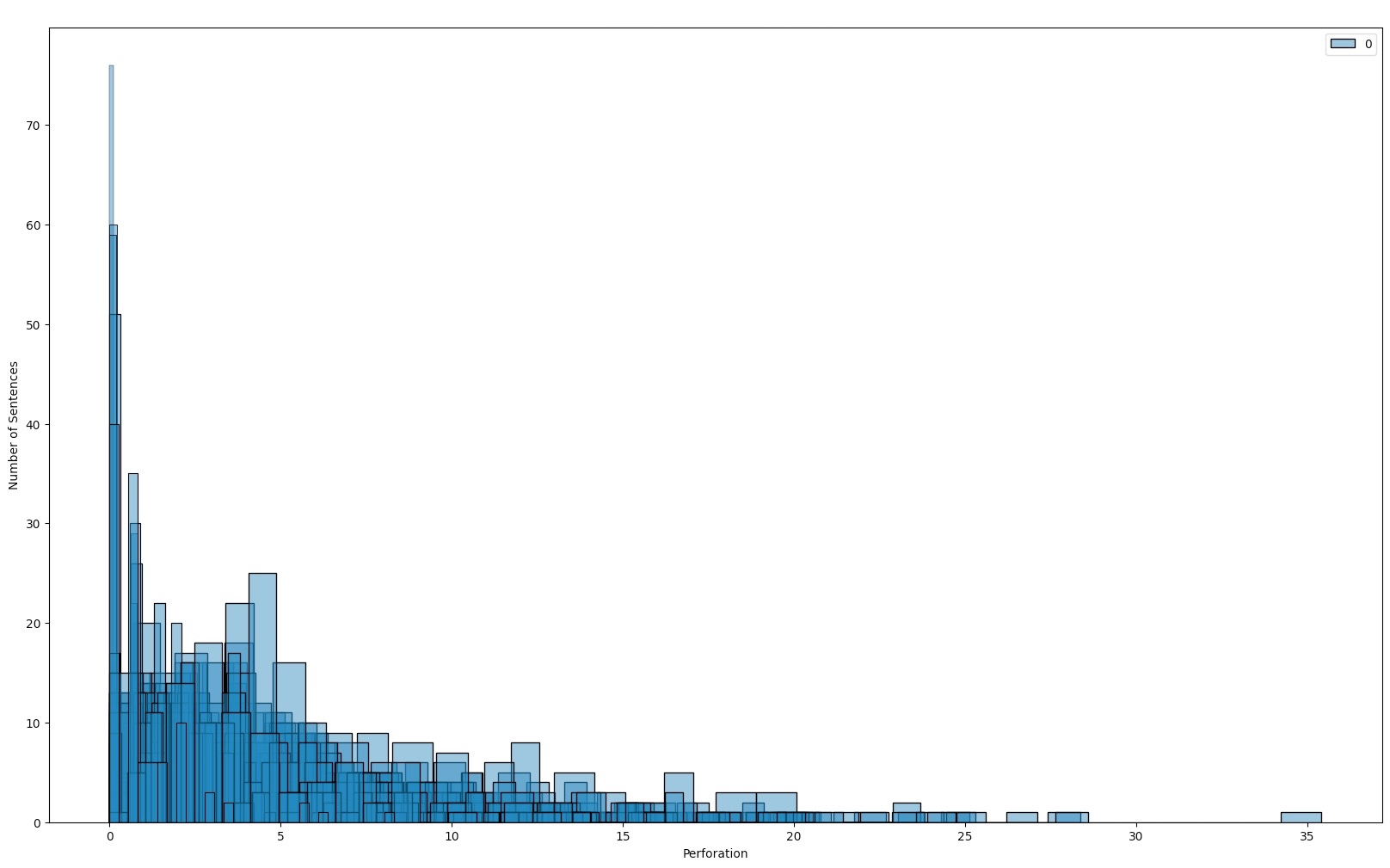}
	\includegraphics[width=0.22\textwidth]{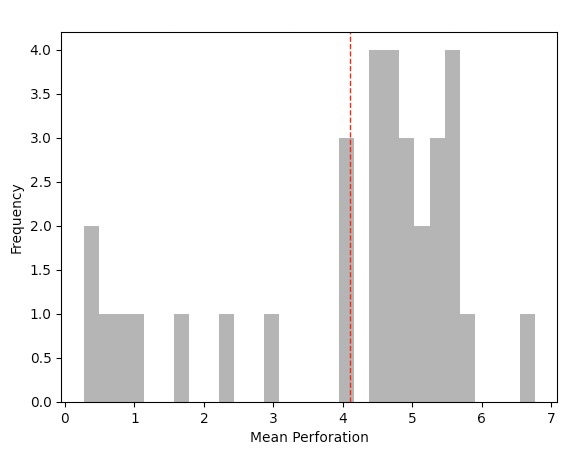}
  \caption{Perforation histograms for 33 transformer layers of LLaMA (overlay), followed by histogram of mean perforation over each layer. Zipf (first pair of plots) and English corpora (the second pair).}
	\label{llama_zipf_eng_mean}
\end{figure}


\end{document}